\documentclass{article}

    \usepackage[final]{neurips_2023}

\usepackage[utf8]{inputenc} 
\usepackage[T1]{fontenc}    
\usepackage[hidelinks]{hyperref}       
\usepackage{url}            
\usepackage{booktabs}       
\usepackage{amsfonts}       
\usepackage{nicefrac}       
\usepackage{microtype}      
\usepackage{xcolor}         

\usepackage{amsmath}
\usepackage{amssymb}
\usepackage{mathtools}
\usepackage{amsthm}

\usepackage[capitalize,noabbrev]{cleveref}

\usepackage{graphicx,psfrag,epsf,color}
\usepackage{enumerate}
\usepackage{subfigure} 
\usepackage{multirow}
\usepackage{array}
\usepackage{bm}
\usepackage{float} 
\usepackage{url}
\setcitestyle{numbers,square}

\usepackage{mathtools, nccmath}

\DeclarePairedDelimiter{\nint}\lfloor\rceil

\theoremstyle{plain}
\newtheorem{theorem}{Theorem}[section]

\theoremstyle{definition}
\newtheorem{definition}[theorem]{Definition}
\theoremstyle{example}
\newtheorem{example}[theorem]{Example}

\theoremstyle{remark}

\newtheorem{condition}{Condition}[section]

\DeclareMathOperator*{\supp}{supp}
\DeclareMathOperator*{\argmin}{arg\,min}

\usepackage[toc,page]{appendix}

\makeatletter
\renewcommand\tableofcontents{%
    \@starttoc{toc}%
}
\makeatother

\usepackage[textsize=tiny]{todonotes}

\usepackage{cleveref}
\crefname{lemma}{lemma}{lemmas}
\Crefname{lemma}{Lemma}{Lemmas}
\crefname{theorem}{thm.}{thms.}
\Crefname{theorem}{Thm.}{Thms.}
\crefname{proposition}{prop.}{props.}
\Crefname{proposition}{Prop.}{Props.}
\crefname{definition}{def.}{defs.}
\Crefname{definition}{Def.}{Defs.}
\creflabelformat{equation}{#1#2#3}
\crefname{equation}{eq.}{eqs.}
\Crefname{equation}{Eq.}{Eqs.}
\Crefname{section}{\S}{\S}
\Crefname{appendix}{App.}{Apps.}
\crefname{figure}{fig.}{figs.}
\Crefname{figure}{Fig.}{Figs.}
\crefname{algorithm}{alg.}{algs.}
\Crefname{algorithm}{Alg.}{Algs.}
\crefname{assumption}{assumption}{assumptions}
\Crefname{assumption}{Assumption}{Assumptions}

\title{On Learning Necessary and Sufficient Causal Graphs}

\author{%
  Hengrui Cai\\
  University of California, Irvine\\
  \texttt{hengrc1@uci.edu} \\
\And
Yixin Wang\\
University of Michigan\\
  \texttt{yixinw@umich.edu}\\
  \AND
  Michael I. Jordan\\
  University of California, Berkeley\\
    \texttt{jordan@cs.berkeley.edu}\\
      \And
Rui Song\\
North Carolina State University\\
 \texttt{songray@gmail.com}
}

\begin{document}

\maketitle

\begin{abstract}
The causal revolution has stimulated interest in understanding complex relationships in various fields. Most of the existing methods aim to discover causal relationships among all variables within a complex large-scale graph. However, in practice, only a small subset of variables in the graph are relevant to the outcomes of interest. Consequently, causal estimation with the full causal graph---particularly given limited data---could lead to numerous \textit{falsely discovered, spurious} variables that exhibit high correlation with, but exert no causal impact on, the target outcome. In this paper, we propose learning a class of \textit{necessary and sufficient causal graphs (NSCG)} that exclusively comprises causally relevant variables for an outcome of interest, which we term \textit{causal features}. The key idea is to employ \textit{probabilities of causation} to systematically evaluate the importance of features in the causal graph, allowing us to identify a subgraph relevant to the outcome of interest. To learn NSCG from data, we develop a \textit{necessary and sufficient causal structural learning (NSCSL)} algorithm, by establishing theoretical properties and relationships between probabilities of causation and natural causal effects of features. Across empirical studies of simulated and real data, we demonstrate that NSCSL outperforms existing algorithms and can reveal crucial yeast genes for target heritable traits of interest.

\end{abstract}




\section{Introduction}

Causal discovery has gained significant attention in recent years for disentangling complex causal relationships in various fields. Building upon the causal graphical model \citep[see e.g., ]
[]{pearl2009causal}, many causal structural learning algorithms have been developed \citep[see e.g.,][]{spirtes2000constructing,chickering2002optimal,shimizu2006linear,kalisch2007estimating,buhlmann2014cam,ramsey2017million,zheng2018dags,yu2019dag,zhu2019causal,cai2020anoce} to infer the causal knowledge (e.g., causal graphs) from observed data. These algorithms are based on the assumption of causal sufficiency (the absence of unmeasured confounders). In real-world applications, to satisfy such an assumption, we strive to learn large-scale causal graphs \citep[see e.g.,][]{nandy2017estimating,chakrabortty2018inference,tang2020long,niu2021counterfactual}, in the hope of \textbf{ sufficiently} describing how an outcome of interest depends on its relevant variables.

\begin{figure}[!t] 
\centering 
\vspace{-0.3cm} 
\begin{subfigure} 
  \centering
  \includegraphics[width=0.35\linewidth]{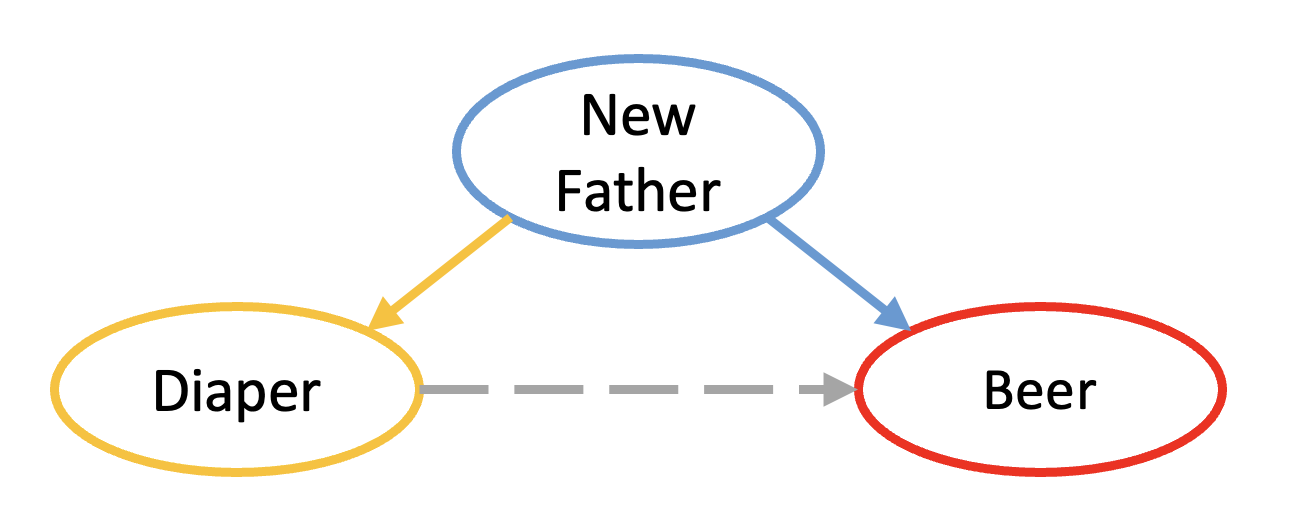} 
\end{subfigure} 
~~~~~~~~~~~~~~~
\begin{subfigure} 
  \centering
  \includegraphics[width=0.35\linewidth]{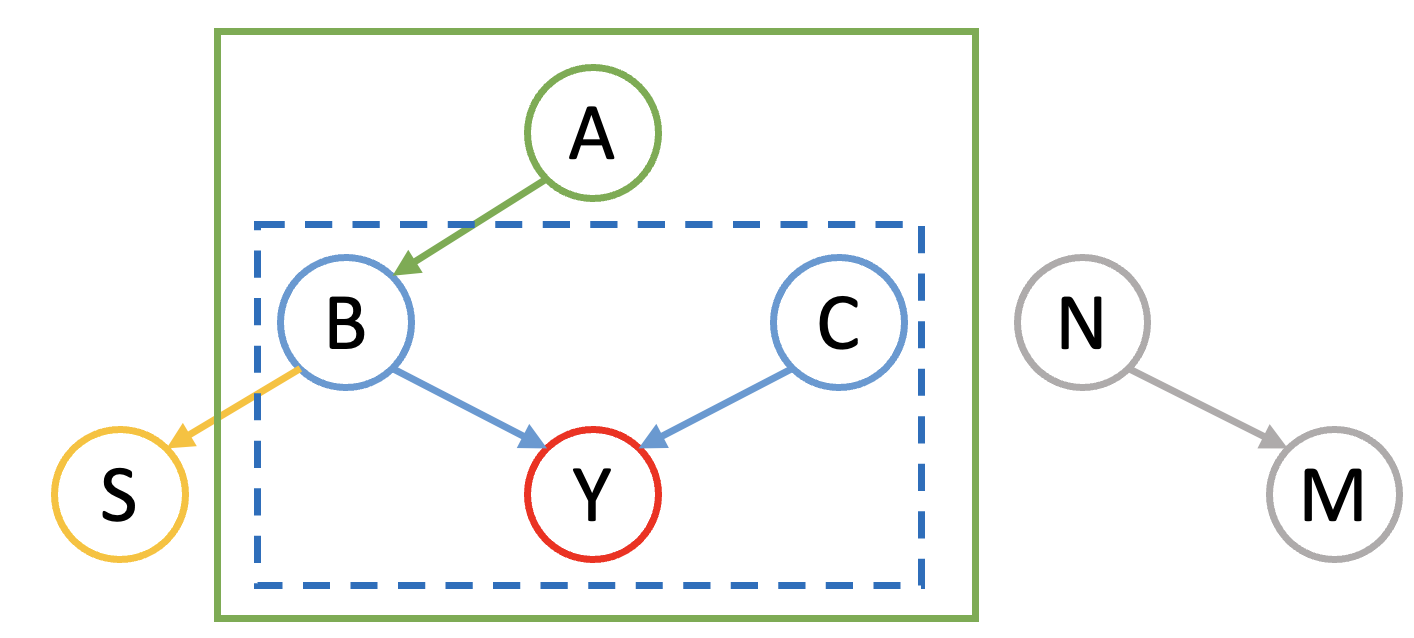} 
\end{subfigure}  
\vspace{-0.3cm} 
\caption{\textbf{Left}: Illustration of the causal relationship between the customer being a new father or not, beer purchasing, and diaper purchasing, where solid lines represent the true model, and the dashed line corresponds to the spurious correlation between beer purchasing and diaper purchasing. \textbf{Right}: Relationship between various causal structures. The nodes $A$, $B$, and $C$ belong to the necessary and sufficient causal graph for the desired result $Y$ and are represented within the solid green square. Among them, nodes $B$ and $C$ are members of the Markov blanket of $Y$, enclosed by the blue square. The node $S$ is the spurious variable for $Y$, while the nodes $N$ and $M$ are not related to the target.} \label{fig:0}
\vspace{-0.3cm}
\end{figure}

In addition to sufficiency, it is also crucial to account for the concept of \textbf{necessity} by excluding redundant variables in explaining the outcome of interest. Failure to do so can result in the inclusion of spurious variables in the learned causal graphs, which are highly correlated but have no causal impact on the outcome. These variables can impede causal estimation with limited data and lead to falsely discovered spurious relationships, leading to poor generalization performance for downstream prediction \citep{scholkopf2021toward}. For example, it might be observed that men aged 30 to 40 who buy diapers are also likely to buy beer. However, beer purchase is a spurious feature for diaper purchases: their correlation is not necessarily causal, as both purchases might be confounded by a shared cause, such as new fathers buying diapers for childcare while also buying beer to alleviate stress. Therefore, simply increasing the availability of diapers or beer will not causally improve the demand for the other (see also \Cref{fig:0}(left)).

Furthermore, the number of variables causally relevant to the outcome of interest is often considerably smaller than the number of variables included in estimating a causal graph (see  \Cref{fig:0}(right)). For example, while an individual's genome may encompass 4 to 5 million single nucleotide polymorphisms (SNPs), only a limited number of non-spurious genes or proteins are found to systematically regulate the expression of the phenotype of interest \citep[e.g.,][]{chakrabortty2018inference}. Similarly, in natural language processing tasks, excluding spurious embeddings such as writing style and dialect can enhance model accuracy and downstream prediction performance \citep[e.g.,][]{feder2021causal}. Thus, a more parsimonious causal graph is required to unveil the {necessary} and {sufficient} causal dependencies.

In this work, we focus on learning \textit{necessary and sufficient causal graphs} (NSCG) that only contain causally relevant variables (which we term \textit{causal features}) for an outcome of interest, offering a compact representation of causal graphs for a target outcome. Our \textbf{contributions} are three-fold. 

$\bullet$ We propose the notion of  NSCG (see an illustration inside the green solid square in the right panel of \Cref{fig:0}). The key idea is to leverage the marginal and conditional probabilities of causation (POC) to systematically characterize the importance of variables (a.k.a. features).\\
$\bullet$ We establish theoretical properties and relationships between POC and the natural causal effects of features, and derive the conditions under which they are equivalent, with lower bounds provided for identification. The natural causal effects of features have explicit forms under parametric models such as the linear structural equation model, enabling convenient estimation of the POC from observed data.\\
$\bullet$ To select necessary and sufficient features 
in causal graphs, 
we propose a necessary and sufficient causal structural learning algorithm (NSCSL) to learn an NSCG containing all necessary and sufficient causes without unnecessary spurious features. This 
enables feature selection for causal discovery.

The proposed method provides concise explanations of causal relationships with high-dimensional data (i.e., with a large number of variables). Empirical studies in simulated datasets show that NSCSL outperforms existing algorithms in distilling relevant subgraphs for outcomes of interest; NSCSL can also identify important quantitative trait loci for the yeast and the causal protein signaling network for single cell data, as demonstrated in real data analyses.



 
 \vspace{-0.2cm}
\subsection{Related Works}
\vspace{-0.15cm}
The literature on \textit{causal structural learning} can be broadly classified into three classes. The first class of methods focuses on using local conditional independence tests to identify the causal skeleton and determine the direction of the edges, such as the PC algorithm~\citep{spirtes2000constructing,kalisch2007estimating,spirtes2013causal}. The second class of methods uses functional causal models with additional assumptions about the data distribution, including ICA-LiNGAM \citep{shimizu2006linear} and the causal additive model (CAM) \citep{buhlmann2014cam}. The third class, the score-based methods, includes greedy equivalence search (GES)~\citep{chickering2002optimal,ramsey2017million,huang2018generalized} and acyclicity optimization methods~\citep{zheng2018dags}. Refer to \citep{yu2019dag,zhu2019causal,lachapelle2019gradient,cai2020anoce,zheng2020learning,vowels2021d} for additional cutting-edge causal structural learning methods. Yet, these works do not consider the necessity of the variables incorporated in the causal graph, i.e., whether the variables are causally relevant to the outcome. Consequently, such algorithms can often produce a redundant or potentially misleading graph, as depicted in \Cref{fig:0} (right).


Our work also links to \textit{ feature selections}
\citep[see an overview in][]{kumar2014feature}. Despite the extensive literature, only a few studies have examined variable selection in causal graphs. One notable exception is \citet{aliferis2010local}, which uses the concept of the Markov blanket to construct a local causal graph for the target variable of interest. In this context, a Markov blanket of a variable $Y$ is the minimal variable subset conditioned upon which all other variables become probabilistically independent of $Y$. Consequently, their algorithm uncovers only direct parents or children in the identified causal graph (such as the blue dotted square in the right panel of \Cref{fig:0}) and thereby overlooks the ancestors that contain atavistic information and indirectly influence the outcome. Recent works \citep{lee2018structural,lee2020characterizing} consider a minimal sufficient action set in bandits. Yet, these methods \citep[also see][]{janzing2013quantifying,janzing2020feature} rely on 
a true or known graph. 
We instead propose to simultaneously learn the causal graph and select the causal features.



Lastly, our work is connected to the body of research on \textit{probability of causation} \citep[e.g.,][]{pearl2000models,tian2000probabilities,pmlr-v202-zhang23ap}, which delineates the necessity and sufficiency of features for the outcome of interest. Recently, \citet{wang2021desiderata} introduced this concept into representation learning, formulating the non-spuriousness and efficiency of representations by generalizing the probabilities of causation to accommodate low-dimensional representations of high-dimensional data. However, these works primarily concentrate on the identification of probabilities of causation, assuming that the causal graph among the variables under consideration is known with causally independent features. We address this gap in our work by incorporating the notion of probabilities of causation into learning complex causal graphs.


 \vspace{-0.25cm}
\section{Framework}\label{sec:2} 
\vspace{-0.2cm}
\textbf{Graph terminology.} Consider a graph $\mathcal{G} =(\boldsymbol{X},\boldsymbol{D}_{\boldsymbol{X}})$ with a node set $\boldsymbol{X}$ and an edge set $\boldsymbol{D}_{\boldsymbol{X}}$ that encompasses all edges in $\mathcal{G}$ for nodes $\boldsymbol{X}$. A node $X_i$ is said to be a parent of $X_j$ if there is a directed edge from $X_i$ to $X_j$, i.e., $X_i$ is a direct cause of $X_j$. A node $X_k$ is said to be an ancestor of $X_j$ if there is a directed path from $X_k$ to $X_j$ regulated by at least one additional node $X_i$ for $i\not =k$ and $i \not =j$, i.e., $X_k$ is an indirect cause of $X_j$. Let the set of all parents/ancestors of node $X_j$ in $\mathcal{G}$ as $\textsc{PA}_{X_j} (\mathcal{G})$. A directed graph $\mathcal{G}$ that does not contain directed cycles is called a directed acyclic graph (DAG). 
The structural causal model (SCM) characterizes the causal relationship among $|\boldsymbol{X}|=d$ nodes via a DAG $\mathcal{G}$ and noises $\boldsymbol{e}_{\boldsymbol{X}} = [e_{X_1},\cdots,e_{X_d}]^\top$ such that
$X_i := h_i\{\textsc{PA}_{X_i} (\mathcal{G}), e_{X_i}\}$ for some unknown $h_i$ and $i=1,\cdots,d$.

 \textbf{Notations and assumptions.}   Denote $\boldsymbol{O}= (\boldsymbol{Z}, Y)$ as a collection of nodes that contains features $\boldsymbol{Z}=[Z_1,\cdots, Z_p]^\top \in \mathcal{Z}\subset \mathbb{R}^p$ and a discrete outcome of interest as $Y \in \mathcal{L} =\{y_1,\cdots,y_l\}$ for $l$ different values. Here, the features 
 can be intervened, such as treatment and mediators. Let $Y(\boldsymbol{Z}=\boldsymbol{z})$ be the potential value of $Y$ that would be observed after setting variable $\boldsymbol{Z}$ as $\boldsymbol{z}$. This is equivalent to the value of $Y$ by imposing a `do-operator' of $do(\boldsymbol{Z}=\boldsymbol{z})$ as in \citet{pearl2009causal}. Similarly, one can define the potential outcome,  $Y(Z_i=z_i)$, by setting an individual variable $Z_i$ as $z_i$, while keeping the rest of the model unchanged. Suppose there exists an SCM that characterizes the causal relationship among $\boldsymbol{O}$, with its DAG as $\mathcal{G}_{\boldsymbol{O}}$. 
 A notation and abbreviation table is provided in \Cref{app_not}. Following the causal inference literature \citep[see e.g.,][]{rosenbaum1983central,pearl2000models,pearl2009causal,wang2021desiderata}, we assume: 
 
 \noindent (A1). \textbf{Consistency}: 
$ 
\boldsymbol{Z}=\boldsymbol{z} \leftrightarrow Y(\boldsymbol{Z}=\boldsymbol{z})=Y, \forall \boldsymbol{z}\in \mathcal{Z}.
$\\
    \noindent (A2). \textbf{Ignorability}: 
  $ \text{(i) }Y(\boldsymbol{Z}=\boldsymbol{z}) \perp \boldsymbol{Z}, \forall \boldsymbol{z}\in \mathcal{Z};\quad
 \text{(ii) } Y(Z_i=z_i)\perp Z_i | \textsc{PA}_{Z_i\cup Y} (\mathcal{G}_{\boldsymbol{O}}) ,\forall z_i\in \mathcal{Z}_i.$
   
  

 Here, (A1) implies that the outcome observed for each unit under study with features as $\boldsymbol{z}$ is identical to the outcome we would have observed had that unit been set with features $\boldsymbol{Z}=\boldsymbol{z}$. In addition, since we include as many confounders as possible, the ignorability assumption in (A2), also known as the no unmeasured confounderness assumption, is satisfied. 

\vspace{-0.25cm}
\section{Necessary and Sufficient Causal Graphs}
\vspace{-0.2cm}  
We care about a subset or a function of $\boldsymbol{Z}$, denoted as $\boldsymbol{X}=[X_1,\cdots,X_d]^\top$ (of $d$ dimension with possibly $d\ll p$), which indeed captures the causal relationship between $\boldsymbol{Z}$ and $Y$. To be specific, let an SCM for causal nodes ${\boldsymbol{V}} = (\boldsymbol{X}, Y)$ with its DAG as $\mathcal{G}_{\boldsymbol{V}} = ({\boldsymbol{V}},\boldsymbol{D}_{\boldsymbol{V}})$ and $e_{\boldsymbol{V}}$ as a $d+1$ dimensional independent noise, to characterize the causal relationship between $\boldsymbol{X}$ and $Y$.  Let $ \mathbb{P}_{\mathcal{G}}$ be the mass/density function for an SCM with its DAG ${\mathcal{G}}$. Following the causal (or disentangled) factorization in the causal graphical model \citep{pearl2009causal}, we define the \textit{sufficient} causal graph as follows. 
  \begin{definition}{(Sufficient Graph)}\label{defi_sufficiency}
  The graph $\mathcal{G}_{\boldsymbol{V}}$ is a \textit{sufficient} causal graph to capture the causal relationship among $\boldsymbol{Z}$ and $Y$ with $\boldsymbol{X}\subset \boldsymbol{Z} $ or $\boldsymbol{X} =f(\boldsymbol{Z}) $ (where  $f$ is within a countable or Vapnik-Chervonenkis (VC) class) if 
 $\mathbb{P}_{\mathcal{G}_{\boldsymbol{V}}}\{Y|\textsc{PA}_{Y} (\mathcal{G}_{\boldsymbol{V}})\} \prod_{X_i \in \textsc{PA}_{Y} (\mathcal{G}_{\boldsymbol{V}})}  \mathbb{P}_{\mathcal{G}_{\boldsymbol{V}}}\{X_i|\textsc{PA}_{X_i} (\mathcal{G}_{\boldsymbol{V}})\}$
 $=  \mathbb{P}_{\mathcal{G}_{\boldsymbol{O}}}\{Y|\textsc{PA}_{Y} (\mathcal{G}_{\boldsymbol{O}})\} \prod_{Z_i \in \textsc{PA}_{Y} (\mathcal{G}_{\boldsymbol{O}})}  \mathbb{P}_{\mathcal{G}_{\boldsymbol{O}}}\{Z_i|\textsc{PA}_{Z_i} (\mathcal{G}_{\boldsymbol{O}})\}.$ 
\end{definition}
Here, \Cref{defi_sufficiency} refers to a sub-structure $\mathcal{G}_{\boldsymbol{V}}$ (from the whole graph $\mathcal{G}_{\boldsymbol{O}}$) containing all directed edges or paths towards $Y$, making it sufficient to describe how $Y$ depends on all its ancestors. Then, the causal graph $\mathcal{G}_{\boldsymbol{V}}$ is said to be \textit{necessary} and \textit{sufficient} if it satisfies the following definition.  
 
   \begin{definition}{(Necessary and Sufficient Graph)}\label{defi_necessity}
   Suppose $\mathcal{G}_{\boldsymbol{V}}$ satisfies \Cref{defi_sufficiency}, then $\mathcal{G}_{\boldsymbol{V}}$ is a \textit{necessary} and \textit{sufficient} causal graph to capture the causal relationship among $\boldsymbol{Z}$ and $Y$ 
   if for any true subset $\boldsymbol{W}$ of $\boldsymbol{X}$, i.e., $\boldsymbol{W}\subset \boldsymbol{X}$ or $\boldsymbol{W}= g(\boldsymbol{X})$ (where $g$ is within a countable or VC class), with  ${\boldsymbol{U}} = (\boldsymbol{W}, Y)$, we have 
 $\mathbb{P}_{\mathcal{G}_{\boldsymbol{V}}}\{Y|\textsc{PA}_{Y} (\mathcal{G}_{\boldsymbol{V}})\} \prod_{X_i \in \textsc{PA}_{Y} (\mathcal{G}_{\boldsymbol{V}})}  \mathbb{P}_{\mathcal{G}_{\boldsymbol{V}}}\{X_i|\textsc{PA}_{X_i} (\mathcal{G}_{\boldsymbol{V}})\}$
 $\not=$ $  \mathbb{P}_{\mathcal{G}_{\boldsymbol{U}}}\{Y|\textsc{PA}_{Y} (\mathcal{G}_{\boldsymbol{U}})\} \prod_{W_i \in \textsc{PA}_{Y} (\mathcal{G}_{\boldsymbol{U}})}  \mathbb{P}_{\mathcal{G}_{\boldsymbol{U}}}\{W_i|\textsc{PA}_{W_i} (\mathcal{G}_{\boldsymbol{U}})\} ,$ 
 where $\mathcal{G}_{\boldsymbol{U}}$ is the causal graph for ${\boldsymbol{U}}$.
\end{definition}
Therefore, by \Cref{defi_necessity}, we can further identify the \textit{minimal sub-structure} $\mathcal{G}_{\boldsymbol{V}}$ 
which includes only all directed edges or paths leading to $Y$. 
 The goal is to learn such a necessary and sufficient causal graph (NSCG) ${\mathcal{G}_{\boldsymbol{V}}}$ from the observed data denoted as $\{\boldsymbol{o}^{(j)}= (\boldsymbol{z}^{(j)}, y^{(j)})\}_{1\leq j\leq n}$ with sample size $n$, by identifying the latent causal features $\boldsymbol{X}$. Denote the resulting estimated graph as $\widehat{\mathcal{G}}_{\boldsymbol{V}}$.

\vspace{-0.25cm}
\section{Probability of Causation and Causal Effects} \label{sec:poc}
\vspace{-0.15cm} 

Obtaining an NSCG ${\mathcal{G}_{\boldsymbol{V}}}$  directly based on \Cref{defi_necessity} poses several challenges, as the latent causal features $\boldsymbol{X}$ driving the causal graph remain unknown. A na\"{\i}ve approach is to search all different combinations of $\boldsymbol{Z}$ for a candidate of $\boldsymbol{X}$ such that \Cref{defi_necessity} holds, which yields a complexity of $\mathcal{O}(p^p)$. 
This motivates us to assess the necessity and sufficiency of features in determining the outcome by introducing the concepts of probabilities of causation and causal effects, which will be elaborated on and interconnected in this section.

\vspace{-0.2cm}
\subsection{Probabilities of Causation and Lower Bounds}
\vspace{-0.15cm} 
The probabilities of a feature being necessary and sufficient, known as the probability of causation (POC), have been proposed and studied \citep[see ][]{pearl2000models,tian2000probabilities,wang2021desiderata}. Specifically, the probability of necessity and sufficiency (PNS) of feature $\boldsymbol{Z}$ is first defined in \citet{tian2000probabilities} as follows.
 \begin{definition}{PNS in \citet{tian2000probabilities} with a univariate binary feature $Z$:}\label{defi_PNS}
 \begin{equation*}
PNS 
\equiv  \mathbb{P}\{Y({Z} \not ={z})\not =y,Y({Z} ={z}) =y\}
\underset{by (A1)}{=} \mathbb{P}({Z} ={z}, Y=y)\cdot PN
+ \mathbb{P}(Z\not=z, Y\not=y)\cdot PS
,
 \end{equation*} 
 where the probability of necessity (PN) is  
 $
 PN
 =\mathbb{P}\{Y({Z} \not ={z})\not =y|{Z} ={z}, Y=y\},$  
  and the probability of sufficiency (PS) is  $ 
  PS
  =\mathbb{P}\{Y({Z} ={z}) =y|{Z} \not ={z}, Y\not =y\}.$ 
\end{definition}
 \vspace{-0.1cm}

 The second equation in \Cref{defi_PNS} holds under (A1) \citep[see details in][]{pearl2000models,tian2000probabilities}. The PN score reflects the necessity of ${Z}$ by evaluating the probability of the outcome becoming worse if revising the features given the good outcome observed. Similarly, the PS score indicates the sufficiency of ${Z}$ by evaluating the probability of the outcome becoming better if changing the features given the bad outcome observed. Therefore, the PNS score shows the causal importance of the features by combining necessary and sufficient properties. The above definition can be generalized to multivariate cases for nonbinary features \citep[see e.g.,][]{wang2021desiderata} to quantify the POC of an individual feature $Z_i$. Let $\boldsymbol{Z}_{-i}\equiv \boldsymbol{Z}\setminus Z_i$ be the set of complementary variables of $Z_i$. In the following, we consider two different POCs for $Z_i$ by extending the work of \citet{wang2021desiderata}. 
    \begin{definition} {Marginal POC (M-POC) for $Z_i$:}\label{defi_tpcPNS_i} 
 \begin{equation*}
  \begin{split}
\text{M-POC}_i(y)
\equiv \mathbb{P}\{Y(Z_i\not =z_i)\not =y,Y(Z_i =z_i) =y\}.
 \end{split}
 \end{equation*}
   \end{definition}
      \vspace{-0.1cm}
%
\vspace{-0.2cm}

 \begin{definition}{Conditional POC (C-POC) for $Z_i$:}\label{defi_dpcPNS_i} 
 \begin{equation*} 
  \text{C-POC}_i(y)
\equiv \mathbb{P}\{ Y(Z_i\not =z_i,\boldsymbol{Z}_{-i}=\boldsymbol{z}_{-i})\not =y, 
 Y(Z_i =z_i,\boldsymbol{Z}_{-i}=\boldsymbol{z}_{-i}) =y\}.
 \end{equation*}
\end{definition} 

   \vspace{-0.1cm}
   
 We introduce the marginal POC (M-POC) as a novel quantity in the literature to summarize the overall causal importance of an individual feature in determining the outcome's value. The conditional POC (C-POC) in \Cref{defi_dpcPNS_i} corresponds to the conditional PNS in \citet{wang2021desiderata}, which quantifies the likelihood of an individual feature being a direct and significant cause of the outcome while holding other features constant. As per Section 9.2.3 in \citet{pearl2000models}, the PNS in \Cref{defi_PNS} is not estimable unless additional conditions (monotonicity) are specified. To alleviate such a condition for identifying \Cref{defi_dpcPNS_i,defi_tpcPNS_i}, we derive the lower bounds for the proposed POCs as follows. 
 \begin{theorem}{(Lower Bound of Probabilities of Causation)}\label{thm1}
 Suppose (A1) and (A2) hold. 
 Then 
 \begin{equation*}
  \begin{split}
      \text{M-POC}_i(y)&
  \geq\mathbb{P}(Y=y|Z_i =z_i)- \mathbb{P}(Y=y|Z_i \not=z_i),\\
   \text{C-POC}_i(y)&
  \geq\mathbb{P}(Y=y|Z_i =z_i, \boldsymbol{Z}_{-i}=\boldsymbol{z}_{-i}) - \mathbb{P}(Y=y|Z_i \not=z_i, \boldsymbol{Z}_{-i}=\boldsymbol{z}_{-i}).
 \end{split}
   \end{equation*} 
\end{theorem}

 \vspace{-0.2cm}
The proofs of \Cref{thm1} are in \Cref{asec:pf}. 
The lower bound equality holds when an additional monotonicity condition is imposed, with details in \Cref{asec:pf}. The results in \Cref{thm1} allow us to estimate the lower bound of POC from observed data by learning the conditional probability of $Y$ given various combinations of confounders. This, in turn, aids in evaluating the significance of features, with details provided in \Cref{asec:algo}. 
Yet, estimating these conditional probabilities of $Y$ based on high-dimensional features is very challenging \citep[e.g.,][]{shah2018hardness,wang2021desiderata}, which motivates us to consider the corresponding expected mean outcome given different combinations of the confounders. 

 \vspace{-0.2cm}
\subsection{Causal Effects and Connection to POCs}
\vspace{-0.15cm}
To connect the proposed POCs and facilitate the empirical estimation, we introduce the natural total effect (TE) and natural 
 direct effect (DE) for $Z_i$ by extending definitions in \citet{pearl2000models}. 
\begin{definition}{
Natural Causal Effects for $Z_i$:}\label{FSTE_FSDE} 
\begin{eqnarray*} 
&&TE_i  =
\mathbb{E}\{Y(Z_i=z_i+1)\}-\mathbb{E}\{Y(Z_i=z_i)\}
,\\
&&DE_i  = 
\mathbb{E}\{Y(Z_i=z_i+1, \boldsymbol{Z}_{-i}=\boldsymbol{z}_{-i}^{(z_i)})\}-\mathbb{E}\{Y(Z_i=z_i)\},
\end{eqnarray*}
where $\boldsymbol{z}_{-i}^{(z_i)}$ is the value of $ \boldsymbol{Z}_{-i}$ if setting $do(Z_i=z_i)$.
\end{definition} 
\vspace{-0.1cm}

The natural total effect ($TE_i$) can be understood as the marginal change in the outcome when increasing $Z_i$ by one unit. Similarly, the natural direct effect ($DE_i$) represents the conditional change in the outcome when $Z_i$ is increased by one unit, with all other features held constant. Indeed, the natural total and direct causal effects delineated in \citet{pearl2000models} emerge as particular instances of \Cref{FSTE_FSDE} when examining a solitary treatment subject to intervention. By comparing \Cref{FSTE_FSDE} with \Cref{defi_dpcPNS_i,defi_tpcPNS_i}, it is natural to establish the relationship between POCs and  causal effects below.

\begin{theorem}{(Relation between POCs and Causal Effects)}\label{thm2} Define $\delta_M(z_i)\equiv \mathbb{E}\{Y|Z_i =z_i\}- \mathbb{E}\{Y|Z_i \not=z_i\}$ and $\delta_C(z_i)\equiv \mathbb{E}\{Y|Z_i =z_i, \boldsymbol{Z}_{-i}=\boldsymbol{z}_{-i}\} - \mathbb{E}\{Y|Z_i \not=z_i, \boldsymbol{Z}_{-i}=\boldsymbol{z}_{-i}\}$. Suppose (A1)-(A2) hold, then
\vspace{-0.1cm}
 \begin{equation*} 
  \begin{split}
\sum_{y\in \mathcal{L}} y\text{M-POC}_i(y)  \geq \delta_M(z_i), \quad \quad 
\sum_{y\in \mathcal{L}} y\text{C-POC}_i(y)  \geq \delta_C(z_i), 
  \end{split} 
 \end{equation*}  
if $Y$ is nonnegative. Further, if $Z_i$ is binary, 
we have 
\vspace{-0.1cm}
 \begin{equation*} 
  \begin{split}
\min\{\sum_{y\in \mathcal{L}} y\text{M-POC}_i(y) , |TE_i| \} \geq \delta_M(z_i), \quad\quad 
\min\{\sum_{y\in \mathcal{L}} y\text{C-POC}_i(y) , |DE_i|\}\geq \delta_C(z_i).
  \end{split} 
 \end{equation*} 
\end{theorem}
\vspace{-0.3cm}

The proofs of \Cref{thm2} can be found in \Cref{asec:pf}, with the lower bound equality holds when an additional monotonicity condition is imposed. We can summarize the findings of \Cref{thm1,thm2} in two key aspects. First, the marginal and conditional POCs assess the \textit{likelihood} of a feature being spurious, while the absolute values of natural causal effects quantify the \textit{size} of such a spurious effect based on the magnitude of the outcome of interest. Both are lower bounded by the same quantity, that is, the differences in expectations based on the corresponding POC, given non-negative outcomes and binary features. With appropriate data processing, we can transform the outcome to be nonnegative, and thus causal effects become a suitable substitute for POCs. Second, these two approaches exhibit consistency under the monotonicity condition. The natural causal effects of features have explicit forms under parametric models (see details in \Cref{sec:est}), such as the linear structural equation model, enabling convenient estimation of the necessity and sufficiency of features from observed data. This section concludes with a toy example illustrating the identification of spurious features through the proposed causal effects and the distinction between total and direct causal effects.

      \begin{table}[!t]
             \centering
             \vskip -0.1in
             \caption{Causal effects from the customer being a new father or
not ($X_F$) and diaper purchasing ($X_D$) on beer purchasing ($X_B$), where  $\omega_1,\omega_2\in(0,1)$ with $\omega_1+\omega_2=1$ and $c\to 1$, in the estimated causal graph ($X_F \overset{\omega_1}{\longrightarrow}X_B$, and $X_F\overset{c}{\longrightarrow}X_D\overset{\omega_2}{\longrightarrow}X_B$).}\label{tab1} 
             \scalebox{0.8}{
\begin{tabular}{llllll}
\toprule
 Variable Name & Direct Effect &  Total Effect  \\
\midrule
New Father ($X_F$)       & $\omega_1 $              &  $\omega_1 +c\omega_2 \to 1 \quad (>\omega_2)$  \\
\midrule
Diaper  ($X_D$)     & $\omega_2$        &$\omega_2$    \\ 
\bottomrule 
\end{tabular}} 
\vspace{-0.2cm}
\end{table}

\begin{example}{(Recall: Beer and Diaper)} 
Given the causal graph in \Cref{fig:0}(left), consider a linear SCM for the customer being a new father or
not ($X_F$), diaper purchasing ($X_D$), and beer purchasing ($X_B$): $X_D=X_F+e_D$ and $X_B=X_F+e_B$, where $e_D$ and $e_B$ are independent mean zero noises. Since the true SCM is unknown, fitting a linear model $X_B\sim \omega_1X_F+\omega_2X_D$ would result in ambiguous coefficients $\omega_1,\omega_2\in[0,1]$ with $\omega_1+\omega_2=1$. By fitting $X_D\sim cX_F$ and obtaining $c$ close to 1, we have the estimated causal graph as $X_F \overset{\omega_1}{\longrightarrow}X_B$, and $X_F\overset{c}{\longrightarrow}X_D\overset{\omega_2}{\longrightarrow}X_B$. 
Based on \Cref{FSTE_FSDE}, we estimate the corresponding causal effects in Table \ref{tab1}. It can be observed that the total effect of $X_F$ on $X_B$ surpasses the total effect from $X_D$, indicating the spurious nature of the diaper purchasing, which should be removed to form the desired NSCG. Yet, using the direct effect may not be able to distinguish their differences due to the high correlation between $X_F$ and $X_D$ if $\omega_1<\omega_2$.

\end{example}

  \vspace{-0.25cm}
\section{Necessary and Sufficient Causal Structural Learning}
\vspace{-0.15cm}

In this section, we formally present how to learn NSCG. Based on \Cref{thm1,thm2},  a simple solution is to first use a pre-screening process to find necessary and sufficient features from $\boldsymbol{Z}$ that achieve high scores of causation, and then estimate the causal graph among the selected nodes and $Y$ to approximate $\mathcal{G}_{\boldsymbol{V}}$. This approach works for general SCMs while may suffer from overfitting.  Instead of using such a two-step learning, we propose to learn necessary and sufficient features and the causal graph simultaneously through a single-step optimization. To this end, in \Cref{sec:est}, we first introduce the structural equation model in order to provide the closed-form expressions of the proposed causal quantities. The main algorithm based on causal effects is presented in \Cref{sec:algo} for the \textit{linear} model, with the POC-based version available in \Cref{asec:algo} for the \textit{nonlinear} model.



\vspace{-0.2cm}
\subsection{Structural Equation Model and Close Form of Causal Effects}\label{sec:est}
 \vspace{-0.1cm}
 \textbf{Structural equation model. }
We define a selection function $g$ that maps the feature set $\boldsymbol{Z}$ to a subset, aiming to maintain good interpretability. That is, $g:\boldsymbol{Z}\in \mathbb{R}^p \to g(\boldsymbol{Z})\in \mathbb{R}^d$ where $d \ll p$, and we denote the $i$-th dimension of $g(\boldsymbol{Z})$ as $g_i(\boldsymbol{Z})$.  Following the causal structure learning literature \citep{spirtes2000constructing,peters2014causal,zheng2018dags,yu2019dag,zhu2019causal,cai2020anoce}, we assume the Markov and faithfulness conditions and consider a linear structural equation model (LSEM) such that $\{g(\boldsymbol{Z}),Y\}$ is characterized by the pair ($\boldsymbol{B}$, $\boldsymbol{\epsilon}$) as 
  \vspace{-0.1cm}
  \begin{eqnarray}\label{lsem_x}
\begin{bmatrix}
   			g(\boldsymbol{Z})\\
			Y\\
		\end{bmatrix}
		\leftarrow \boldsymbol{B}  \begin{bmatrix}
   			g(\boldsymbol{Z})\\
			Y\\
		\end{bmatrix}+\boldsymbol{\epsilon}
		\equiv \begin{bmatrix}
   			 
			 \boldsymbol{B}_g &0\\
   			\boldsymbol{\theta} &0\\
		\end{bmatrix}
		\begin{bmatrix}
   			g(\boldsymbol{Z})\\
			Y\\
		\end{bmatrix}
		+\begin{bmatrix} 
   			\boldsymbol{\epsilon}_{\boldsymbol{Z}}\\
			\epsilon_Y\\
		\end{bmatrix},
\end{eqnarray}
where $\boldsymbol{B}$ is a $(d+1)\times (d+1)$ weighted adjacent matrix that characterizes the causal relationship among $\{g(\boldsymbol{Z}),Y\}$, and $\boldsymbol{\epsilon}\equiv [\boldsymbol{\epsilon}_{\boldsymbol{Z}}^\top, \epsilon_Y]^\top $ is a $d+1$ dimensional random vector of jointly independent errors. Here, $\boldsymbol{B}$ consists of three components: (1). a $d\times d$ matrix $\boldsymbol{B}_g =\{b_{i,j}\}_{1\leq i\leq d,1\leq j\leq d}$ with $b_{i,j}$ as the weight of the edge $g_i(\boldsymbol{Z}_i)\rightarrow g_j(\boldsymbol{Z}_i)$ if exists and $b_{i,j}=0$ otherwise; (2) a $1\times d$ vector $\boldsymbol{\theta}=[\theta_1,\cdots,\theta_p]$ for $\theta_i$ presenting the weight of the direct edge $g_i(\boldsymbol{Z})\rightarrow Y$; and (3). a $(d+1)\times 1$ zero vector indicating the outcome of interest $Y$ 
cannot be any parent of the features. 
Without further assumptions, 
the model in  \eqref{lsem_x} given a particular selector $g$ can be identified only up to a Markov equivalence class (MEC) 
\citep{spirtes2000constructing,peters2014causal}. In the following, we focus on cases where the DAG can be uniquely identifiable, such as LSEM with  Gaussian noises of equal variance \citep{spirtes2000constructing, peters2017elements,peters2014identifiability}, and linear model with non-Gaussian noise \citep{shimizu2006linear,zheng2020learning}. See more details and extensions to MEC in  \Cref{asec:mec}.

\textbf{Close form and estimation of causal effects.}   We next provide the close forms of the causal effects in \Cref{FSTE_FSDE} under the model in  \eqref{lsem_x}. 
Recall that $\theta_i$ presents the weight of the direct edge $g(\boldsymbol{Z})_i\rightarrow Y$. According to  \eqref{lsem_x} and \Cref{FSTE_FSDE}, we have  
\vspace{-0.2cm}
\begin{eqnarray*}
DE_i(\boldsymbol{B};g)= \theta_i.
\end{eqnarray*}
The total causal effect can be quantified by the path method \citep[see e.g., ][]{Wright1921CorrelationAndCausation,nandy2017estimating}. Specifically, the causal effect of $g_i(\boldsymbol{Z})$ on $g_j(\boldsymbol{Z})$ along a directed path from $g_i(\boldsymbol{Z})\rightarrow g_j(\boldsymbol{Z})$ in $\mathcal{G}$ can be calculated by multiplying all edge weights along the path, under LSEM. 
Denote the set of directed paths that starts with $g_i(\boldsymbol{Z})$ and ends with $Y$ as $\pi_{i}=\{ g_i(\boldsymbol{Z})\rightarrow \cdots \rightarrow Y\}$ with the size as $m_{i}$. Then the causal effect of $g_i(\boldsymbol{Z})$ on $Y$ through the directed path $\pi_{i}^{(k)} =\{i,l_1, \cdots, l_{\tau_k}, d+1\} \in \pi_{i}$ with length ${\tau_k}+1$ is 
$ PE\{\pi_{i}^{(k)}\} = 
b_{i,l_1} \cdots b_{l_{\tau_k},(d+1)}$,
by the path method, where $b_{i,j}$ is the weight of the edge $g_i(\boldsymbol{Z})\rightarrow g_j(\boldsymbol{Z})$ if it exists, and $b_{i,j}=0$ otherwise, for $i, j \in \{1,\cdots, d\}$, and $b_{l_{\tau_k},(d+1)} = \theta_{l_{\tau_k}}$ as the direct edge from $g_{l_{\tau_k}}(\boldsymbol{Z})$ to $Y$. Thus, 
\vspace{-0.25cm}
\begin{align*}
TE_i(\boldsymbol{B};g)= \sum_{k=1}^{m_i} PE\{\pi_{i}^{(k)}\}.
\end{align*}
Both $TE_i$ and $DE_i$ can be explicitly calculated given a matrix $\boldsymbol{B}$ under a selector $g$. We denote their estimates as $\widehat{TE}_i$ and $\widehat{DE}_i$ given the estimated matrix $\widehat{\boldsymbol{B}}$ and $g$. 
 
\vspace{-0.15cm}
\subsection{Learning Algorithm based on Causal Effects} \label{sec:algo}
\vspace{-0.1cm}
The primary algorithm based on causal effects comprises three steps that quantify two sources of loss and learn the causal graph, specifically: loss of causal structural learning, loss of discovering causal features, and minimizing the overall loss to learn NSCG based on data $\{\boldsymbol{o}^{(j)}= (\boldsymbol{z}^{(j)}, y^{(j)})\}_{1\leq j\leq n}$.

 \textbf{Step 1: Form the loss from causal structural learning.} 
To estimate the matrix $\boldsymbol{B}$ in  \eqref{lsem_x}, we adopt the acyclicity constraint \citep{yu2019dag,zheng2018dags}   as $
h_1(\boldsymbol{B})\equiv \text{tr}\big[(I_{d+1}+t \boldsymbol{B} \circ \boldsymbol{B})^{d+1}\big]-(d+1)=0, $
where $I_{d+1}$ is a $d+1$-dimensional identity matrix, and $\text{tr}(\cdot)$ is the trace of a matrix and $t$ is a hyperparameter that depends on the estimated largest eigenvalue of $\boldsymbol{B}$. 
The first loss by the augmented Lagrangian is 
  \vspace{-0.1cm}
\begin{eqnarray}\label{loss1}
L_1(\boldsymbol{B},g,\theta,\lambda_1|\{\boldsymbol{o}^{(j)}\}
)=f(\boldsymbol{B},g,\theta|\{\boldsymbol{o}^{(j)}\})+\lambda_1 h_1(\boldsymbol{B}),
\end{eqnarray}
where $f(\boldsymbol{B},g,\theta|\{\boldsymbol{o}^{(j)}\})$ is some loss such as the least square error in NOTEARS \citep{zheng2018dags} or the Kullback-Leibler divergence in DAG-GNN \citep{yu2019dag} with parameters $\theta$, and $\lambda_1$ is the Lagrange multiplier. Other causal structural leaning algorithms \citep[see e.g.,][]{spirtes2000constructing,chickering2002optimal,shimizu2006linear,kalisch2007estimating,buhlmann2014cam,ramsey2017million,zhu2019causal} can also be applied by formulating the corresponding score or loss function.

 \textbf{Step 2: Constraints for causal relevance and causal identifiability.} We next measure the causal relevance of the selection function $g$  by 
 the natural causal effects.  
 We convert the outcome $Y$ to be nonnegative. 
 According to \Cref{thm2}, we can avoid the estimation of POCs by using the related causal effects in \Cref{FSTE_FSDE} with their explicit expressions. This part of loss thus becomes 
   \vspace{-0.1cm}
   \begin{equation}\label{loss2_causaleffects}
L_2^{CE}(\boldsymbol{B},g,\gamma|\{\boldsymbol{o}^{(j)}\}
)=- \sum_{i=1}^{d} |\widehat{CE}_i(\boldsymbol{B};g)| +\sum_{i=1}^{d+1} |b_{i,d+1}| +\gamma |g|.
 \end{equation} 
Here, $\widehat{CE}_i$ can either take the estimated $DE_i$ or the estimated $TE_i$ given the matrix $\boldsymbol{B}$ and $g$, as detailed in \Cref{sec:est}. The second term corresponds to the causal identification constraint on the last column of $\boldsymbol{B}$, requiring all elements to be zeros as in \eqref{lsem_x}. This constraint restricts the causal structural learning to a smaller class of DAGs. Finally, $|g|$ denotes the number of selected nodes in $g$, with a penalty $\gamma$ to control the complexity of the selector.

 \textbf{Step 3: Necessary and sufficient causal structural learning.} Combining two sources of loss functions in 
 \eqref{loss1} with \eqref{loss2_causaleffects}, leads to the objective as
\begin{eqnarray}\label{loss}
 \min_{\boldsymbol{B},g} \left[ L_1(\boldsymbol{B},g,\theta,\lambda_1|\{\boldsymbol{o}^{(j)}\}
 ) 
 +\alpha L_2^{CE}(\boldsymbol{B},g,\gamma|\{\boldsymbol{o}^{(j)}\}
 ) \right],
 \end{eqnarray}
 where $\alpha$ can be reviewed as a trade-off parameter between two loss functions. Next, we provide a solution for \eqref{loss} \textit{without} tuning $\alpha$. Specifically, based on no unmeasured confounders in (A2), we can calculate the highest causal effects that could be achieved given all variables without any penalty. 
 Denote the estimated highest absolute causal effects 
 in data as $\delta^*$.  
The goal is to find a subset 
of $\boldsymbol{Z}$ such that $g(\boldsymbol{Z})$ achieves a similar level of necessity and sufficiency, i.e., the resulting score is close to $\delta^*$. Hence, we set the second loss as a constraint by comparing the overall causal effects of the selected nodes with the highest reference over the entire observed feature space, i.e.,
\vspace{-0.25cm}
 \begin{eqnarray*}
  h_2(\boldsymbol{B};g)=\delta^*- \sum_{i=1}^{d} |\widehat{CE}_i(\boldsymbol{B};g)| +\sum_{i=1}^{d+1} |b_{i,d+1}| ,
 \end{eqnarray*}   
 should be approaching 0 given a good selector $g$.  
 This yields a new objective function as
 \vspace{-0.1cm}
 \begin{eqnarray}\label{loss_new}
 \min_{\boldsymbol{B},g}& f(\boldsymbol{B},g,\theta|\{\boldsymbol{o}^{(j)}\}
 )+\lambda_1 h_1(\boldsymbol{B})+\lambda_2 h_2(\boldsymbol{B};g) +c|h_1(\boldsymbol{B})|^2+d|h_2(\boldsymbol{B};g)|^2+\gamma |g|,
 \end{eqnarray}
 where $\lambda_2$ is the Lagrange multiplier for the new constraint, and $c$ and $d$ are penalty terms. To minimize the loss in \eqref{loss_new} and satisfy both $h_1(B)\to 0$ and $h_2(\boldsymbol{B};g)\to 0$, we simultaneously update $\lambda_1$ and $\lambda_2$ and increase $c$ and $d$ to infinity, by modifying the updating technique \citep[see e.g., ][]{zheng2018dags,yu2019dag} for multiple constraints, with the class of functions $g$ specified as the subset of $\boldsymbol{Z}$ and its penalty $\gamma$ as the size of selected nodes in $g$. 
 Here, the minimization can be solved using a black-box stochastic optimization such as `Adam'  \citep{kingma2014adam}. Denote the estimated matrix as $\widehat{\boldsymbol{B}}$, based on which we can obtain the estimated causal graph as $\widehat{\mathcal{G}}_{\boldsymbol{V}}$ consisting of nodes $\widehat{g}(\boldsymbol{Z})$. 
Finally, we name this proposed algorithm as necessary and sufficient causal structural learning (NSCSL). The computational complexity of NSCSL is provided in \Cref{app:complex}. We next establish the consistency of estimated causal graphs below.

\begin{theorem}\label{theoremA}
    Assume Model \eqref{lsem_x} holds with independent Gaussian error and equal variance. Suppose the topological ordering of the true bounded matrix $\boldsymbol{B}$ is consistently estimated. Then the estimated matrix $\widehat{\boldsymbol{B}}$ minimizing the loss in \eqref{loss_new} converges to $\boldsymbol{B}$ with the probability going to 1 as $n\to\infty$.
\end{theorem} 

The proof and detailed conditions for \Cref{theoremA} are provided in \Cref{asec:pf}, which align with those commonly imposed in causal structural learning \citep[e.g.,][]{shi2021testing}. Our proof follows similar strategies but accounts for the extra penalty term from causal effects. Notice that the explicit forms of causal effects under LSEM are linear combinations of elements of $\boldsymbol{B}$. 
This implies our new regulation can similarly vanish away as $n$ goes to infinity.   


 \vspace{-0.25cm}
\section{Experiments}\label{simu} 
\vspace{-0.15cm}

\textbf{Experiment design.} Scenarios are generated as follows. We consider the dimension of variables in the graph as $p=5$ in Scenarios 1 to 3 (S1 to S3), $p=20$ in Scenario 4 (S4), and $p=50$ in Scenario 5 (S5), to examine the scalability of NSCSL. 
For each scenario, the DAG that characterizes the causal relationship among variables $\boldsymbol{O} = (\boldsymbol{Z}, Y)$ is generated from the Erd\H os-Re\'nyi (ER) model with an expected degree as 2 for S1 to S3 and degree as 5 for S4 to S5. We also generate the graph from the scale-free (SF) model for S5 with the degree of 5 to examine the robustness of the proposed method under diverse synthetic graphs. Each edge is assigned positive weights. 
We set the last variable as the outcome of interest $Y$, and generate the data based on LSEM by
$\boldsymbol{Z}=\boldsymbol{B}^\top \boldsymbol{Z}+\boldsymbol{\epsilon}$,  where $\boldsymbol{\epsilon} $ is a random vector of jointly independent binary variables with equal probability taking value one or zero. Thus, the outcome is nonnegative and discrete. In addition, we also include a nonlinear structural equation model for S4 where $O_i := \psi_i\{\textsc{PA}_{O_i} (\mathcal{G})\}+ e_{O_i}$ and $\psi_i(x)=\nint{2\log(x+1)}$ where $\nint{x}$ rounds to nearest integer for $x$. 
In S1, the true causal graph contains one spurious node (indexed by 0) and three non-spurious nodes (indexed by 1, 2, and 3), as shown in sub-figures (a) of \Cref{fig_scen_res}, with the associated true NSCG in sub-figures (b) of \Cref{fig_scen_res}. Moreover, we design a balanced setting with half spurious variables and half non-spurious variables in S2, as depicted in \Cref{fig_scen_res3}, and a baseline setting without any spurious variables in S3, shown in \Cref{fig_scen_res00}. Finally, S4 contains 2 non-spurious variables with 17 spurious variables in \Cref{fig_scen_res5}. The experiments are conducted on a Google Cloud Platform virtual machine with 8 processor cores and 32GB memory. 

 \begin{table}[!t]
\centering
\caption{Comparison results 
across S1 to S3 under different sample sizes ($n$). Methods are evaluated by FDR, TPR, and SHD, with the standard error (SE) reported for each metric, over 50 replications.}
\label{tab:comparison}
\centering
  \scalebox{0.75}{
\begin{tabular}{llllllll}
\toprule
& & \multicolumn{2}{l}{FDR$\pm$SE} & \multicolumn{2}{l}{TPR$\pm$SE} & \multicolumn{2}{l}{SHD$\pm$SE} \\
Scenario & Method & $n_1$ (small) & $n_2$ (large) & $n_1$ (small) & $n_2$ (large)& $n_1$ (small) & $n_2$ (large) \\ \midrule
S1 & NSCSL-TE & 0.09$\pm$0.03& 0.02$\pm$0.01 & 0.95$\pm$0.02 & 1.00$\pm$0.00 & 0.64$\pm$0.20& 0.14$\pm$0.06  \\
$p=5$& NSCSL-DE & 0.08$\pm$0.03 & 0.02$\pm$0.01 & 0.95$\pm$0.03 & 1.00$\pm$0.00 & 0.72$\pm$0.20 & 0.14$\pm$0.06\\
$n_1=30$& NOTEARS & 0.39$\pm$0.01 & 0.34$\pm$0.00 & 0.96$\pm$0.02 & 1.00$\pm$0.00 & 2.56$\pm$0.14 & 2.02$\pm$0.01\\
$n_2=100$& PC & 0.53$\pm$0.02 & 0.53$\pm$0.01& 0.47$\pm$0.05 & 0.41$\pm$0.01 & 3.12$\pm$0.11 & 3.20$\pm$0.07\\
ER Model& LiNGAM & 0.31$\pm$0.02 & 0.33$\pm$0.00 & 0.78$\pm$0.02 & 0.98$\pm$0.01 & 2.30$\pm$0.10 & 2.00$\pm$0.00\\\midrule
S2 & NSCSL-TE & 0.10$\pm$0.03& 0.02$\pm$0.01 & 1.00$\pm$0.00 & 0.99$\pm$0.01  & 0.38$\pm$0.14& 0.12$\pm$0.06  \\
$p=5$& NSCSL-DE & 0.12$\pm$0.04 & 0.01$\pm$0.01 & 0.67$\pm$0.04 & 0.50$\pm$0.00 & 1.00$\pm$0.12 & 1.02$\pm$0.01\\
$n_1=30$& NOTEARS & 0.63$\pm$0.01 & 0.60$\pm$0.00 & 1.00$\pm$0.00 & 1.00$\pm$0.00 & 3.46$\pm$0.13 & 3.02$\pm$0.01\\
$n_2=100$& PC & 0.73$\pm$0.02 & 0.79$\pm$0.00& 0.50$\pm$0.00 & 0.50$\pm$0.00 & 3.62$\pm$0.15 & 3.88$\pm$0.03\\
ER Model& LiNGAM & 0.62$\pm$0.01 & 0.60$\pm$0.00 & 0.98$\pm$0.02 & 1.00$\pm$0.00 & 3.18$\pm$0.09 & 3.00$\pm$0.00\\\midrule
S3 & NSCSL-TE & 0.10$\pm$0.02& 0.01$\pm$0.00 & 0.94$\pm$0.02 & 1.00$\pm$0.00  & 0.84$\pm$0.16& 0.06$\pm$0.02  \\
$p=5$& NSCSL-DE & 0.08$\pm$0.02 & 0.01$\pm$0.00 & 0.84$\pm$0.03 & 0.81$\pm$0.00 & 1.30$\pm$0.14 & 1.00$\pm$0.00\\
$n_1=30$& NOTEARS & 0.11$\pm$0.02 & 0.01$\pm$0.00 & 0.96$\pm$0.02 & 1.00$\pm$0.00 & 0.76$\pm$0.16 & 0.06$\pm$0.02\\
$n_2=100$& PC & 0.00$\pm$0.00 & 0.00$\pm$0.00& 0.66$\pm$0.02 & 0.80$\pm$0.00 & 1.68$\pm$0.12 & 1.00$\pm$0.00\\
ER Model& LiNGAM & 0.03$\pm$0.01 & 0.00$\pm$0.00 & 0.98$\pm$0.01 & 1.00$\pm$0.00 & 0.24$\pm$0.09 & 0.02$\pm$0.01\\
\bottomrule
\end{tabular}}
\end{table}

 \begin{table}[!t]
\centering
\caption{Comparison studies 
under S4 to S5 under different sample sizes ($n$) and dimensions ($p$) with the Erd\H os-Re\'nyi (ER) model. Methods are evaluated by FDR, TPR, SHD, and runtime (seconds), with standard errors (SE) reported for each metric, over 50 replications.}
\label{tab:comparison1}
\centering
  \scalebox{0.7}{
\begin{tabular}{llllllllll}
\toprule
& & \multicolumn{2}{l}{FDR$\pm$SE} & \multicolumn{2}{l}{TPR$\pm$SE} & \multicolumn{2}{l}{SHD$\pm$SE}& \multicolumn{2}{l}{Time$\pm$SE} \\
Scenario & Method & $n_1$ (small) & $n_2$ (large) & $n_1$ (small) & $n_2$ (large)& $n_1$ (small) & $n_2$ (large) & $n_1$ (small) & $n_2$ (large) \\ \midrule
S4 & NSCSL-TE & 0.11$\pm$0.02& 0.00$\pm$0.01 & 1.00$\pm$0.00 & 1.00$\pm$0.00  & 0.86$\pm$0.24& 0.04$\pm$0.01 & 10.8$\pm$0.3 & 56.2$\pm$1.1 \\

$p=20$& NSCSL-DE & 0.11$\pm$0.02 & 0.00$\pm$0.01 & 0.69$\pm$0.01 & 0.67$\pm$0.00 & 1.34$\pm$0.08 & 1.00$\pm$0.01& 12.4$\pm$0.8 & 54.5$\pm$1.3 \\

$n_1=100$& NOTEARS & 0.93$\pm$0.00 & 0.92$\pm$0.00 & 1.00$\pm$0.00 & 1.00$\pm$0.00 & 40.80$\pm$0.20 & 36.40$\pm$0.04& 22.9$\pm$6.4 & 69.8$\pm$8.7 \\

$n_2=1000$& PC & 0.92$\pm$0.00 & 0.95$\pm$0.00& 0.51$\pm$0.02 & 0.67$\pm$0.00 & 19.08$\pm$0.17 & 33.34$\pm$0.03& 6.9$\pm$0.5 & 16.3$\pm$0.8 \\

ER Model& LiNGAM & 0.92$\pm$0.00 & 0.93$\pm$0.00& 0.99$\pm$0.01 & 1.00$\pm$0.00 & 33.00$\pm$0.20 & 37.10$\pm$0.02& 8.1$\pm$0.5 & 24.6$\pm$0.6 \\

Degree=5& DAGGNN & 0.93$\pm$0.00 & 0.93$\pm$0.00& 0.97$\pm$0.01 & 0.97$\pm$0.00 & 41.34$\pm$0.19 & 40.10$\pm$0.06& $28.3^2\pm$4.3 & $39.1^2\pm$7.2 \\

& GSGES & 0.98$\pm$0.01 & 0.98$\pm$0.00& 0.22$\pm$0.02 & 0.20$\pm$0.01 & 43.20$\pm$0.26 & 45.70$\pm$0.34& $14.9^2\pm$7.5 & $26.3^2\pm$9.1\\

& FCI & 0.98$\pm$0.01 & 0.99$\pm$0.00& 0.10$\pm$0.02 & 0.06$\pm$0.01 & 22.70$\pm$0.17 & 32.90$\pm$0.02& 6.6$\pm$0.3 & 12.7$\pm$0.2 \\

& CAM & 0.93$\pm$0.00 & 0.94$\pm$0.00&0.63$\pm$0.02 & 1.00$\pm$0.00 & 29.80$\pm$0.38 & 40.30$\pm$0.07& $19.6^2\pm$18.3 & $25.6^2\pm$23.2 \\
\midrule

S5 & NSCSL-TE & 0.03$\pm$0.01& 0.02$\pm$0.01 & 0.86$\pm$0.03 & 0.93$\pm$0.01  & 2.18$\pm$0.13& 1.58$\pm$0.07 &110.1$\pm$3.9 & $21.1^2\pm$12.1\\

$p=50$& NSCSL-DE &0.02$\pm$0.02& 0.01$\pm$0.01 & 0.29$\pm$0.02 & 0.28$\pm$0.01 & 10.08$\pm$0.21 & 9.76$\pm$0.13&119.0$\pm$5.5 & $23.5^2\pm$10.9\\

$n_1=1000$& NOTEARS & 0.86$\pm$0.04 & 0.85$\pm$0.01 & 0.93$\pm$0.03 & 0.92$\pm$0.01 & 79.20$\pm$1.40 & 77.12$\pm$0.53& 128.3$\pm$8.2 & $26.9^2\pm$15.1\\

$n_2=3000$& PC & 0.96$\pm$0.03 & 0.97$\pm$0.02& 0.07$\pm$0.02 & 0.06$\pm$0.01 & 82.12$\pm$1.21 & 88.28$\pm$1.63& 20.9$\pm$0.5 & 35.9$\pm$3.2\\

ER Model& LiNGAM & 0.86$\pm$0.02 & 0.86$\pm$0.01& 0.97$\pm$0.02 & 0.99$\pm$0.01 & 86.12$\pm$1.20 & 85.70$\pm$0.84& 43.1$\pm$6.3 & 145.3$\pm$7.9\\

Degree=5& DAGGNN & 0.87$\pm$0.02 & 0.88$\pm$0.01& 0.93$\pm$0.02 & 0.94$\pm$0.01 & 87.50$\pm$1.10 & 85.62$\pm$0.96& $49.3^2\pm$7.5 & $81.1^2\pm$87.6 \\

& GSGES & 0.89$\pm$0.03 & 0.93$\pm$0.01& 0.19$\pm$0.03 & 0.12$\pm$0.01 & 93.54$\pm$1.46 & 95.70$\pm$0.79& $31.2^2\pm$10.1 & $45.3^2\pm$18.1\\

& FCI & 0.96$\pm$0.02 & 0.97$\pm$0.01& 0.08$\pm$0.01 & 0.07$\pm$0.01 & 84.00$\pm$0.80 & 88.50$\pm$0.60 & 14.3$\pm$0.8 & 17.7$\pm$0.5 \\

& CAM & 0.93$\pm$0.04 & 0.95$\pm$0.02& 0.66$\pm$0.03 & 0.67$\pm$0.02 & 126.00$\pm$3.64 & 127.80$\pm$2.06& $28.4^2\pm$31.7 & $75.6^2\pm$63.9 \\




\bottomrule
\end{tabular}} 
\end{table}

\textbf{Methods and benchmark specification.} We apply the proposed NSCSL based on TE and DE as the criteria of necessity and sufficiency, respectively, to capture the marginal and conditional causal effect of the confounders. Note that we consider fully identifiable models in the experiments so that it is meaningful to evaluate causal effects from the estimated graph. The underlying causal structure learning algorithm is set to NOTEARS \citep{zheng2018dags}. 
We also compare the proposed method against other methods,  
including PC \citep{spirtes2000constructing} and LiNGAM \citep{shimizu2006linear} for S1 to S5; DAG-GNN \citep{yu2019dag}, GES with generalized score \citep[GSGES, ][]{huang2018generalized}, fast causal inference \citep[FCI, ][]{spirtes2013causal}, and the causal additive model \citep[CAM, ][]{buhlmann2014cam}, for all high-dimensional settings in S4 and S5. 
Here, we use a graph threshold of 0.3 (commonly used in the literature \citep{zheng2018dags,yu2019dag,zhu2019causal,cai2020anoce}) to prune the noise edges for a fair comparison. The training details are provided in \Cref{tab:train}. The true and estimated graphs with the associated matrix under different approaches are illustrated in Figs. \ref{fig_scen_res2} to \ref{fig_scen_res8} and Figs. \ref{fig_scen_res} to \ref{fig_scen_res7} in \Cref{asec:simu} for S1 to S4. 
The comparison results across different sample sizes ($n$) are presented in \Cref{tab:comparison} for S1 to S3, in \Cref{tab:comparison1} for S4 to S5 with linear ER model, in \Cref{tab:comparison2} for S4 with nonlinear ER model, and in \Cref{tab:comparison3} for S5 with linear SF model. All the results are evaluated by false discovery rate (FDR), true positive rate (TPR), and the structural Hamming distance (SHD) to the true causal graph, with their standard errors, over 50 replications. 
The average running time of these methods is also reported in \Cref{tab:comparison} to \Cref{tab:comparison3} to reflect the computational complexities. In addition, the sensitivity analyses concerning all hyperparameters in \Cref{tab:train} are conducted using S4, as presented in Fig. 2.

 \textbf{Results and conclusion.} NSCSL performs the best 
 in discovering NSCG in S1, S2, S4, and S5, and shows comparably best results in S3 (a setting without any spurious  variables). To be specific, the benchmark methods for causal structural learning aim to reveal causal relationships in the whole graph (i.e., sub-figures (a) in Figs. \ref{fig_scen_res2} to \ref{fig_scen_res8}), which contains spurious effects on the target outcome $Y$. The proposed algorithm overcomes this drawback by purely identifying the true important causal relationships (i.e., sub-figures (b) in Figs. \ref{fig_scen_res2} to \ref{fig_scen_res8}). By comparing the sub-figures (c) and (d) in Figs. \ref{fig_scen_res2} to \ref{fig_scen_res8} as well as \Cref{tab:comparison} to \Cref{tab:comparison3}, NSCSL-TE detects all necessary and sufficient causal paths towards the outcome, while NSCSL-DE only extracts direct causal relationships between the features and the outcome, resulting in a slightly lower TPR and slightly higher SHD than NSCSL-TE. Furthermore, from \Cref{tab:comparison} to \Cref{tab:comparison3}, the results under the proposed algorithm approach the truth more closely as the sample size increases in all scenarios, regardless of the underlying graph models and data-generating process. In contrast, the benchmark methods all fail to discover NSCG in the high-dimensional setting, exhibiting extremely high FDRs and SHDs. In addition, \Cref{tab:comparison1} to \Cref{tab:comparison3} reveal that NSCSL is as fast as the quickest benchmarks such as PC, LiNGAM, and FCI, and significantly faster than others like DAGGNN, GSGES, and CAM. Our method's integration of treatment effects into the optimization adds efficiency and restricts the searching space, making it practical and even beating NOTEARS in computation. The sensitivity analyses in Fig. 2 indicate that our method remains robust to these parameters 
 set within a reasonable range.

\begin{table}[!t]
\begin{minipage}{0.65\textwidth} 
\centering
\caption{Comparison studies under the nonlinear structural equation model for S4 
over 50 replications.}
\label{tab:comparison2}
\centering
  \scalebox{0.75}{
\begin{tabular}{llllll}
\toprule
Scenario & Method & FDR$\pm$SE & TPR$\pm$SE & SHD$\pm$SE & Time$\pm$SE \\ \midrule
S4 & NSCSL-TE  & 0.03$\pm$0.01   & 0.83$\pm$0.01  & 0.60$\pm$0.02  & 55.8$\pm$0.3 \\

$p=20$& NSCSL-DE  & 0.03$\pm$0.01   & 0.50$\pm$0.01 &  1.60$\pm$0.02 & 56.8$\pm$0.2 \\

$n=1000$& NOTEARS  & 0.91$\pm$0.01  & 0.83$\pm$0.01   & 35.90$\pm$0.04& 56.5$\pm$0.7 \\

ER Model& PC               & 0.99$\pm$0.01 & 0.12$\pm$0.01  & 44.12$\pm$0.03 & 15.7$\pm$0.2 \\

Degree=5& LiNGAM             & 0.93$\pm$0.01 & 0.70$\pm$0.00  & 37.30$\pm$0.03  & 11.6$\pm$0.1 \\

& DAGGNN                    & 0.94$\pm$0.01 & 0.86$\pm$0.01  & 34.80$\pm$0.06 & $41.3^2\pm$9.8 \\

& GSGES                       & 0.98$\pm$0.01 & 0.23$\pm$0.01  & 52.40$\pm$0.38 & $22.1^2\pm$7.5\\

& FCI                             & 0.97$\pm$0.01 & 0.13$\pm$0.01  & 33.80$\pm$0.06  & 12.6$\pm$0.3 \\

& CAM                          & 0.95$\pm$0.01 & 1.00$\pm$0.01  & 31.60$\pm$0.08  & $27.9^2\pm$33.7 \\







\bottomrule
\end{tabular}}

\bigskip


\centering
\caption{Comparison studies under the scale-free (SF) model for S5 
over 50 replications.}
\label{tab:comparison3}
\centering
  \scalebox{0.75}{
\begin{tabular}{llllll}
\toprule
Scenario & Method & FDR$\pm$SE & TPR$\pm$SE & SHD$\pm$SE & Time$\pm$SE \\ \midrule
S5 & NSCSL-TE & 0.02$\pm$0.02 & 0.78$\pm$0.03   & 5.08$\pm$0.11 &135.7$\pm$5.6 \\

$p=50$& NSCSL-DE &0.02$\pm$0.02 & 0.51$\pm$0.02  & 17.20$\pm$0.35  &147.0$\pm$6.3  \\

$n=1000$& NOTEARS & 0.88$\pm$0.04  & 0.75$\pm$0.03   & 123.10$\pm$1.50  & 160.5$\pm$8.1 \\

SF Model& PC & 0.97$\pm$0.03  & 0.06$\pm$0.02  & 79.34$\pm$1.17 & 32.1$\pm$0.8  \\

Degree=5& LiNGAM & 0.91$\pm$0.02  & 0.98$\pm$0.01   & 212.00$\pm$5.12  & 117.7$\pm$6.3 \\

& DAGGNN & 0.92$\pm$0.02  & 0.85$\pm$0.02   & 203.50$\pm$7.80  & $57.3^2\pm$13.6  \\

& GSGES & 0.96$\pm$0.03 & 0.10$\pm$0.03  & 98.34$\pm$2.15  & $35.9^2\pm$13.5  \\

& FCI & 0.97$\pm$0.02 & 0.07$\pm$0.01 & 81.70$\pm$1.40 & 15.7$\pm$1.1  \\

& CAM & 0.98$\pm$0.04 & 0.24$\pm$0.03   & 218.00$\pm$8.15  & $37.1^2\pm$53.6  \\
\bottomrule
\end{tabular}}

\bigskip


\centering
\caption{Real data results for the single-cell data by \citet{sachs2005causal}, 
evaluated by total edges, correct edges, and SHD, based on the true NSCG with respect to the protein Akt.}
\label{tab:real1}
\centering
  \scalebox{0.68}{
\begin{tabular}{llllllllll}
\toprule
Method & NSCSL &NOTEARS & PC  & LiNGAM & DAGGNN & GSGES   & FCI  & 
CAM\\
\midrule
Total Edges &8&20&25&6&33 &28&24&7\\
Correct Edges &4&2&2&1&4 &2&2&1\\
SHD &8  &21&28&11&30&32&27 &10\\
\bottomrule
\end{tabular}} 
\end{minipage}\hfill
\begin{minipage}[!t]{0.3\textwidth}
\centering
\includegraphics[width=1\textwidth]{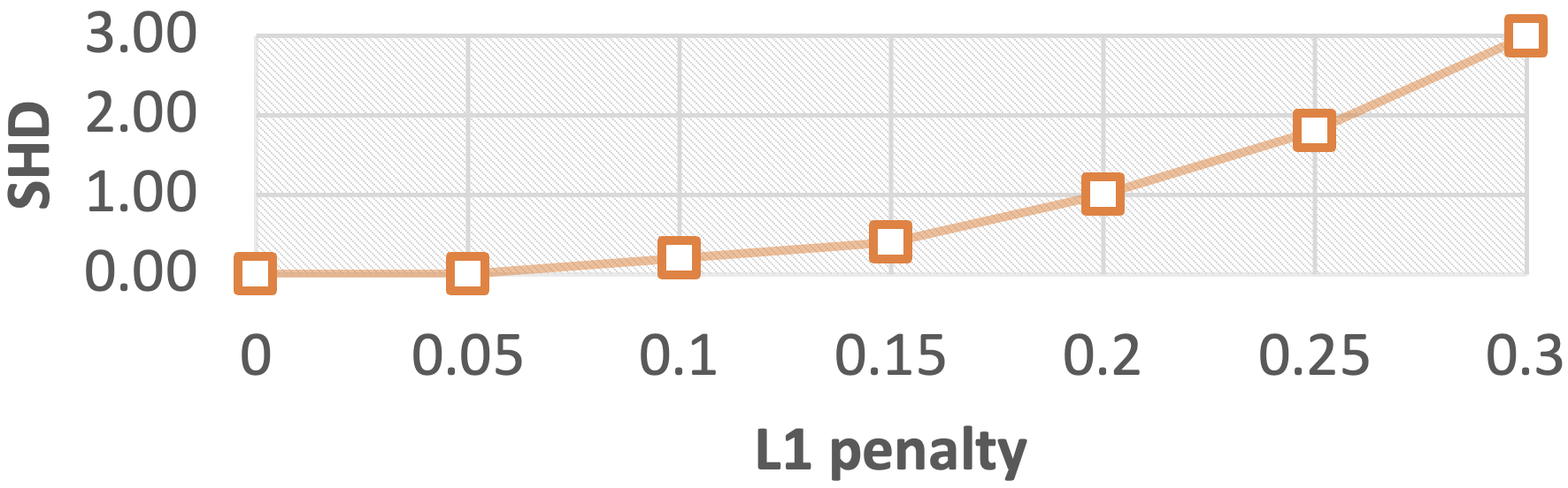}

\includegraphics[width=1\textwidth]{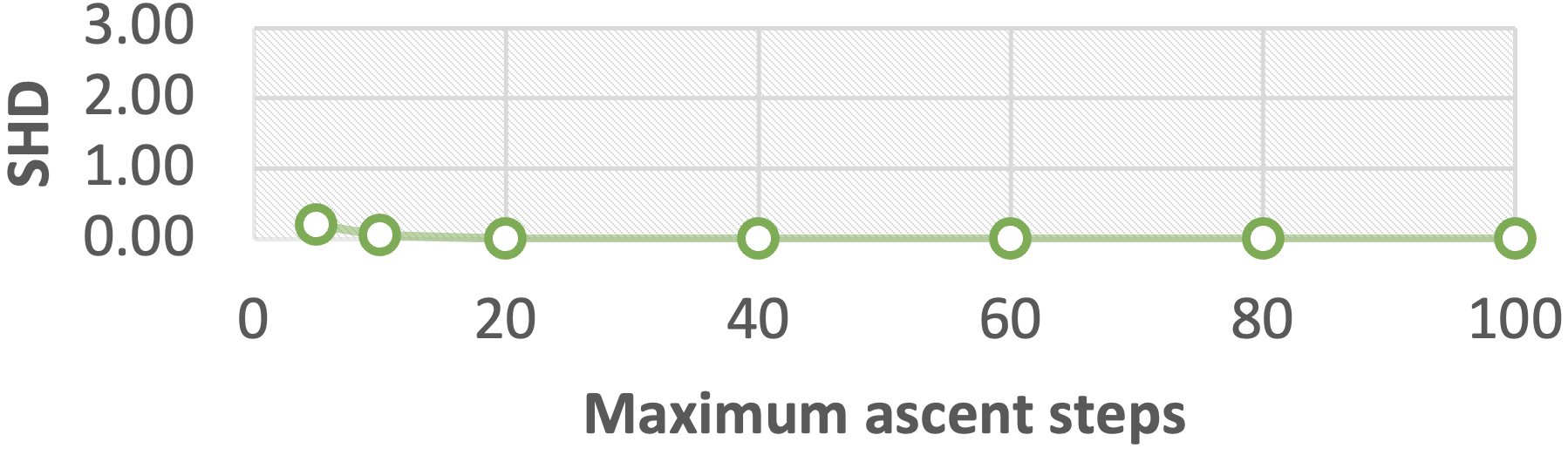}

\includegraphics[width=1\textwidth]{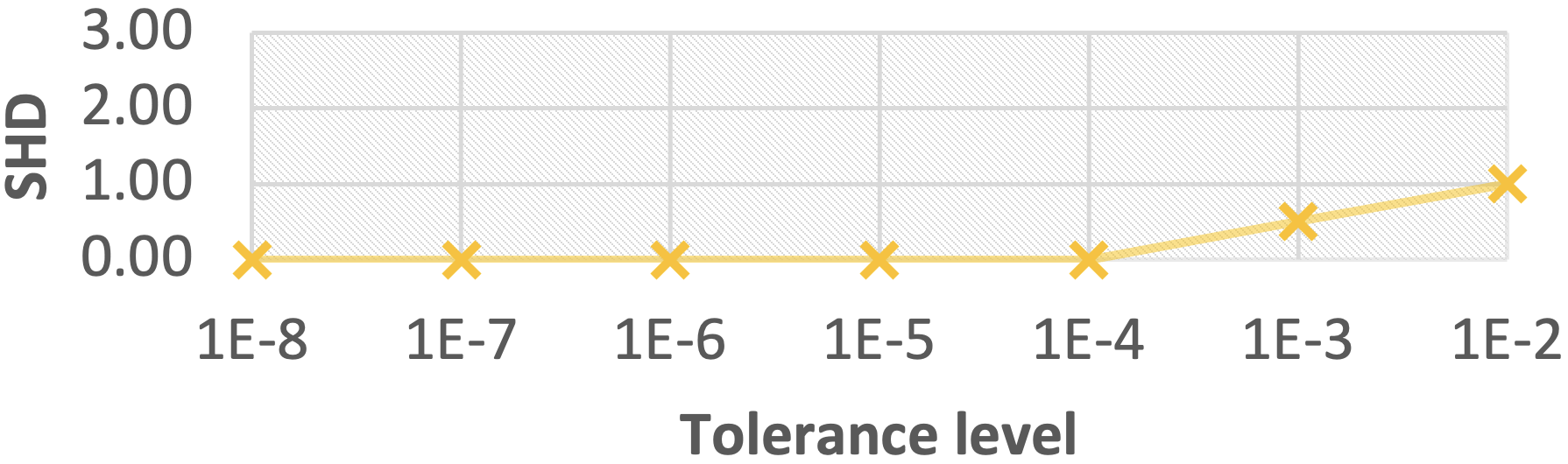}

\includegraphics[width=1\textwidth]{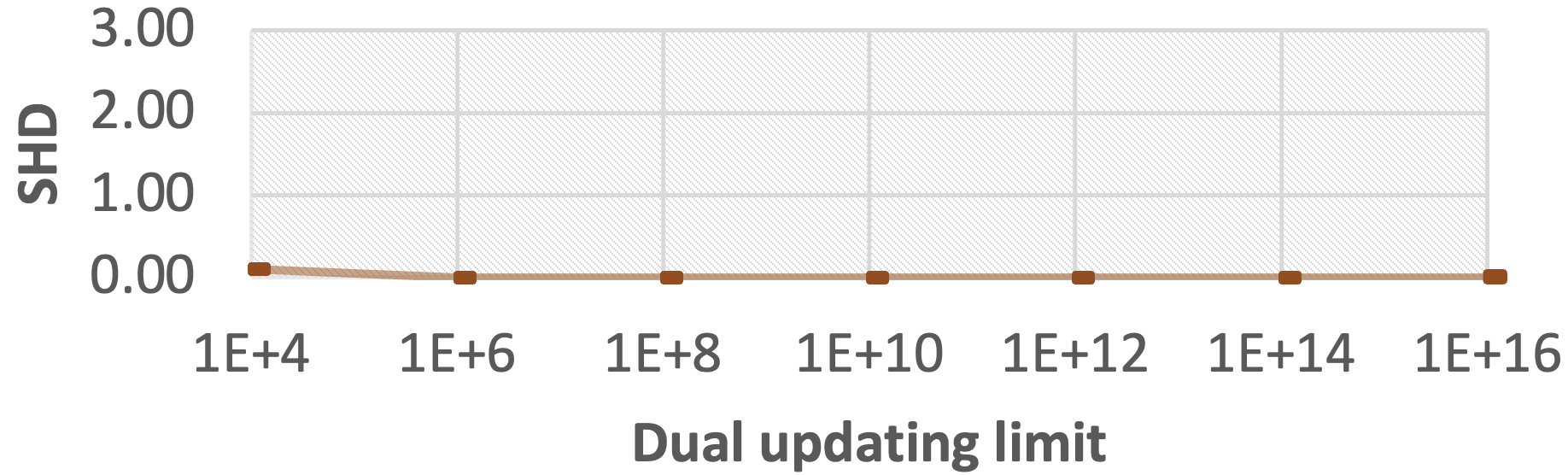}

Figure 2: Sensitivity analyses.

\bigskip

\centering

\includegraphics[width=1\textwidth]{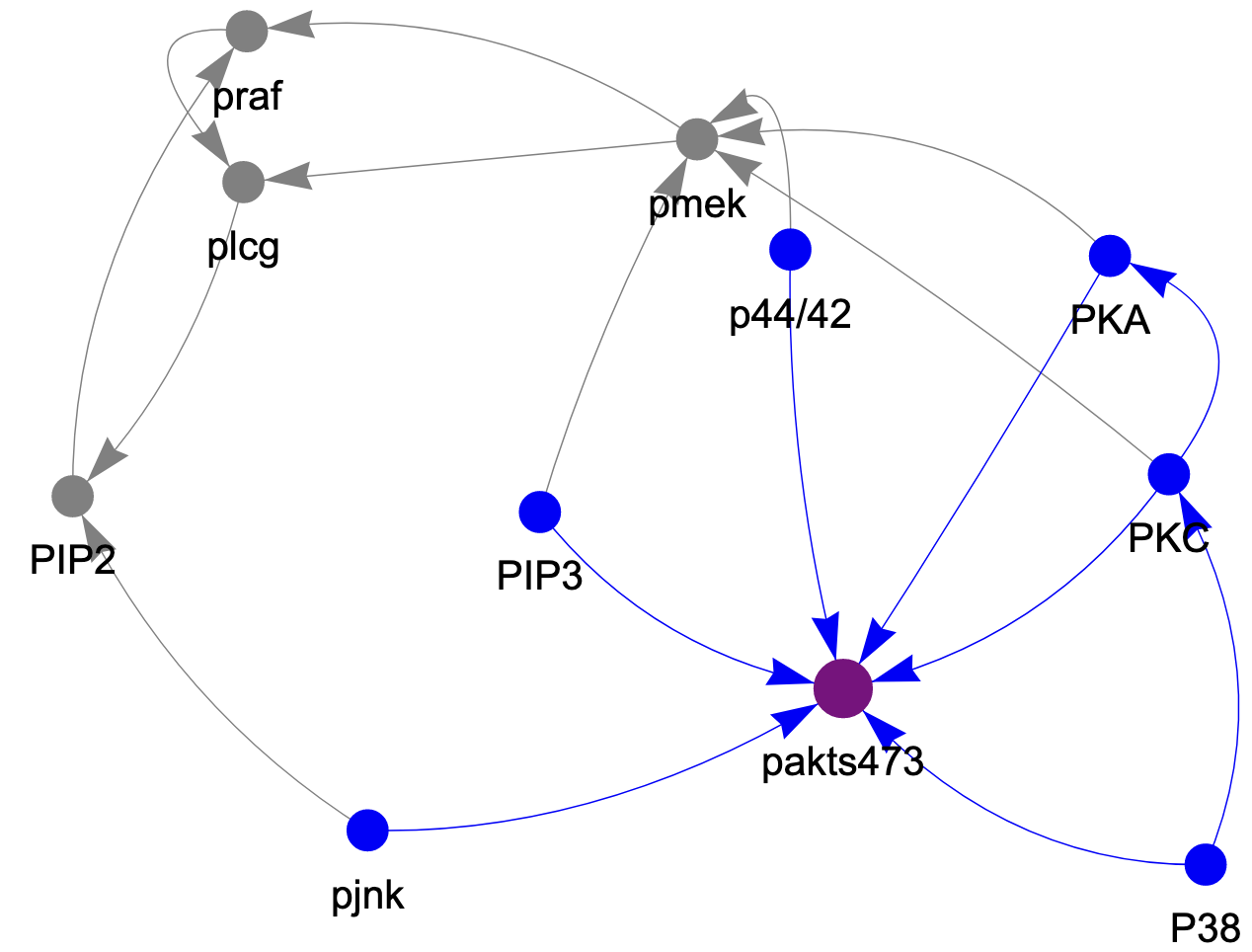}

Figure 3: The causal signaling network in \citet{sachs2005causal}, where the blue-colored sub-graph is the true NSCG for protein Akt (purple).

\label{fig:image}
\end{minipage}
\end{table}

 \begin{table}[!t] 
\centering
\caption{Real data results for the yeast gene data, 
 evaluated by total edges, the identified parents/ancestors of the variant YER124C, and the edges towards YER124C.}
\label{tab:real2}
\centering
  \scalebox{0.75}{
\begin{tabular}{llllllllll}
\toprule
Method & NSCSL  &NOTEARS & PC  & LiNGAM & DAGGNN & GSGES   & FCI  & CAM\\
\midrule
Total Edges & 11              &25              &22        &15     &35     &27       &22     &33\\
\# Parents/Ancestors of YER124C& 8  &8  &6   &4     &8        &7          &5       &7\\ 
\# Edges towards YER124C &11    &9    &8         &6     &10     &9           &8     &10\\
\bottomrule
\end{tabular}} 
\end{table}

\textbf{Real data analysis - \citet{sachs2005causal}.} We conduct real data analysis using the benchmark data from \citet{sachs2005causal}. To validate our method's capacity to find the NSCG and align with \Cref{defi_necessity}, we designated the protein Akt as the target outcome. This designation ensures that NSCG exists (see Fig. 3) and that finding an NSCG is meaningful. Our method (NSCSL-TE) and seven baseline methods were applied and evaluated against the true NSCG associated with the protein Akt. \Cref{tab:real1} shows that our method achieves the best performance in finding the NSCG concerning the protein Akt.

\textbf{Real data analysis - \citet{brem2005landscape}.} 
Furthermore, we apply NSCSL to gene expression traits in yeast \citep{brem2005landscape} using a dataset of 104 yeast segregants with diverse genotypes. The goal is to identify genes, known as quantitative trait loci (QTLs), influencing the expression level (nonnegative) of the genetic variant YER124C, a daughter cell-specific protein involved in cell wall metabolism. Following a similar approach as in \citet{chakrabortty2018inference}, we include 492 QTLs by filtering out genes with missing or low variability in expression levels. The total sample size is 262. Given the high-dimensional QTLs, constructing an NSCG with only causally relevant variables for the outcome of interest is essential.   
We apply NSCSL-TE and compare it with 
all baseline methods. The summarized results in \Cref{tab:real2} highlight our method's ability to identify relevant genetic influencers for the variant YER124C without contamination by irrelevant genes. The estimated causal graph and causal effects as well as more detailed analyses are provided in \Cref{more_yeast}. 
These observations align with findings from our simulation studies, further supporting NSCSL's superiority in revealing important causal features.

\vspace{-0.3cm}
\section{Limitations and Future Research}  
\vspace{-0.2cm}

In this work, we introduced NSCSL that leverages causal effects/POCs to systematically assess feature importance while learning a causal graph. By identifying a subgraph closely related to the outcome, our method filters irrelevant variables, presenting a significant advancement in the field. Extensive empirical evaluations on simulated and real-world data underscore NSCSL's superior performance over existing algorithms, including important findings on yeast genes and the protein signaling network. However, this promising advancement is not without limitations. First, NSCSL, like most existing causal structural learning methods, assumes no unmeasured confounders (A2) and the causal Markov condition. These assumptions may not hold in practice, leading to biased causal effect estimates and potential errors in the causal graph. Second, NSCSL employs absolute causal effects as a substitute for POCs to facilitate estimation in high-dimensional settings. Although theoretically consistent under certain conditions, examining the differences between these two methods in general feature and outcome spaces is an area for future research.

\begin{ack}
Research reported in this publication was supported in part by the Amazon Web Services (AWS) Cloud Research program, the Office of Naval Research under grant N00014-23-1-2590, and the National Science Foundation under grant No. 2231174 and No. 2310831. The authors thank the area chair and the reviewers for their constructive comments which have led to a significant improvement of the earlier version of this article.
\end{ack}

\clearpage

\setcounter{tocdepth}{-1}

\appendix

\begin{appendices}
  \renewcommand{\contentsname}{Table of Contents}
  \tableofcontents
  \addtocontents{toc}{\protect\setcounter{tocdepth}{3}}






\newpage

 \counterwithin{figure}{section}
\counterwithin{table}{section}
\counterwithin{equation}{section}  
\section{Notations and Abbreviations}\label{app_not}

\begin{table}[h]
\centering
\caption{The table of notations and abbreviation.}
\begin{tabular}{lp{8cm}}
\toprule
\textbf{Concept} & \textbf{Description} \\
\midrule
Graph $\mathcal{G}$ & A graph with a node set $\boldsymbol{X}$ and edge set $\boldsymbol{D}_{\boldsymbol{X}}$ \\
\midrule
Node $X_i$ & A member of the node set $\boldsymbol{X}$ \\
\midrule
Parent of $X_j$ & A node $X_i$ such that there is a directed edge from $X_i$ to $X_j$ (i.e., $X_i$ is a direct cause of $X_j$) \\
\midrule
Ancestor of $X_j$ & A node $X_k$ such that there's a directed path from $X_k$ to $X_j$ regulated by at least one additional node $X_i$ for $i \neq k$ and $i \neq j$ (i.e., $X_k$ is an indirect cause of $X_j$) \\
\midrule
$\textsc{PA}_{X_j} (\mathcal{G})$ & The set of all parents/ancestors of node $X_j$ in $\mathcal{G}$ \\
\midrule
DAG & A directed acyclic graph; a directed graph $\mathcal{G}$ that does not contain directed cycles \\
\midrule
SCM & The structural causal model characterizes the causal relationship among $|\boldsymbol{X}|=d$ nodes via a DAG $\mathcal{G}$ and noises $\boldsymbol{e}_{\boldsymbol{X}} = [e_{X_1},\cdots,e_{X_d}]^\top$ such that $X_i := h_i\{\textsc{PA}_{X_i} (\mathcal{G}), e_{X_i}\}$ for some unknown $h_i$ and $i=1,\cdots,d$ \\
\midrule
$Y(Z_i=z_i)$ & Potential outcome after setting individual variable $Z_i$ to $z_i$ \\
\midrule
$\mathcal{G}_{\boldsymbol{O}}$ & DAG characterizing the causal relationship among $\boldsymbol{O}$ \\
\midrule
$ \mathbb{P}_{\mathcal{G}}$ & The mass/density function for an SCM with its DAG ${\mathcal{G}}$ \\
\midrule
$\boldsymbol{Z}_{-i}\equiv \boldsymbol{Z}\setminus Z_i$ &  The complementary variable set of $Z_i$\\
\bottomrule
\end{tabular}
 
\end{table} 

\section{NSCSL for Nonlinear Models: The POC-based Learning Algorithm} \label{asec:algo}


In this section, we outline a method for learning the NSCG for a nonlinear model. Given the lack of an explicit form for both causal effects and POCs in such models, we employ an iterative learning method based on \Cref{thm1} to capture complex nonlinear relationships among variables. The method begins with a pre-screening process, identifying necessary and sufficient features from $\boldsymbol{Z}$ with high causation scores. Following this, the causal graph is estimated among the selected nodes and $Y$, approximating $\mathcal{G}_{\boldsymbol{V}}$ in the nonlinear structural equation model. This iterative approach is applicable for general SCMs and the process is repeated until convergence is achieved. \Cref{asec:est_POC} introduces the nonlinear structural equation model and describes the estimation of POC in such a model. The main algorithm based on POCs is presented in \Cref{asec:algo_POCs}.
\subsection{Nonlinear Structural Equation Model and  Estimation of POC} \label{asec:est_POC}

 \subsubsection{Nonlinear Structural Equation Model} 
 
 While the linear structural equation model has good properties such as easy implementation and nice interpretation, it cannot capture complex nonlinear causal relationships. To address this, we consider the non-linear additive form following \cite{peters2014causal,zheng2020learning,rolland2022score}. Specifically, for variables $\bm{D}=\{g(\boldsymbol{Z}),Y\}$ as a $d+1$-dimensional vector, we consider replacing the model in  \eqref{lsem_x} as follows,
\begin{equation}\label{NonPara-SEM} 
D_i := \psi_i\{\textsc{PA}_{D_i} (\mathcal{G})\}+ e_{D_i},
\end{equation}
for the $i$-th element/node $D_i$ in $\bm{D}$ with some unknown nonlinear function $\psi_i$ and independent noise $e_{D_i}$, for $i=1,\cdots,d+1$. 
As mentioned in \citep{zheng2020learning,rolland2022score}, this model is identifiable from observational data. We further define a new functional matrix $\bm{B}(\psi) = \bm{B}(\psi_1, \cdots,\psi_{d+1})$ that encodes the unknown causal relationship among variables. The element in the $i$-th row and $j$-th column is defined as:
\begin{equation}\label{NonPara-B}
[\bm{B}(\psi)]_{ij}:=||\partial_j \psi_i||_2,
\end{equation}
 where $||\cdot||_2$ presents the $L_2$ norm. This matrix thus describes the dependency among $\bm{D}$; if $D_i$ does not depend on $D_j$, we have $||\partial_j \psi_i||_2=0$. We can also incorporate background knowledge as done in \cref{sec:algo} to reflect the causal roles in different types of variables. We do this by specifying the last column of $\bm{B}(\psi)$ to be all zeros, so that $h_c{\bm{B}(\psi)}= \sum_{i=1}^{d+1} |[\bm{B}(\psi)]_{i,d+1}| = 0$.
 
 \subsubsection{Estimation of POC}
 We detail the estimation of POC based on \Cref{thm1} and the model \eqref{NonPara-SEM} as follows. Denote the estimator of the conditional probability of the outcome given the $i$-th selected feature $g_i(\boldsymbol{Z})$ as $\widehat{\mathbb{P}}(Y=y|g_i(\boldsymbol{Z}))$. This can be achieved by either parametric models (such as logistic regression for the binary outcome) or non-parametric models (such as random forest or neural network). Then, the estimated marginal POC across $n$ data points is given by
    \begin{equation*}
  \begin{split}
\widehat{\text{M-POC}}(g_i|\{\boldsymbol{o}^{(j)}\})  = \prod_{j=1}^{n} &\left|\widehat{\mathbb{P}}\{Y=y^{(j)}|g_i(\boldsymbol{Z})=g_i(\boldsymbol{z}^{(j)})\} - \widehat{\mathbb{P}}\{Y=y^{(j)}|g_i(\boldsymbol{Z})\not=g_i(\boldsymbol{z}^{(j)})\}\right|,
 \end{split}
 \end{equation*}
   where $g_i(\boldsymbol{Z})$ is the $i$-th dimension of $g(\boldsymbol{Z})$. Similarly, we can estimate the conditional probability of the outcome given all selected feature $g(\boldsymbol{Z})$ as $\widehat{\mathbb{P}}(Y=y|g(\boldsymbol{Z}))$.  Likewise, we have the estimated conditional POC as 
\begin{align*}
 \widehat{\text{C-POC}}(g_i|\{\boldsymbol{o}^{(j)}\}) 
&= \prod_{j=1}^{n} \left|\widehat{\mathbb{P}}\{Y=y^{(j)}|g_i(\boldsymbol{Z})=g_i(\boldsymbol{z}^{(j)}), g_{-i}(\boldsymbol{Z})=g_{-i}(\boldsymbol{z}_{-i}^{(j)})\}\right.\\&\qquad \left.~~~- \widehat{\mathbb{P}}\{Y=y^{(j)}|g_i(\boldsymbol{Z})\not=g_i(\boldsymbol{z}^{(j)}), g_{-i}(\boldsymbol{Z})=g_{-i}(\boldsymbol{z}_{-i}^{(j)})\}\right|,
 \end{align*}
 where $g_{-i}(\cdot) \equiv g(\cdot)\setminus g_i(\cdot) $ is the complement  of $g_i(\cdot)$.

\subsection{Learning Algorithm based on POCs}\label{asec:algo_POCs}

Given the lack of an explicit form for causal quantities in the nonlinear models, we employ an iterative learning method based on \Cref{thm1} to capture complex nonlinear relationships among variables. The method begins with an initialized causal graph and then identifies necessary and sufficient features from $\boldsymbol{Z}$ with high POCs by setting $g$ as a subset selection function. Following this, the causal graph can be updated among the selected nodes and $Y$, approximating $\mathcal{G}_{\boldsymbol{V}}$ in the nonlinear structural equation model based on data $\{\boldsymbol{o}^{(j)}= (\boldsymbol{z}^{(j)}, y^{(j)})\}_{1\leq j\leq n}$. This process is repeated until convergence is achieved. The main algorithm based on POCs is presented as follows.

\subsubsection{Step 1: Nonlinear Causal Structural Learning} 

In the first step, we employ a causal structural learning algorithm for nonlinear models to estimate the functional matrix $\bm{B}(\psi)$ as presented in \eqref{NonPara-B}. This estimation is performed considering the selector $g^{k}$ at the $k$-th iteration, where $g^{1}(\boldsymbol{Z}):=\boldsymbol{Z}$. Aligning with the main text, we adopt the nonparametric acyclicity constraint on $\bm{B}(\psi)$ proposed in \citet{zheng2020learning}, represented as $
 h_n(\bm{B}(\psi))=0.$ The loss function, defined by the augmented Lagrangian for the $k$-th iteration, is then formulated as:
\begin{eqnarray}\label{asec:loss1}
\tilde{L}(\bm{B}(\psi),\theta,\lambda|g^{k},\{\boldsymbol{o}^{(j)}\} 
)=\tilde{f}(\bm{B}(\psi),\theta|g^{k},\{\boldsymbol{o}^{(j)}\})+\lambda \{h_c(\bm{B} (\psi))+h_n(\bm{B}(\psi))\},
\end{eqnarray}
where $\tilde{f}(\bm{B}(\psi),\theta|g^{k},\{\boldsymbol{o}^{(j)}\})$ denotes a nonlinear loss function with parameters $\theta$, and $\lambda$ represents the Lagrange multiplier. The objective in \eqref{asec:loss1} can be solved using existing nonlinear causal structural learning methods (refer to \citet{yu2019dag}, \citet{zhu2019causal}, \citet{zheng2020learning}) given the selector $g^{k}$ and data $\{\boldsymbol{o}^{(j)}= (\boldsymbol{z}^{(j)}, y^{(j)})\}_{1\leq j\leq n}$. The estimated functional matrix resulting from this process is represented as $\widehat{\bm{B}}^k(\widehat{\psi})$.


\subsubsection{Step 2: Causal Relevance Measurement Using POCs}
Following the estimation of the functional matrix $\widehat{\bm{B}}^k(\widehat{\psi})$, we assess the causal relevance of nodes through the probabilities of causation (POCs) to update the selection function $g$. The loss function relating to the selector is defined as follows: 
\begin{eqnarray}\label{asec:loss2}
\tilde{L}^{P}(g,\gamma|\widehat{\bm{B}}^k(\widehat{\psi}),{\boldsymbol{o}^{(j)}})
=- \sum_{i=1}^{d} \widehat{\text{P}}(g_i|{\boldsymbol{o}^{(j)}})
+\gamma |g|,
\end{eqnarray} 
where $|g|$ signifies the number of selected nodes in $g$ with a penalty $\gamma$ for controlling the complexity of the selector. The term $g_i$ refers to the $i$-th dimension of $g$ for $i=1,\cdots,d$. Here, $\widehat{\text{P}}(\cdot)$ can represent either the estimated $\text{M-POC}$ or $\text{C-POC}$ as outlined in \Cref{asec:est_POC}. By optimizing different POCs, we can learn the corresponding best selector $g$ according to the loss function detailed in \eqref{asec:loss2} using the CAUSAL-REP algorithm as proposed in \citet{wang2021desiderata} by considering the selector $g$ as the subset of $\boldsymbol{Z}$. The resulting selector is denoted as $g^{k+1}$.
   


\subsubsection{Step 3: Necessary and Sufficient Causal Structural Learning}
Repeat the optimization process for loss functions in \eqref{asec:loss1} and \eqref{asec:loss2} for $k=1,\cdots,K$ iterations until either the maximum iteration number $K$ is reached, or the change in loss functions in \eqref{asec:loss1} and \eqref{asec:loss2} falls below a predefined tolerance level $\tau$. The resultant estimated matrix is denoted as $\widehat{\bm{B}}(\widehat{\psi})$, from which the estimated causal graph, $\widehat{\mathcal{G}}_{\boldsymbol{V}}$, can be derived.

  \subsection{The Computational Complexity of the NSCSL Algorithm}\label{app:complex}
      The computational complexity of NSCSL comprises two parts: the cost from causal discovery as $G(n,p)$, and the estimation of causal effects/scores $F(n,p)$, where $n$ is the data sample size and $p$ is the number of nodes. In the linear case, our method learns the features and causal graph through single-step optimization in \eqref{loss_new}, with complexity cubic in the number of nodes, $G(n,p) = \mathcal{O}(p^3)$ following \citet{zheng2018dags}. Here, the causal effect computation is linear-time and thus is dominated. In the nonlinear case, according to \Cref{asec:algo_POCs}, the time complexity depends on the base causal discovery method and the number of max iterations $K$, yielding $ \mathcal{O}[K(G(n,p) + F(n,p))]$. Supporting runtime details are provided in \Cref{simu}. 

\subsection{Discussion on Scale Invariance}
      NSCSL is scale-invariant when we appropriately choose the causal discovery base learner and model the treatment effects/POCs. Though NOTEARS lacks scale invariance, our method's flexibility allows for the integration of scale-invariant causal discovery methods like NSCGL with FCI. Additionally, under LSEM, the rescaling will not affect the relative rank of the features based on absolute causal effects. In the nonlinear case, we proposed to use POCs which by their definitions are scale-invariant. 
      
\section{Extension to Markov Equivalence Class}\label{asec:mec}
Our proposed algorithm can also be extended to manage the Markov equivalence class of partial directed acyclic graphs when the causal graph cannot be uniquely identified from the observational studies.
 
 \subsection{Additional Graph Terminology}
 A graph $\mathcal{G}$ that contains directed and/or undirected edges is termed as a partially directed graph. If this graph doesn't contain a directed cycle, it's referred to as a partially directed acyclic graph or PDAG. The DAG $\mathcal{G}$ is generally not identifiable from the distribution of $\bm{X}$ based on conditional independence relationships, as per observational data \citep{pearl2000models}. This is because multiple DAGs can represent the same conditional independence relationships, and these DAGs form a Markov equivalence class (MEC), denoted as $MEC({\mathcal{G}})$. Two DAGs belong to the same MEC if and only if they have the same skeleton and the same $v$-structures \citep{pearl2000models}. A MEC of DAGs can be uniquely symbolized by a completed partially directed acyclic graph (CPDAG) \citep{spirtes2000constructing,chickering2002optimal}, a graph that may contain both directed and undirected edges. A CPDAG adheres to the following: if $X_i\rightarrow X_j$ exists in the CPDAG, then $X_i\rightarrow X_j$ is present in every DAG in the MEC; and if $X_i - X_j$ exists in the CPDAG, then the MEC contains a DAG for which $X_i\rightarrow X_j$ as well as a DAG for which $X_j\rightarrow X_i$.

\subsection{Model Identifiabilities}

In the absence of further assumptions regarding the form of functions and/or noises, the model in \eqref{lsem_x} can only be identified up to MEC following the Markov and faithful assumptions \citep{spirtes2000constructing,peters2014causal}. Below, we explore the conditions for the unique identifiability of the DAG and potential strategies for addressing scenarios involving the MEC.

Initially, we consolidate cases where the DAG is uniquely identifiable. In the context of the LSEM, when the noises $\boldsymbol{\epsilon}$ follow a Gaussian distribution, the resulting model corresponds to the standard linear-Gaussian model class, as investigated in \citet{spirtes2000constructing} and \citet{peters2017elements}. In instances where the noises $\boldsymbol{\epsilon}$ maintain equal variances, according to \citet{peters2014identifiability}, the DAG $\mathcal{G}$ can be uniquely identified from observational data. Further, when the functions are linear but the noises are non-Gaussian, one can derive the LiNGAM as described in \citet{shimizu2006linear}, where the true DAG can be uniquely identified under certain favorable conditions. In addition, as cited in \citet{zheng2020learning,rolland2022score}, the nonlinear additive model can be identified from observational data. Another scenario of note arises when the corresponding MEC encompasses only one DAG; here, the DAG can be inherently identified from observational data. Recent score-based causal discovery algorithms \citep{zheng2018dags,yu2019dag,zhu2019causal,cai2020anoce} typically take into account synthetic datasets generated from fully identifiable models, which provides practical relevance in evaluating the estimated graph in relation to the true DAG.

In instances where the true DAG is not identifiable, we reference the discussion in \Cref{asec:mec}. In such cases, a CPDAG uniquely symbolizes a MEC of DAGs that yield the same joint distribution of variables. This CPDAG can be inferred from observational data via a variety of causal discovery algorithms \citep[see e.g.,][]{spirtes2000constructing,chickering2002optimal,shimizu2006linear,kalisch2007estimating,buhlmann2014cam,ramsey2017million}. One feasible approach to dealing with MEC involves enumerating all DAGs in the MEC derived from a given CPDAG \citep{chakrabortty2018inference}. It is conventional to encapsulate a range of potential effects or probabilities by their average or the minimum absolute value \citep{chakrabortty2018inference,shi2021testing}. However, such an approach typically proves computationally prohibitive for large graphs, necessitating computational shortcuts to acquire the causal effects or probabilities of causation without enumerating all DAGs in the MEC of the estimated CPDAG.

\subsection{Extended Algorithm for Markov Equivalence Class}

In contrast to existing causal discovery algorithms, we consider a causal graph that is necessary and sufficient to portray causal relationships influencing the outcome variable $Y$. This aim is expressed as a causal identification constraint designed to restrict the causal structural learning to a smaller class of DAGs, as detailed in the ensuing section. Additionally, we employ causal effects or probabilities of causation as another loss function within the objective. This assists in identifying the DAG with the highest score of conditional scores of causation or causal effects, wherein the $v$-structures of interest are also constrained to have an endpoint at $Y$. Based on the estimated CPDAG, which involves a significantly smaller number of nodes, we can generate all DAGs within the MEC and prune superfluous nodes following aggregation.

Specifically, the proposed NSCSL algorithm can be adapted to handle  MEC of PDAGs when the causal graph is not uniquely identifiable from observational data. Let's recall either the estimated matrix $\widehat{\boldsymbol{B}}$ obtained by NSCSL based on causal effects, as presented in \Cref{sec:algo} for the \textit{linear} model, or the estimated functional matrix $\widehat{\boldsymbol{B}}(\widehat{\psi})$ by the POC-based NSCSL in \Cref{asec:algo} for the \textit{nonlinear} model. Based on these estimated (functional) matrices, we can derive the estimated causal graph $\widehat{\mathcal{G}}_{\boldsymbol{V}}$. This can further lead to estimation under its MEC by averaging over possible DAGs, as follows:
\begin{eqnarray}
\widehat{\mathcal{G}}_{\boldsymbol{V}}^* = \frac{1}{|MEC(\widehat{\mathcal{G}}_{\boldsymbol{V}})|} \sum_{\mathcal{G}_i \in MEC(\widehat{\mathcal{G}}_{\boldsymbol{V}})} \mathcal{G}_i,
\end{eqnarray}
where $|MEC(\widehat{\mathcal{G}}_{\boldsymbol{V}})|$ is the size of MEC for $\widehat{\mathcal{G}}_{\boldsymbol{V}}$. 
As we have previously noted, the number of nodes in $\widehat{\mathcal{G}}_{\boldsymbol{V}}$ is much smaller than $p$, making it feasible to generate all DAGs in the MEC and prune extraneous nodes following aggregation.

 \section{Technical Proofs}\label{asec:pf}
 
 In this section, we provide proofs for \Cref{thm1,thm2}. Let the symbol $\wedge$ denote the logical connective $and$, and the symbol $\vee$ denote the logical connective $or$. For two events $A$ and $B$, $A\wedge B =$ True if $A=B=$ True, and $A\wedge B=$ False otherwise. Furthermore, $A\vee B =$ False if $A=B=$ False, and $A\vee B=$ True otherwise.

  \subsection{Proofs of \Cref{thm1}} 
Consider the marginal probability of causation (M-POC) for $Z_i$ in \Cref{defi_tpcPNS_i} such that
 \begin{equation*}
  \begin{split}
\text{M-POC}_i(y)
\equiv \mathbb{P}\{Y(Z_i\not =z_i)\not =y,Y(Z_i =z_i) =y\},
 \end{split}
 \end{equation*}
 and the conditional probability of causation (C-POC) for $Z_i$ in \Cref{defi_dpcPNS_i} such that
 \begin{equation*} 
  \text{C-POC}_i(y)
\equiv \mathbb{P}\{ Y(Z_i\not =z_i,\boldsymbol{Z}_{-i}=\boldsymbol{z}_{-i})\not =y, 
 Y(Z_i =z_i,\boldsymbol{Z}_{-i}=\boldsymbol{z}_{-i}) =y\}. 
 \end{equation*}
  Our goal is to establish their lower bounds. In the following, we derive the lower bound for M-POC in Part 1 and the lower bound for C-POC in Part 2. Finally, we discuss conditions for the lower bound equality to be held in Part 3.

\subsubsection{Part 1: Lower Bound for M-POC}
We focus on M-POC first. Based on the consistency assumption (A1), we have either $ \{Y(Z_i\not =z_i)=y\}$ or $\{Y(Z_i\not=z_i) \not= y \}$ holds.  Since the events $\{Y(Z_i\not=z_i)= y \}$ and $\{Y(Z_i\not=z_i) \not= y \}$ are disjoint, we have 
\begin{align}\label{pf_1_1}
    \{Y(Z_i\not=z_i)= y \} \vee  \{Y(Z_i\not=z_i) \not= y \} =\text{True},
\end{align}
where the symbol $\vee$ denote the logical connective $or$, meaning one of the above event holds. Based on this fact, we focus on the second event $\{Y(Z_i =z_i)=y\}$ in M-POC which yields 
 \begin{eqnarray}\label{pf_1}
 && \{Y(Z_i =z_i)=y\} \\\nonumber
 =&& \{Y(Z_i =z_i)=y\} \wedge \text{True}\\\nonumber
=&& \{Y(Z_i =z_i)=y\} \wedge\left[ \{Y(Z_i\not=z_i)= y \} \vee  \{Y(Z_i\not=z_i) \not= y \}\right]\\\nonumber
=&& \left[ \{Y(Z_i =z_i)=y\} \wedge \{Y(Z_i\not=z_i)= y  \}\right]\vee\left[ \{Y(Z_i =z_i)=y\} \wedge\{Y(Z_i\not=z_i) \not= y \}\right],
\end{eqnarray}
where the first equality is owing to the definition of the logical connective $ \wedge $, the second equality comes from  \eqref{pf_1_1}, and the last equality follows the rule of interchange in the logical connectives, i.e, $A\wedge(B\vee C) = (A\wedge B)\vee (A\wedge C)$ for events $A$, $B$, and $C$. By noticing the last line is the logical connective $or$ of two events, taking the probability on both sides of \eqref{pf_1}  gives
 \begin{eqnarray}\label{pf_1_new}
&& \mathbb{P}\{Y(Z_i =z_i)=y\} \\\nonumber
 \leq && \mathbb{P}\left[ \{Y(Z_i =z_i)=y\} \wedge \{Y(Z_i\not=z_i)= y \} \right]+\mathbb{P}\left[ \{Y(Z_i =z_i)=y\} \wedge\{Y(Z_i\not=z_i) \not= y \}\right]\\\nonumber
  = &&\underbrace{\mathbb{P}\left[ Y(Z_i =z_i)=y , Y(Z_i\not=z_i)= y \right]}_{\eta_1}+\mathbb{P}\left[ Y(Z_i =z_i)=y ,Y(Z_i\not=z_i) \not= y \right]\\\nonumber
  =&&\eta_1 + \text{M-POC}_i(y),
\end{eqnarray}
where the first inequality is owing to $\mathbb{P}(A\vee B) \leq\mathbb{P}(A)+\mathbb{P}(B)$, and the equalities are due to the definitions of probabilities.

Similarly, by (A1), since the events $\{Y(Z_i =z_i)= y \}$ and $\{Y(Z_i =z_i) \not= y \}$ are disjoint, we have 
\begin{align*} 
    \{Y(Z_i =z_i)= y \} \vee  \{Y(Z_i =z_i) \not= y \} =\text{True}.
\end{align*}
Based on this fact, the first event $\{Y(Z_i \not =z_i)=y\}$ in M-POC is 
 \begin{eqnarray}\label{pf_2}
  && \{Y(Z_i\not =z_i)=y\} \\\nonumber
 =&& \{Y(Z_i \not =z_i)=y\} \wedge \text{True}\\\nonumber 
=&& \{Y(Z_i \not =z_i)=y\} \wedge\left[ \{Y(Z_i=z_i)= y \} \vee  \{Y(Z_i=z_i) \not= y \}\right]\\\nonumber
=&& \left[ \{Y(Z_i \not=z_i)=y\} \wedge \{Y(Z_i=z_i)= y \} \}\right]\vee\left[ \{Y(Z_i \not=z_i)=y\} \wedge\{Y(Z_i=z_i) \not= y \}\right].
\end{eqnarray}
By noticing the last line is the logical connective $or$ of two events, taking the probability on both sides of \eqref{pf_2}  gives
 \begin{eqnarray}\label{pf_2_new}
 \mathbb{P}\{Y(Z_i\not =z_i)=y\} &\geq \mathbb{P}\left[ \{Y(Z_i \not=z_i)=y\} \wedge \{Y(Z_i=z_i)= y  \}\right]\\\nonumber
 &=\mathbb{P}\left[  Y(Z_i \not=z_i)=y, Y(Z_i=z_i)= y  \right] =\eta_1,
\end{eqnarray}
where the first inequality is owing to $\mathbb{P}(A\vee B)\geq \mathbb{P}(A)$ and the last equality comes from the definition of $\eta_1$. 

Combining \eqref{pf_1_new} and \eqref{pf_2_new}, we have 
 \begin{eqnarray}\label{pf_5_new}
&&\text{M-POC}_i(y)\\\nonumber
\text{(by \eqref{pf_1_new})} \quad \geq &&\mathbb{P}\{Y(Z_i =z_i)=y\}  -\eta_1 \\\nonumber
\text{(by \eqref{pf_2_new})} \quad  \geq &&\mathbb{P}\{Y(Z_i =z_i)=y\}  -  \mathbb{P}\{Y(Z_i\not =z_i)=y\}\\\nonumber
=&& \mathbb{P} \{Y=y|Z_i =z_i\}- \mathbb{P} \{Y=y|Z_i\not =z_i\},
\end{eqnarray}
where the last equation follows the results that $ \mathbb{P} \{Y=y|do(X=x)\}=\mathbb{P} \{Y=y|X=x\}$ under the ignorability assumption (A2) following \citet{rosenbaum1983central} and \citet{pearl2000models,pearl2009causal}. The proof of the first part thus is completed.

\subsubsection{Part 2: Lower Bound for C-POC}  We next show the lower bound of conditional POC in \Cref{thm1} following the same logic as in Part 1. 
Based on the consistency assumption (A1), we have either $ \{Y(Z_i\not =z_i,\boldsymbol{Z}_{-i}=\boldsymbol{z}_{-i})=y\}$ or $\{Y(Z_i\not=z_i,\boldsymbol{Z}_{-i}=\boldsymbol{z}_{-i}) \not= y \}$ holds, i.e., these two events are disjoint, thus, we have 
\begin{align}\label{pf_1_1_cpoc}
    \{Y(Z_i\not=z_i,\boldsymbol{Z}_{-i}=\boldsymbol{z}_{-i})= y \} \vee  \{Y(Z_i\not=z_i,\boldsymbol{Z}_{-i}=\boldsymbol{z}_{-i}) \not= y \} =\text{True},
\end{align}
where the symbol $\vee$ denote the logical connective $or$, meaning one of the above event holds. Based on this fact, we focus on the second event $\{Y(Z_i =z_i,\boldsymbol{Z}_{-i}=\boldsymbol{z}_{-i})=y\}$ in C-POC which yields 
 \begin{eqnarray}\label{pf_1_cpoc}
 && \{Y(Z_i =z_i,\boldsymbol{Z}_{-i}=\boldsymbol{z}_{-i})=y\} \\\nonumber
 =&& \{Y(Z_i =z_i,\boldsymbol{Z}_{-i}=\boldsymbol{z}_{-i})=y\} \wedge \text{True}\\\nonumber
=&& \{Y(Z_i =z_i,\boldsymbol{Z}_{-i}=\boldsymbol{z}_{-i})=y\}  \\\nonumber
&&  ~~~~~~~~~~~~~~~~\wedge\left[ \{Y(Z_i\not=z_i,\boldsymbol{Z}_{-i}=\boldsymbol{z}_{-i})= y \} \vee  \{Y(Z_i\not=z_i,\boldsymbol{Z}_{-i}=\boldsymbol{z}_{-i}) \not= y \}\right]\\\nonumber
=&& \left[ \{Y(Z_i =z_i,\boldsymbol{Z}_{-i}=\boldsymbol{z}_{-i})=y\} \wedge \{Y(Z_i\not=z_i,\boldsymbol{Z}_{-i}=\boldsymbol{z}_{-i})= y  \}\right] \\\nonumber
&&~~~~~~~~~~~~~~~~\vee\left[ \{Y(Z_i =z_i,\boldsymbol{Z}_{-i}=\boldsymbol{z}_{-i})=y\} \wedge\{Y(Z_i\not=z_i,\boldsymbol{Z}_{-i}=\boldsymbol{z}_{-i}) \not= y \}\right],
\end{eqnarray}
where the first equality is owing to the definition of the logical connective $ \wedge $, the second equality comes from  \eqref{pf_1_1_cpoc}, and the last equality follows the rule of interchange in the logical connectives, i.e, $A\wedge(B\vee C) = (A\wedge B)\vee (A\wedge C)$ for events $A$, $B$, and $C$. By noticing the last line is the logical connective $or$ of two events, taking the probability on both sides of \eqref{pf_1_cpoc}  gives
 \begin{eqnarray}\label{pf_1_new_cpoc}
&& \mathbb{P}\{Y(Z_i =z_i,\boldsymbol{Z}_{-i}=\boldsymbol{z}_{-i})=y\} \\\nonumber
 \leq && \mathbb{P}\left[ \{Y(Z_i =z_i,\boldsymbol{Z}_{-i}=\boldsymbol{z}_{-i})=y\} \wedge \{Y(Z_i\not=z_i,\boldsymbol{Z}_{-i}=\boldsymbol{z}_{-i})= y \} \right]\\\nonumber
&&~~~~~~~~~~~~~~~~+\mathbb{P}\left[ \{Y(Z_i =z_i,\boldsymbol{Z}_{-i}=\boldsymbol{z}_{-i})=y\} \wedge\{Y(Z_i\not=z_i,\boldsymbol{Z}_{-i}=\boldsymbol{z}_{-i}) \not= y \}\right]\\\nonumber
  = &&\underbrace{\mathbb{P}\left[ Y(Z_i =z_i,\boldsymbol{Z}_{-i}=\boldsymbol{z}_{-i})=y , Y(Z_i\not=z_i,\boldsymbol{Z}_{-i}=\boldsymbol{z}_{-i})= y \right]}_{\eta_2}\\\nonumber
&&~~~~~~~~~~~~~~~~+\mathbb{P}\left[ Y(Z_i =z_i,\boldsymbol{Z}_{-i}=\boldsymbol{z}_{-i})=y ,Y(Z_i\not=z_i,\boldsymbol{Z}_{-i}=\boldsymbol{z}_{-i}) \not= y \right]\\\nonumber
  =&&\eta_2 + \text{C-POC}_i(y),
\end{eqnarray}
where the first inequality is owing to $\mathbb{P}(A\vee B) \leq\mathbb{P}(A)+\mathbb{P}(B)$, and the equalities are due to the definitions of probabilities. Similarly, by (A1), since the events $\{Y(Z_i =z_i,\boldsymbol{Z}_{-i}=\boldsymbol{z}_{-i})= y \}$ and $\{Y(Z_i =z_i,\boldsymbol{Z}_{-i}=\boldsymbol{z}_{-i}) \not= y \}$ are disjoint, we have 
\begin{align*} 
    \{Y(Z_i =z_i,\boldsymbol{Z}_{-i}=\boldsymbol{z}_{-i})= y \} \vee  \{Y(Z_i =z_i,\boldsymbol{Z}_{-i}=\boldsymbol{z}_{-i}) \not= y \} =\text{True}.
\end{align*}
Based on this fact, the first event $\{Y(Z_i \not =z_i,\boldsymbol{Z}_{-i}=\boldsymbol{z}_{-i})=y\}$ in C-POC is 
 \begin{eqnarray}\label{pf_2_cpoc}
  && \{Y(Z_i\not =z_i,\boldsymbol{Z}_{-i}=\boldsymbol{z}_{-i})=y\} \\\nonumber
 =&& \{Y(Z_i \not =z_i,\boldsymbol{Z}_{-i}=\boldsymbol{z}_{-i})=y\} \wedge \text{True}\\\nonumber 
=&& \{Y(Z_i \not =z_i,\boldsymbol{Z}_{-i}=\boldsymbol{z}_{-i})=y\} \\\nonumber
&&~~~~~~~~~~~~~~~~\wedge\left[ \{Y(Z_i=z_i,\boldsymbol{Z}_{-i}=\boldsymbol{z}_{-i})= y \} \vee  \{Y(Z_i=z_i,\boldsymbol{Z}_{-i}=\boldsymbol{z}_{-i}) \not= y \}\right]\\\nonumber
=&& \left[ \{Y(Z_i \not=z_i,\boldsymbol{Z}_{-i}=\boldsymbol{z}_{-i})=y\} \wedge \{Y(Z_i=z_i,\boldsymbol{Z}_{-i}=\boldsymbol{z}_{-i})= y \} \}\right]\\\nonumber
&&~~~~~~~~~~~~~~~~\vee\left[ \{Y(Z_i \not=z_i,\boldsymbol{Z}_{-i}=\boldsymbol{z}_{-i})=y\} \wedge\{Y(Z_i=z_i,\boldsymbol{Z}_{-i}=\boldsymbol{z}_{-i}) \not= y \}\right].
\end{eqnarray}
By noticing the last line is the logical connective $or$ of two events, taking the probability on both sides of \eqref{pf_2_cpoc}  gives
 \begin{eqnarray}\label{pf_2_new_cpoc}
 &&\mathbb{P}\{Y(Z_i\not =z_i,\boldsymbol{Z}_{-i}=\boldsymbol{z}_{-i})=y\}  \\\nonumber 
  \geq&& \mathbb{P}\left[ \{Y(Z_i \not=z_i,\boldsymbol{Z}_{-i}=\boldsymbol{z}_{-i})=y\} \wedge \{Y(Z_i=z_i,\boldsymbol{Z}_{-i}=\boldsymbol{z}_{-i})= y  \}\right]\\\nonumber
  =&&\mathbb{P}\left[  Y(Z_i \not=z_i,\boldsymbol{Z}_{-i}=\boldsymbol{z}_{-i})=y, Y(Z_i=z_i,\boldsymbol{Z}_{-i}=\boldsymbol{z}_{-i})= y  \right] =\eta_2,
\end{eqnarray}
where the first inequality is owing to $\mathbb{P}(A\vee B)\geq \mathbb{P}(A)$ and the last equality comes from the definition of $\eta_2$. Combining \eqref{pf_1_new_cpoc} and \eqref{pf_2_new_cpoc}, we have 
 \begin{eqnarray}\label{pf_5_new_cpoc}
&&\text{C-POC}_i(y)\\\nonumber
\text{(by \eqref{pf_1_new_cpoc})} \quad \geq &&\mathbb{P}\{Y(Z_i =z_i,\boldsymbol{Z}_{-i}=\boldsymbol{z}_{-i})=y\}  -\eta_2 \\\nonumber
\text{(by \eqref{pf_2_new_cpoc})} \quad  \geq &&\mathbb{P}\{Y(Z_i =z_i,\boldsymbol{Z}_{-i}=\boldsymbol{z}_{-i})=y\}  -  \mathbb{P}\{Y(Z_i\not =z_i,\boldsymbol{Z}_{-i}=\boldsymbol{z}_{-i})=y\}\\\nonumber
=&& \mathbb{P} \{Y=y|Z_i =z_i,\boldsymbol{Z}_{-i}=\boldsymbol{z}_{-i}\}- \mathbb{P} \{Y=y|Z_i\not =z_i,\boldsymbol{Z}_{-i}=\boldsymbol{z}_{-i}\},
\end{eqnarray}
where the last equation follows the results that $ \mathbb{P} \{Y=y|do(X=x)\}=\mathbb{P} \{Y=y|X=x\}$ under the ignorability assumption (A2) following \citet{rosenbaum1983central} and \citet{pearl2000models,pearl2009causal}. The proof of the second part thus is completed.

\subsubsection{Part 3: Conditions to Achieve  Lower Bounds}\label{part_3_equal}
In this part, we discuss the conditions under which the POCs are equal to their corresponding lower bounds. To this end, we introduce the following monotonicity condition.

   \noindent (C1*). \textbf{Monotonicity}:\\ 
      \noindent (i) 
      $\{Y(\boldsymbol{Z}\not=\boldsymbol{z}) = y \} \wedge  \{Y(\boldsymbol{Z}=\boldsymbol{z}) \not= y \} =\text{False};$ \\
  \noindent (ii) 
  $ \{Y(Z_i\not=z_i)= y \} \wedge  \{Y(Z_i=z_i) \not= y \} =\text{False}.$

 Here, (C1*.i) is proposed in Section 9.2.3 in \citet{pearl2000models} and also \citet{tian2000probabilities} to establish the identifiability of the probability of causation.  
We generalize the condition (C1*.i) to the condition (C1*.ii) so that we can extend the results in Theorem 9.2.14 in \citet{pearl2000models} for POCs in \Cref{defi_dpcPNS_i,defi_tpcPNS_i}.

We detail the case of M-POC first. By noticing the monotonicity condition in (C1*.ii) such that $\left[ \{Y(Z_i \not=z_i)=y\} \wedge\{Y(Z_i=z_i) \not= y \}\right]=\text{False}$, we can simplify \eqref{pf_2} as
 \begin{eqnarray}\label{pf_3}
 \{Y(Z_i\not =z_i)=y\} = \left[ \{Y(Z_i \not=z_i)=y\} \wedge \{Y(Z_i=z_i)= y   \}\right].
\end{eqnarray}
Substituting  \eqref{pf_3} into  \eqref{pf_1} yields
 \begin{eqnarray}\label{pf_4}
 \{Y(Z_i =z_i)=y\} = \{Y(Z_i\not =z_i)=y\} \vee\left[ \{Y(Z_i =z_i)=y\} \wedge\{Y(Z_i\not=z_i) \not= y \}\right].
\end{eqnarray}
Based on the consistency assumption (A1), we have either $ \{Y(Z_i\not =z_i)=y\}$ or $\{Y(Z_i\not=z_i) \not= y \}$ holds, and thus the events $\{Y(Z_i\not =z_i)=y\}$ and $\left[ \{Y(Z_i =z_i)=y\} \wedge\{Y(Z_i\not=z_i) \not= y \}\right]$ are disjoint. Therefore, taking the probability on both sides of \eqref{pf_4}  gives
 \begin{eqnarray}\label{pf_5}
\mathbb{P} \{Y(Z_i =z_i)=y\} = \mathbb{P} \{Y(Z_i\not =z_i)=y\} + \mathbb{P} \{Y(Z_i =z_i)=y,Y(Z_i\not=z_i) \not= y \} .
\end{eqnarray}
Recall \Cref{defi_tpcPNS_i}. Based on \eqref{pf_5}, we have
 \begin{eqnarray*}
\text{M-POC}_i(y)~~~=&&\mathbb{P}\{Y(Z_i\not =z_i)\not =y,Y(Z_i =z_i) =y\}\\
=&& \mathbb{P} \{Y(Z_i =z_i)=y\}- \mathbb{P} \{Y(Z_i\not =z_i)=y\}\\
=&& \mathbb{P} \{Y=y|Z_i =z_i\}- \mathbb{P} \{Y=y|Z_i\not =z_i\},
\end{eqnarray*}
where the last equation follows the results that $ \mathbb{P} \{Y=y|do(X=x)\}=\mathbb{P} \{Y=y|X=x\}$ under the Ignorability assumption by \citet{rosenbaum1983central} and \citet{pearl2000models,pearl2009causal}.   
Thus, the lower bound equality for M-POC holds when an additional monotonicity condition is imposed.
Following the same logic, we can show the lower bound equality for C-POC holds given the monotonicity condition. We omit the details for brevity.

 \subsection{Proofs of \Cref{thm2}}
As a direct result of \Cref{thm1}, below we can establish the relationship between POC and the corresponding expected mean outcome given different combinations of the confounders involving $Z_i$. Specifically, we take expectations over $Y$ on both sides of \eqref{pf_5_new} and \eqref{pf_5_new_cpoc}. When $Y$ is nonnegative, this yields
 \begin{equation*} 
  \begin{split} 
\sum_{y\in \mathcal{L}} y\text{M-POC}_i(y)  \geq &\sum_{y\in \mathcal{L}} y \left[ \mathbb{P}\{Y(Z_i =z_i)=y\}  -  \mathbb{P}\{Y(Z_i\not =z_i)=y\}\right]\\
\geq & \mathbb{E}\{Y(Z_i =z_i)\}- \mathbb{E}\{Y(Z_i \not=z_i)\}, 
  \end{split} 
 \end{equation*} 
 and
 \begin{equation*} 
  \begin{split} 
\sum_{y\in \mathcal{L}} y\text{C-POC}_i(y) \geq & \sum_{y\in \mathcal{L}} y \left[\mathbb{P}\{Y(Z_i =z_i,\boldsymbol{Z}_{-i}=\boldsymbol{z}_{-i})=y\}  -  \mathbb{P}\{Y(Z_i\not =z_i,\boldsymbol{Z}_{-i}=\boldsymbol{z}_{-i})=y\}\right]\\
\geq &  \mathbb{E}\{Y(Z_i =z_i, \boldsymbol{Z}_{-i}=\boldsymbol{z}_{-i})\} - \mathbb{E}\{Y(Z_i \not=z_i, \boldsymbol{Z}_{-i}=\boldsymbol{z}_{-i})\}.
  \end{split} 
 \end{equation*} 
 Under the ignorability assumption (A2) following \citet{rosenbaum1983central} and \citet{pearl2000models,pearl2009causal}, we have 
\begin{equation}\label{fist_res_poc} 
  \begin{split}
& \sum_{y\in \mathcal{L}} y\text{M-POC}_i(y)  \geq  \delta_M(z_i) \equiv \mathbb{E}\{Y|Z_i =z_i\}- \mathbb{E}\{Y|Z_i \not=z_i\} ,\\
& \sum_{y\in \mathcal{L}} y\text{C-POC}_i(y)  \geq \delta_C(z_i) \equiv\mathbb{E}\{Y|Z_i =z_i, \boldsymbol{Z}_{-i}=\boldsymbol{z}_{-i}\} - \mathbb{E}\{Y|Z_i \not=z_i, \boldsymbol{Z}_{-i}=\boldsymbol{z}_{-i}\},
  \end{split} 
 \end{equation}   
 where $\delta_M(z_i)$ and $\delta_C(z_i)$ are defined as the marginal and conditional causal effects using the differences of expectations based on the corresponding POC. 
Recall the definitions of natural causal effects for $Z_i$ as
\begin{eqnarray*} 
&&TE_i  =
\mathbb{E}\{Y(Z_i=z_i+1)\}-\mathbb{E}\{Y(Z_i=z_i)\}
,\\
&&DE_i  = 
\mathbb{E}\{Y(Z_i=z_i+1, \boldsymbol{Z}_{-i}=\boldsymbol{z}_{-i}^{(z_i)})\}-\mathbb{E}\{Y(Z_i=z_i)\},
\end{eqnarray*}
where $\boldsymbol{z}_{-i}^{(z_i)}$ is the value of $ \boldsymbol{Z}_{-i}$ if setting $do(Z_i=z_i)$. When $Z_i$ is binary, by comparing the definitions with $Z_i\in\{0,1\}$, we have 
\begin{eqnarray} \label{second_res_poc} 
|TE_i| \geq \delta_M(z_i), \quad |DE_i| \geq \delta_C(z_i).
\end{eqnarray}
Therefore, combining \eqref{fist_res_poc} 
 and \eqref{second_res_poc} yields the second conclusion in \Cref{thm2} that 
\begin{equation*} 
  \begin{split}
\min\{\sum_{y\in \mathcal{L}} y\text{M-POC}_i(y) , |TE_i| \}&\geq \mathbb{E}\{Y|Z_i =z_i\}- \mathbb{E}\{Y|Z_i \not=z_i\} ,\\
\min\{\sum_{y\in \mathcal{L}} y\text{C-POC}_i(y) , |DE_i|\}&\geq \mathbb{E}\{Y|Z_i =z_i, \boldsymbol{Z}_{-i}=\boldsymbol{z}_{-i}\} - \mathbb{E}\{Y|Z_i \not=z_i, \boldsymbol{Z}_{-i}=\boldsymbol{z}_{-i}\}.
  \end{split} 
 \end{equation*}  
Further, when $Z_i$ is binary, the absolute values of $\delta_M(z_i)$ and $\delta_C(z_i)$ is equal to the absolute values of $TE_i$ and $DE_i$, respectively.
Combining this with  \eqref{fist_res_poc} yields the second conclusion in \Cref{thm2}.
Finally, following the same logic in \Cref{part_3_equal}, we can show the lower bound equality for \Cref{thm2} holds when the monotonicity condition is imposed. We omit the details for brevity. The proof is thus completed.

\subsection{Proofs of \Cref{theoremA}}

We investigate the theoretical consistency of the proposed causal structural learning methods under Model \eqref{lsem_x} with independent Gaussian error and equal variance, using the score-based method such as NOTEARS that minimizes the loss in \eqref{loss_new}.  

\textbf{Notations and Conditions:} We first detail some notations and the required conditions in \Cref{theoremA} below. 
Denote $\bm{W}_{j}\in\mathbb{R}^{n\times 1}$ as the row vector of a matrix $\bm{W}\in\mathbb{R}^{(d+1)\times n}$ for $j=1,\cdots d+1$. Let $\supp(v)=\{i:v_i\ne 0\}$ denote the support of a vector $v$, which is the set of indices of nonzero terms of $v$. The first condition requires a bounded true matrix $\bm{B}$ as follows.

\begin{condition}\label{condi:zero_row_column}
The true matrix $\bm{B}\in\mathbb{R}^{(d+1)\times (d+1)}$ is bounded such that $\|\bm{B}\|_2=\mathcal{O}(1)$, 
and the maximum degree across different rows is less than the number of nodes in the graph, i.e., $s_0 = \max_j{\supp(\bm{B}_j)}\le d+1$. 
\end{condition}

Next, we recall the linear structural equation model condition on $\boldsymbol{W}\equiv[g(\boldsymbol{Z})^\top,Y]^\top$ as follows.

\begin{condition}\label{condi:lsem}
Suppose Model \eqref{lsem_x} such that $\boldsymbol{W}=\bm{B}^\top \boldsymbol{W}+\bm{\epsilon}$, with independent Gaussian error $\bm{\epsilon}$ and the error variance is a constant $\sigma^2$. 
\end{condition}

Furthermore, define the order as $\nu=[\nu_{1},\nu_{2},\cdots,\nu_{d+1}]^\top$ is a permutation of indices $\{1,2,...,d,d+1\}$ of nodes in the causal graph, such that $\nu_{j}\in\{1,2,...,d+1\}$. If we set the last node to be the target variable, then the last index is fixed to be $d+1$. 
Let $\bm{B}(\nu)$ as $\bm{B}(\nu)=[\bm{B}_1(\nu),\cdots,\bm{B}_{d+1}(\nu)]^\top$, such that \begin{equation*}
    \bm{B}_{j}(\nu)=\argmin_{\bm{\beta}:\supp(\bm{\beta})\in\{\nu_1,\cdots,\nu_{j-1}\}}
    \mathbb{E}(\bm{W}_{\nu_j}-\bm{\beta}^\top \bm{W})^2, \quad j=1,2,\cdots,d+1.
\end{equation*} 
Note that the true causal graph $\bm{B}$ that generates $\bm{W}$ also has a topological order, $\nu^*$, which is called the true topological order (this true order may not be unique). Then the true causal graph $\bm{B}$ can also be denoted as $\bm{B}(\nu^*)$. Denote the set of all permuations of $\{1,2,\cdots,d+1\}$ as $\Upsilon$ and the set of true order as $\Upsilon^*$ such that $\Upsilon^*\subset \Upsilon$.  Denote the order of $\widehat{\bm{B}}$ as $\nu\in \Upsilon$, so $\widehat{\bm{B} }$ can also be notated as $\widehat{\bm{B} }(\nu)$. Then let $s_j(\nu)=\supp(\bm{B}_j(\nu)),\widehat{s}_j(\nu)=\supp(\widehat{\bm{B}}_j(\nu))$. 
The last condition assumes the consistency of the topological ordering.


\begin{condition}\label{condi:true_order}
The true topological ordering of $\bm{B}$, i.e., $\nu^*$, is consistently estimated.
\end{condition}
Condition \ref{condi:true_order} is commonly imposed when proving the error bound of causal structural learning results \citep[see Condition (A6) in ][]{shi2021testing}.  

\textbf{Overview of Proof:} With the aforementioned three conditions, our proof follows similar strategies of the causal structural learning literature \citep[e.g.,][]{shi2021testing} but accounts for the extra penalty term from causal effects. Notice that the explicit forms of causal effects under LSEM are linear combinations of elements of $\boldsymbol{B}$. 
This implies our new regulation can similarly vanish away as $n$ goes to infinity. For simplicity, in the rest of the proof, we show the consistency given a selection function $g$ 
for brevity. 
To start with, proving the consistency of $\widehat{\bm{B}}$ 
 returned by NSCSL with NOTEARS as baseline is equivalent to showing that the $\widehat{\bm{B}}$  solving \begin{equation}\label{eqa:soft_constraint_solve_notears}
    \begin{aligned}
        \widehat{\bm{B} } = \argmin {\|\bm{W}-\bm{B}^\top \bm{W}\|_2^2} + \lambda h_2 (\bm{B}), \quad \text{subject to } h_1(\bm{B}) = 0.
    \end{aligned}
    \end{equation}
    is consistent.  
 Then the proof of \Cref{theoremA} follows the same strategy as the proof of Proposition 1 in \cite{shi2021testing}. The key difference between \Cref{theoremA} and Proposition 1 in \cite{shi2021testing} is that the loss function in   \eqref{eqa:soft_constraint_solve_notears} contains extra penalty term $  h_2(\boldsymbol{B})=\delta^*- \sum_{i=1}^{d} |\widehat{CE}_i(\boldsymbol{B})| +\sum_{i=1}^{d+1} |b_{i,d+1}|$, that involves causal effect information compared to the original loss function \begin{equation*}\label{eqa:original_notears_solve}
    \begin{aligned}
\widetilde{\bm{B} } = \argmin {\|\bm{W}-\bm{B}^\top \bm{W}\|_2^2}, \quad \text{subject to } h_1(\bm{B}) = 0,
    \end{aligned}
\end{equation*}  
in \cite{zheng2018dags}. 
In this proof, we need to show a main statement and the rest of the proof will follow the same procedure of Steps 2-3 in Appendix Section 9 of \cite{shi2021testing}. The \textbf{Main Statement} is: 
\begin{equation*}
    \begin{aligned}
        \big\|\widehat{\bm{B}}(\nu)-\bm{B}(\nu)\big\|_2 = \mathcal{O}\big(\sqrt{\sum_{j=1}^{d+1}{\frac{\log n}{n}(s_j(\nu)+\widehat{s}_j(\nu))}}+\frac{\lambda}{n} \sqrt{s_0 (d+1)} \big),
    \end{aligned}
\end{equation*}
with $\lambda$  satisfy the bound of $\mathcal{O}({(n\log{n} )}^{1/2})$ for arbitrary $\nu \in \Upsilon$.  

        \textbf{Proof of Main Statement}:
Since $\widehat{\bm{B}}(\nu)$ solves \eqref{eqa:soft_constraint_solve_notears}, 
we have  
\begin{equation}\label{eqa:equiv_62_prop_5_1}
    \begin{aligned}
{\|\bm{W}-\widehat{\bm{B}}(\nu)^\top \bm{W}\|_2^2} + \lambda h_2 (\widehat{\bm{B}}(\nu))
 & \le
{\|\bm{W}-\bm{B}(\nu)^\top \bm{W}\|_2^2} + \lambda h_2 (\bm{B}(\nu)).
    \end{aligned}
\end{equation}
Following Theorem 7.1 in \citet{van20130} and \citet{shi2021testing}, we can transform \eqref{eqa:equiv_62_prop_5_1}  to
\begin{equation}\label{eqa:equiv_follow_64_prop_5_1}
    \begin{aligned}
{{1\over 2}\|{\bm{B}}(\nu)^\top \bm{W}-\widehat{\bm{B}}(\nu)^\top \bm{W}\|_2^2}+ \lambda h_2 (\widehat{\bm{B}}(\nu))\leq 2\sum_{j=1}^{d+1}\kappa(s_j(\nu)+\widehat{s}_j(\nu))\log{n}+ \lambda h_2 (\bm{B}(\nu)),
    \end{aligned}
\end{equation}
for some $\kappa>0$. 
Recall \Cref{condi:lsem} such that  $
    \bm{W}=\bm{B}^\top \bm{W}+\bm{\epsilon}$ with $\bm{\epsilon}\sim N(\bm{0}^{n\times 1},\bm{I}_n\times\sigma^2)$. 
From Equation (65) in \cite{shi2021testing}, we have $0<\kappa^*<\sigma^2$ such that \begin{equation}\label{eqa:equiv_65_prop_5_1}
    \kappa^* n\sum_{j=1}^{d+1}\|\bm{B}_j(\nu)-\widehat{\bm{B}}_j(\nu)\|_2^2\le {{1\over 2}\|{\bm{B}}(\nu)^\top \bm{W}-\widehat{\bm{B}}(\nu)^\top \bm{W}\|_2^2}\le    \frac{1}{\kappa^* } n\sum_{j=1}^{d+1}\|\bm{B}_j(\nu)-\widehat{\bm{B}}_j(\nu)\|_2^2, \forall \nu . 
\end{equation}
Combining \eqref{eqa:equiv_65_prop_5_1} with \eqref{eqa:equiv_follow_64_prop_5_1}, we have \begin{equation}\label{eqa:equiv_66_prop_5_1}
    \begin{aligned}
        \kappa^* n\sum_{j=1}^{d+1}\|\bm{B}_j(\nu)-\widehat{\bm{B}}_j(\nu)\|_2^2 + \lambda h_2 (\widehat{\bm{B}}(\nu))  \le 2\sum_{j=1}^{d+1}\kappa(s_j(\nu)+\widehat{s}_j(\nu))\log{n}+ \lambda h_2 ({\bm{B}}(\nu)). 
    \end{aligned}
\end{equation}
If $ \lambda h_2 (\widehat{\bm{B}}(\nu)) > \lambda h_2 ({\bm{B}}(\nu))$, the main statement directly holds. 

In the other case, we focus on showing the main statement when $ \lambda h_2 (\widehat{\bm{B}}(\nu)) \leq \lambda h_2 ({\bm{B}}(\nu))$. Specifically, we first define a vector 
\begin{equation*}
    \bm{\mu}(\nu)=[\widehat{\bm{B}}_1(\nu)-\bm{B}_1(\nu),\widehat{\bm{B}}_2(\nu)-\bm{B}_2(\nu),\cdots,\widehat{\bm{B}}_p(\nu)-\bm{B}_{d+1}(\nu)]^\top\in\mathbb{R}^{(d+1) \times 1}.
\end{equation*}
Our goal is to bound $\|\bm{\mu}(\nu)\|_2 = \sum_{j=1}^{d+1}\|\widehat{\bm{B}}_j(\nu)-\bm{B}_j(\nu)\|^2_2$. Define $\mathcal{M}(\nu)=\supp(\bm{\mu}(\nu))$ and $\mathcal{M}^c(\nu)$ is the complementary set. 
Then denote $\bm{\mu}(\nu)_{\mathcal{M}(\nu)}$ as the vector formed by elements of $\bm{\mu}(\nu)$ in $\mathcal{M}(\nu)$. Denote $\bm{\mu}(\nu)_{\mathcal{M}^c(\nu)}$ as the vector formed by elements of $\bm{\mu}(\nu)$ in $\mathcal{M}^c(\nu)$. Hence, we can show \begin{equation}\label{eqa:equiv_follow_67_key1_prop_5_1}
\begin{aligned}
&\|\lambda h_2 ({\bm{B}}(\nu)) \|_1-
\|\lambda h_2 (\widehat{\bm{B}}(\nu))  \|_1
 \\ \le 	& \left\|\lambda \sum_{i=1}^{d} \left\{{CE}_i(\boldsymbol{B}(\nu))-\widehat{CE}_i(\widehat{\boldsymbol{B}}(\nu))\right\} +\lambda \left\{\boldsymbol{B}_{d+1}(\nu)- \widehat{\boldsymbol{B}}_{d+1}(\nu)\right\} \right\|_1 
 \\ \le &	   \lambda  \sum_{i=1}^{d}\left\| {CE}_i(\boldsymbol{B}(\nu))-\widehat{CE}_i(\widehat{\boldsymbol{B}}(\nu))  \right\|_1 +\lambda\big\|\boldsymbol{B}_{d+1}(\nu)- \widehat{\boldsymbol{B}}_{d+1}(\nu)  \big\|_1.
 \end{aligned}
\end{equation}
Recall the close form of causal effects in \Cref{FSTE_FSDE} derived in \Cref{sec:est} under the LSEM model. We have 
\begin{eqnarray*}
DE_i(\boldsymbol{B};g)= \theta_i,
\end{eqnarray*}
where $\theta_i$ presents the weight of the direct edge $g(\boldsymbol{Z})_i\rightarrow Y$ according to  \eqref{lsem_x} and \Cref{FSTE_FSDE}. In addition, the total causal effect can be quantified by the path method \citep[see e.g., ][]{Wright1921CorrelationAndCausation,nandy2017estimating} as
\begin{align*}
TE_i(\boldsymbol{B};g)= \sum_{k=1}^{m_i} PE\{\pi_{i}^{(k)}\},
\end{align*}
where 
$ PE\{\pi_{i}^{(k)}\} = 
b_{i,l_1} \cdots b_{l_{\tau_k},(d+1)}$ is the causal effect of $g_i(\boldsymbol{Z})$ on $Y$ through the directed path $\pi_{i}^{(k)} =\{i,l_1, \cdots, l_{\tau_k}, d+1\} \in \pi_{i}$ with length ${\tau_k}+1$, 
and $b_{i,j}$ is the weight of the edge $g_i(\boldsymbol{Z})\rightarrow g_j(\boldsymbol{Z})$ if it exists, and $b_{i,j}=0$ otherwise, for $i, j \in \{1,\cdots, d\}$, and $b_{l_{\tau_k},(d+1)} = \theta_{l_{\tau_k}}$ as the direct edge from $g_{l_{\tau_k}}(\boldsymbol{Z})$ to $Y$.  
Both $TE_i$ and $DE_i$ can be explicitly calculated given a matrix $\boldsymbol{B}$ under a selector $g$. We denote their estimates as $\widehat{TE}_i$ and $\widehat{DE}_i$ given the estimated matrix $\widehat{\boldsymbol{B}}$ and $g$. Using the direct causal effects as an example, we have \eqref{eqa:equiv_follow_67_key1_prop_5_1} be further bounded by
\begin{equation}\label{eqa:equiv_follow_67_key1_prop_5_1_exp0}
\begin{aligned}
&\lambda   \sum_{i=1}^{d} \left\| {CE}_i(\boldsymbol{B}(\nu))-\widehat{CE}_i(\widehat{\boldsymbol{B}}(\nu))  \right\|_1 +\lambda\big\|\boldsymbol{B}_{d+1}(\nu)- \widehat{\boldsymbol{B}}_{d+1}(\nu)  \big\|_1
 \\ \le 	&  C_1 \lambda \sum_{i=1}^d\big\| \bm{B}_i(\nu) -\widehat{\bm{B}}_i(\nu)
  \big\|_1 + \lambda \big\|\boldsymbol{B}_{d+1}(\nu)- \widehat{\boldsymbol{B}}_{d+1}(\nu)  \big\|_1
\\ \le 	&  \max\{C_1,1\} \lambda  \sum_{i=1}^{d+1} \big\| \bm{B}_i(\nu) -\widehat{\bm{B}}_i(\nu)
  \big\|_1,
\end{aligned}    
\end{equation}
for some constant $C_1>0$. 
Recall $\|\bm{\mu}(\nu)\|_2 = \sum_{j=1}^{d+1}\|\widehat{\bm{B}}_j(\nu)-\bm{B}_j(\nu)\|^2_2$. Combining \eqref{eqa:equiv_follow_67_key1_prop_5_1} and \eqref{eqa:equiv_follow_67_key1_prop_5_1_exp0}, we have
\begin{equation}\label{eqa:equiv_follow_67_key1_prop_5_1_exp}
\begin{aligned}
&\|\lambda h_2 ({\bm{B}}(\nu)) \|_1-
\|\lambda h_2 (\widehat{\bm{B}}(\nu))  \|_1
  \le 	 \max\{C_1,1\} \lambda  \sum_{i=1}^{d+1} \big\| \bm{B}_i(\nu) -\widehat{\bm{B}}_i(\nu)
  \big\|_1  
    \\ \le	  & \max\{C_1,1\} \lambda\|\bm{\mu}(\nu)\|_1 \le  \max\{C_1,1\} \lambda(\| \bm{\mu}(\nu)_{\mathcal{M}(\nu)}\|_1+\| \bm{\mu}(\nu)_{\mathcal{M}^c(\nu)}\|_1 ) 
 \\ \le	  & 2 \max\{C_1,1\} \lambda \sqrt{s_0 (d+1)} \|\bm{\mu}(\nu)_{\mathcal{M}(\nu)}\|_2,
\end{aligned}    
\end{equation}
 where the last inequality follows Equation (67) in \citet{shi2021testing}. 
Together with \eqref{eqa:equiv_66_prop_5_1}, we have
\begin{equation*}
    \begin{aligned}
       & \kappa^* n\sum_{j=1}^{d+1}\|\bm{B}_j(\nu)-\widehat{\bm{B}}_j(\nu)\|_2^2 -2\sum_{j=1}^{d+1}\kappa(s_j(\nu)+\widehat{s}_j(\nu))\log{n}
\\ =&         \kappa^* n\|\bm{\mu}(\nu)\|_2 -2\sum_{j=1}^{d+1}\kappa(s_j(\nu)+\widehat{s}_j(\nu))\log{n} \le 2 \max\{C_1,1\} \lambda \sqrt{s_0 (d+1)} \|\bm{\mu}(\nu)_{\mathcal{M}(\nu)}\|_2,
    \end{aligned}
\end{equation*}
and thus
\begin{equation}\label{eqa:equiv_by_66_prop_5_1}
    \begin{aligned}
        &\kappa^* \|\bm{\mu}(\nu)\|_2 
\le  & 2\sum_{j=1}^{d+1}\kappa(s_j(\nu)+\widehat{s}_j(\nu))\frac{\log{n}}{n} +2 \max\{C_1,1\} {\lambda \sqrt{s_0 (d+1)} \over n}\|\bm{\mu}(\nu)_{\mathcal{M}(\nu)}\|_2.
    \end{aligned}
\end{equation}
Rearranging \eqref{eqa:equiv_by_66_prop_5_1} 
leads to 
\begin{equation*}\label{eqa:equiv_before_68_prop_5_1}
    \begin{aligned}
      \|\bm{\mu}(\nu)\|_2
\le \mathcal{O}\big(\sqrt{\sum_{j=1}^{d+1}{\frac{\log n}{n}(s_j(\nu)+\widehat{s}_j(\nu))}}+{\lambda \sqrt{s_0 (d+1)} \over n} \big),
    \end{aligned}
\end{equation*}
for arbitrary $\nu \in \Upsilon$. 
Hence 
the proof of the \textbf{Main Statement} is completed. The rest of the proofs follow the similar arguments in Proposition 1 of \cite{shi2021testing}. For $\lambda$  satisfying the bound of $\mathcal{O}({(n\log{n} )}^{1/2})$, the consistency of the estimated matrix holds for arbitrary $\nu \in \Upsilon$. The proof of \Cref{theoremA} is hence completed.

    \section{Additional Simulation Results}\label{asec:simu}

In this section, we provide additional simulation configurations and results. 
\subsection{Simulation Configurations}
    \begin{table}[!htp]
  \caption{Hyper-parameters information.}\label{tab:train}
  \centering
  \begin{tabular}{ll}
    \toprule  
    Hyper-parameters     & Values   \\
    \midrule
    Maximum number of dual ascent steps in NOTEARS/NSCSL & 100\\
    Tolerance $\tau$ of acyclic constraint $h_1$ to be violated in NOTEARS/NSCSL & 1e-8 \\
    Maximum of parameters for the hard constraints in NOTEARS/NSCSL &1e+16\\
    $L_1$ penalty term $l$ in NOTEARS/NSCSL &   0  \\
         Conditional independent testing in PC  &   ``Fisher-Z''  \\
         Pruning threshold for all methods & 0.3\\
    \bottomrule
  \end{tabular}
\end{table}
\subsection{More Real Data Analyses on Yeast Data}\label{more_yeast}
\begin{figure}[!t]
\centering
\vspace{-0.3cm}
\begin{subfigure}[]
 \centering
 \includegraphics[width=0.45\textwidth]{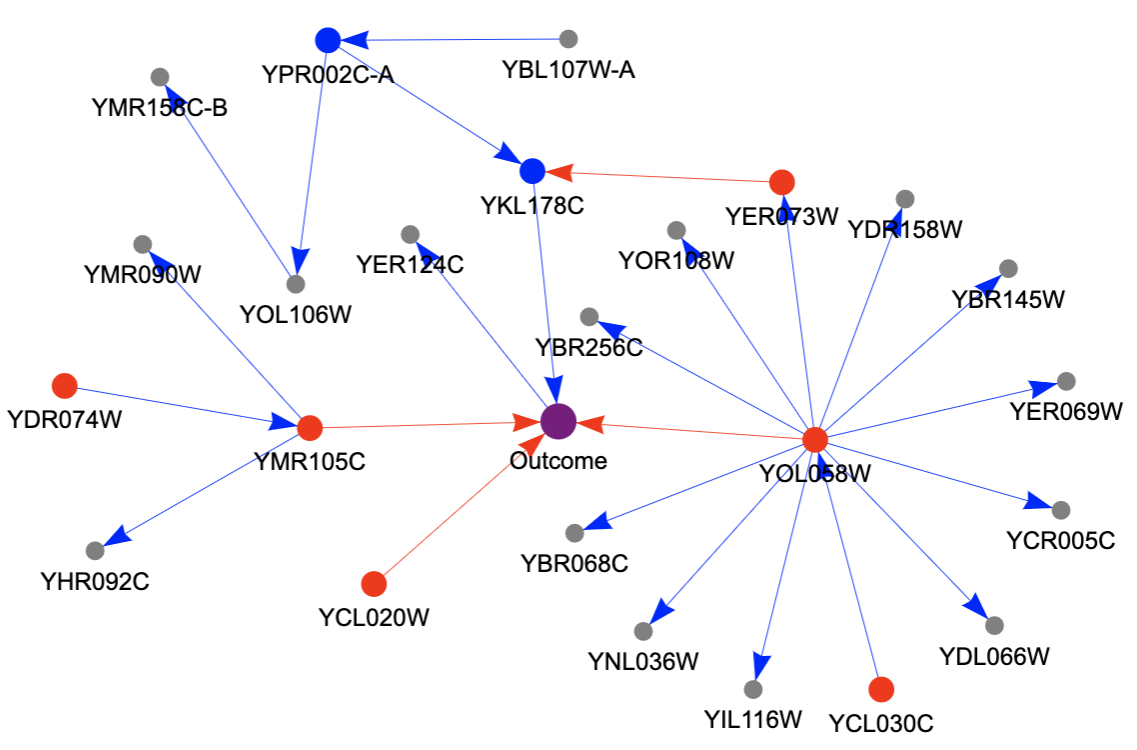} 
\end{subfigure}%
~~~~~~~~~~~~
\begin{subfigure}[]
 \centering
 \includegraphics[width=0.33\textwidth]{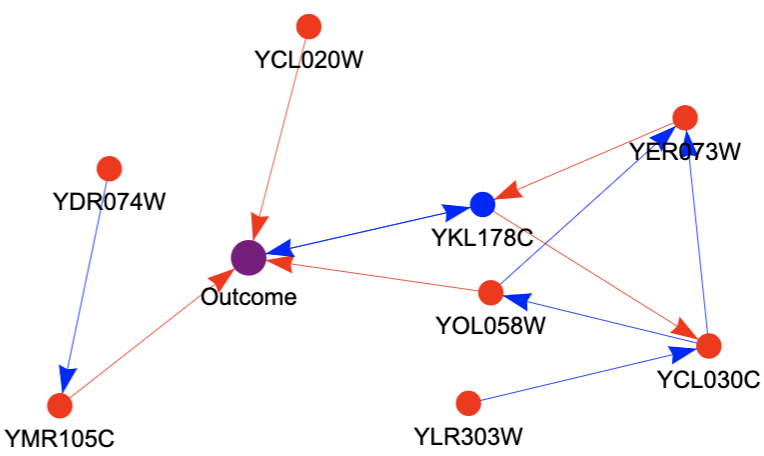} 
\end{subfigure} 
\vspace{-0.2cm}
\caption{Causal graphs for candidate genes that affect variant YER124C in yeast: (a). the estimated graph $\widehat{\mathcal{G}}$ by NOTEARS (benchmark); (b). the estimated graph $\widehat{\mathcal{G}}$ by NSCSL using TE.}
\label{fig_real_yeast}  
\vspace{-0.25cm}
\end{figure}

 The estimated causal graph among candidate QTLs and the outcome is shown in \Cref{fig_real_yeast} under the proposed method and
NOTEARS \citep{zheng2018dags} for illustration. 
The purple node represents the outcome, the blue nodes indicate QTLs with a positive causal impact, and the red nodes denote QTLs with a negative impact. Grey nodes are noisy QTLs without causal impact. Blue and red arrows represent positive and negative causal links, respectively. Causal effects from candidate genes on the genetic variant YER124C in yeast gene data are summarized using NSCSL in \Cref{tab_real}. 
\Cref{fig_real_yeast} demonstrates that the proposed algorithm can discover necessary and sufficient causal relationships with better performance compared to the current causal discovery benchmark. Specifically, all nodes with causal effects (either blue or red nodes in the causal graph) on the outcome are identified under NSCSL. Furthermore, the proposed algorithm identifies an additional gene, `YLR303W', which is not found in NOTEARS. Here, `YLR303W' is essential for sulfur amino acid synthesis \citep{brzywczy1993role}, with an estimated total causal effect of -0.06 on the target gene expression, as shown in \Cref{tab_real}. And `YER124C' of interest is a daughter cell-specific protein involved in cell wall metabolism \citep{colman2001yeast}. It has been shown that sulfur amino acid synthesis can influent cell wall metabolism \citep{takahashi2001sulfur,de2019regulation}. These indicate that NSCGL which additionally identified `YLR303W' performs better than NOTEARS. These observations align with findings from our simulation studies, further supporting NSCSL's superiority in revealing important causal features.
 
  \begin{table}[!t]
  \vspace{-0.3cm}
            \centering
            \caption{Summary of candidate genes that affect the variant YER124C in yeast gene data by NSCSL.}\label{tab_real} 
            \scalebox{0.8}{
\begin{tabular}{lllll}
\toprule
Gene Code  &Gene Function & Direct Effect  & Total Effect  \\ 
\midrule
YOL058W       &Arginosuccinate synthetase & -0.20                & -0.22      \\
\midrule
YCL020W   & Genotype regulators& -0.26          & -0.26       \\
\midrule
YDR074W  & Trehalose-6-phosphate phosphatase & 0.0     & -0.15       \\
\midrule
YMR105C    &Phosphoglucomutase   & -0.28               & -0.28       \\
\midrule
YKL178C  & Cell surface a factor receptor & 0.06         & 0.07    \\
\midrule
YLR303W & Required for sulfur amino acid synthesis & 0.0           & -0.06     \\
\midrule
YCL030C    &Multifunctional enzyme   & 0.0           &  -0.22    \\  
\midrule
YER073W        &Mitochondrial aldehyde dehydrogenase & 0.0            & -0.06           \\ 
\bottomrule  
\end{tabular}}
\end{table}

 \clearpage
     \subsection{Additional Simulation Results:  True and Estimated Matrix}
     
 \begin{figure}[!thp]
\centering
\begin{subfigure}[]
  \centering
  \includegraphics[width=0.45\linewidth]{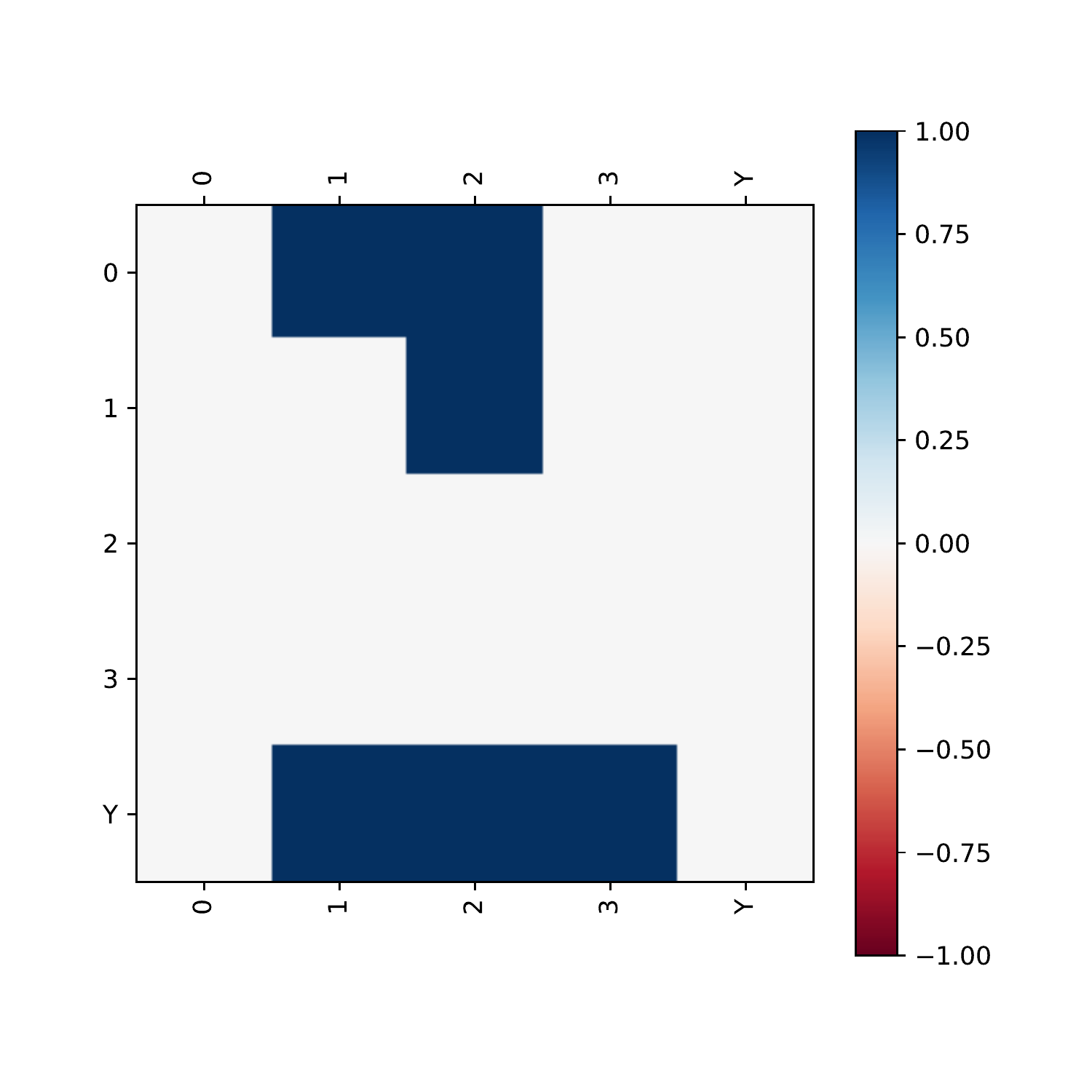} 
\end{subfigure}%
\begin{subfigure}[]
  \centering
  \includegraphics[width=0.45\linewidth]{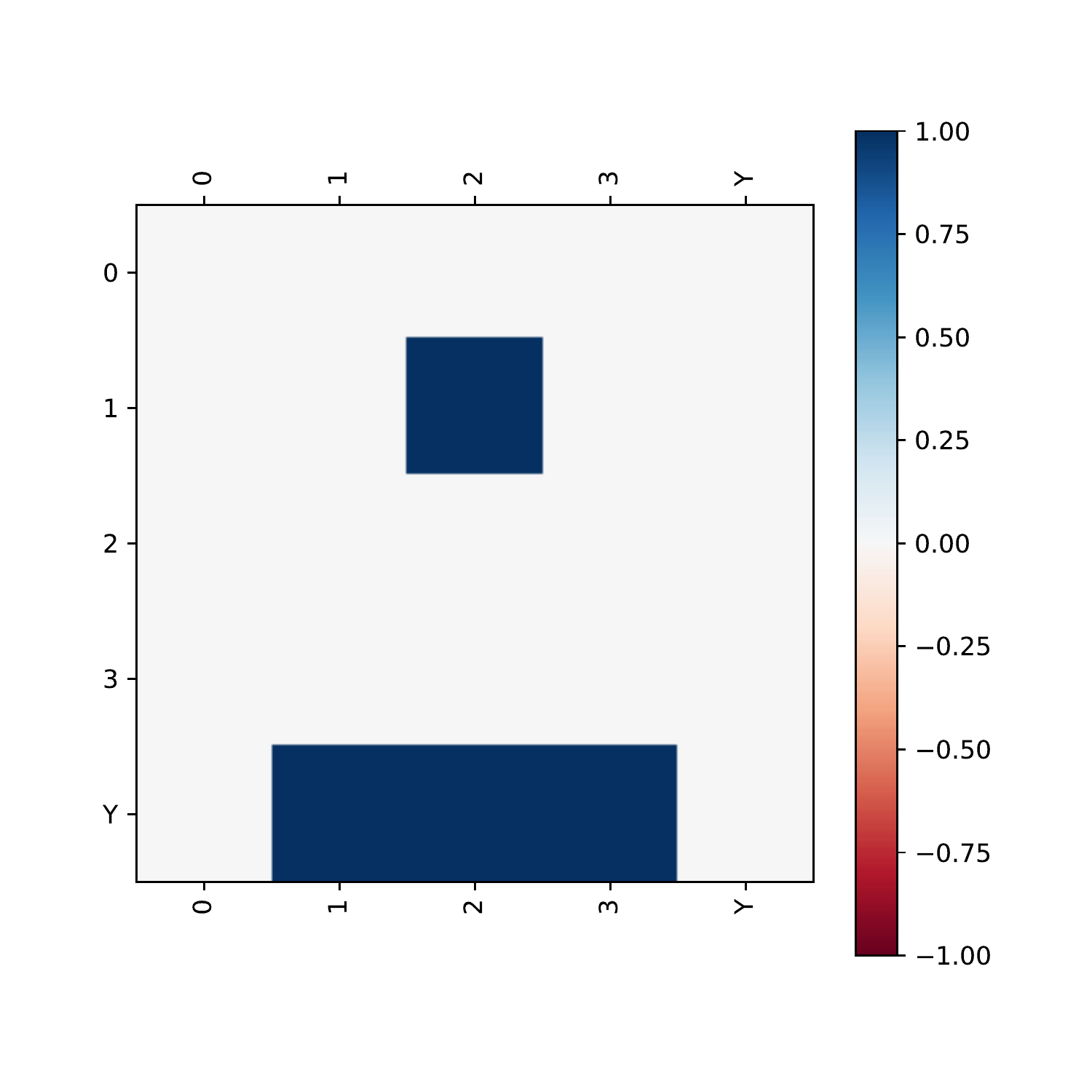} 
\end{subfigure}\\
\begin{subfigure}[] 
  \centering
  \includegraphics[width=0.45\linewidth]{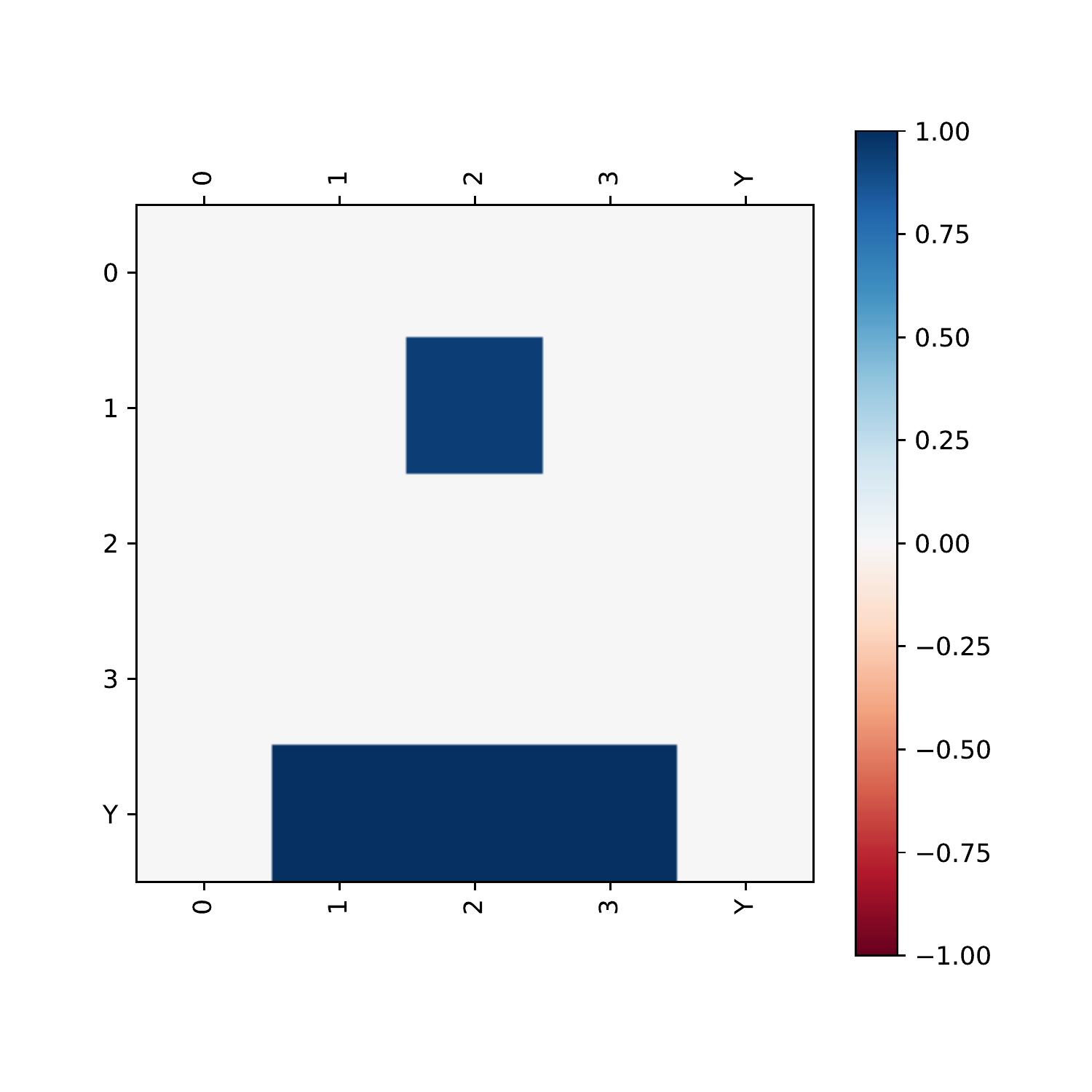} 
\end{subfigure}%
\begin{subfigure}[] 
  \centering
  \includegraphics[width=0.45\linewidth]{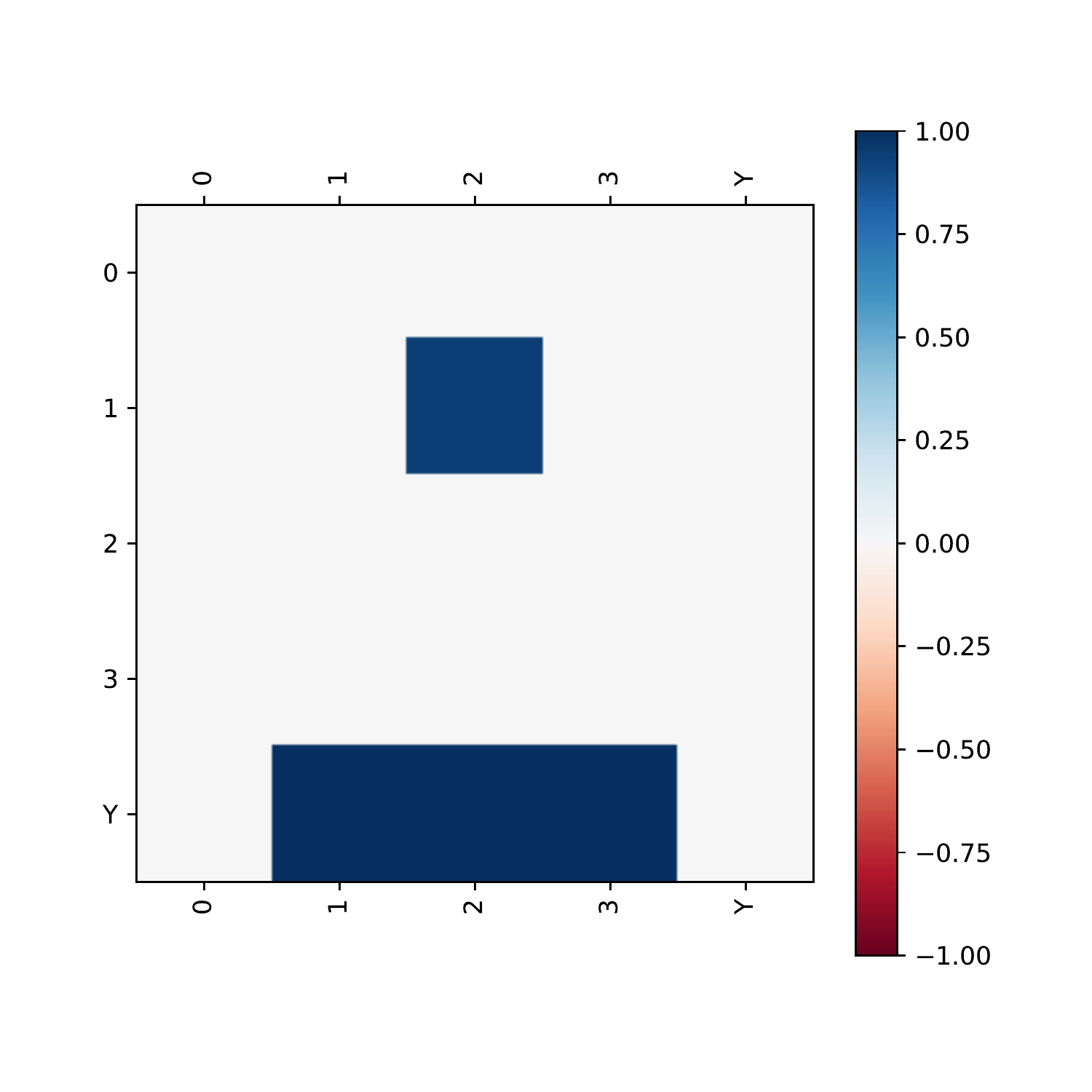} 
\end{subfigure}%
\begin{subfigure}[] 
  \centering
  \includegraphics[width=0.31\linewidth]{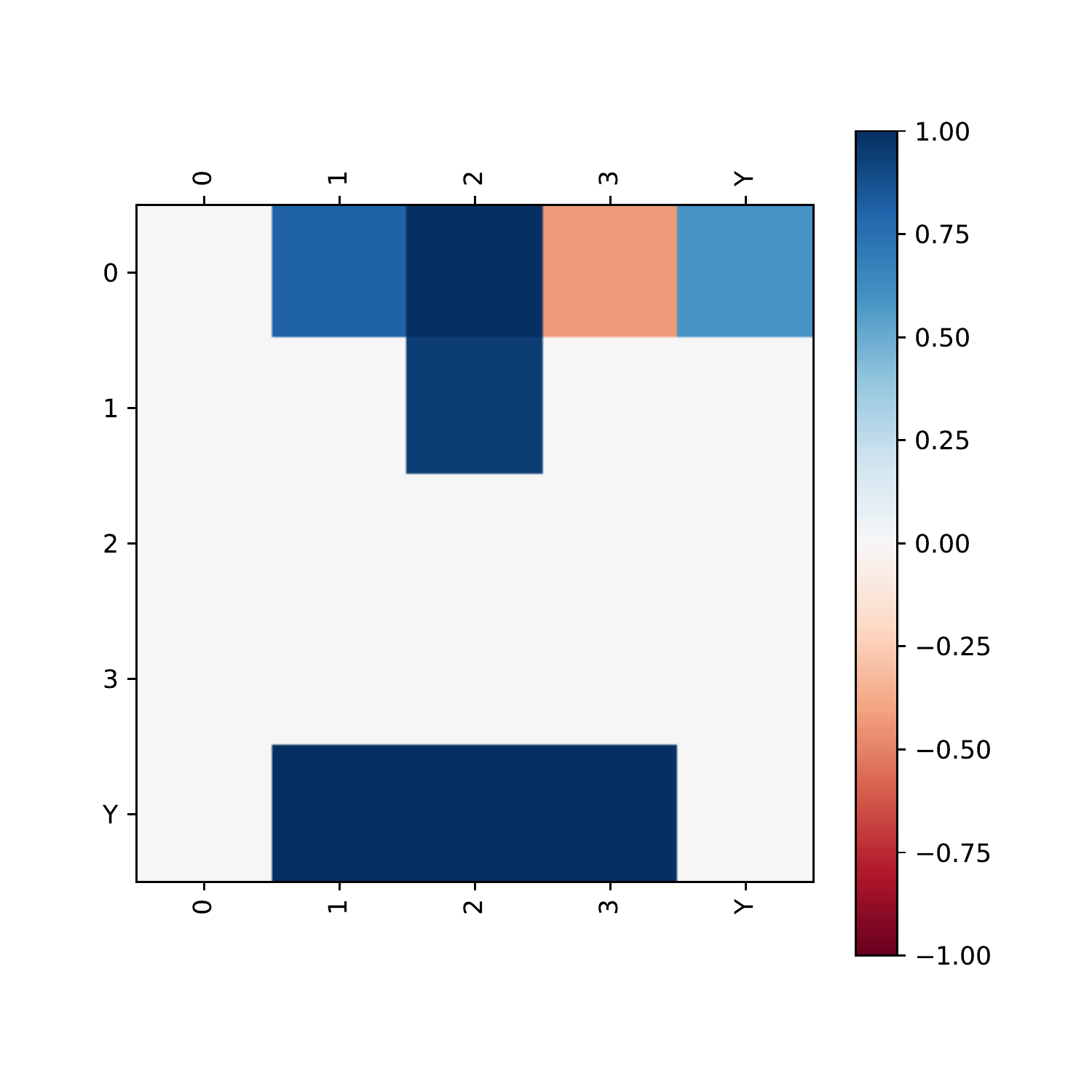} 
\end{subfigure}
\begin{subfigure}[] 
  \centering
  \includegraphics[width=0.31\linewidth]{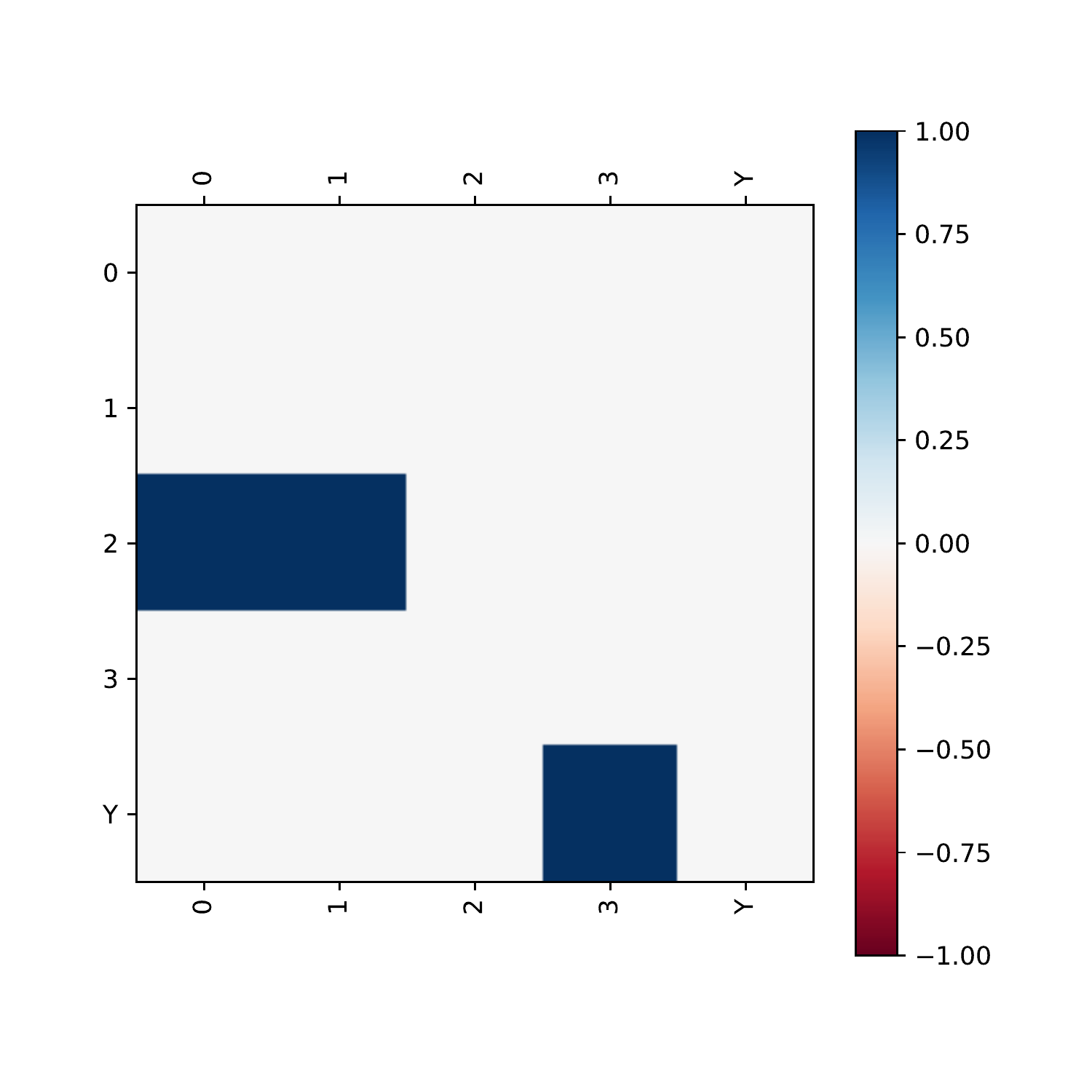} 
\end{subfigure}
\begin{subfigure}[] 
  \centering
  \includegraphics[width=0.31\linewidth]{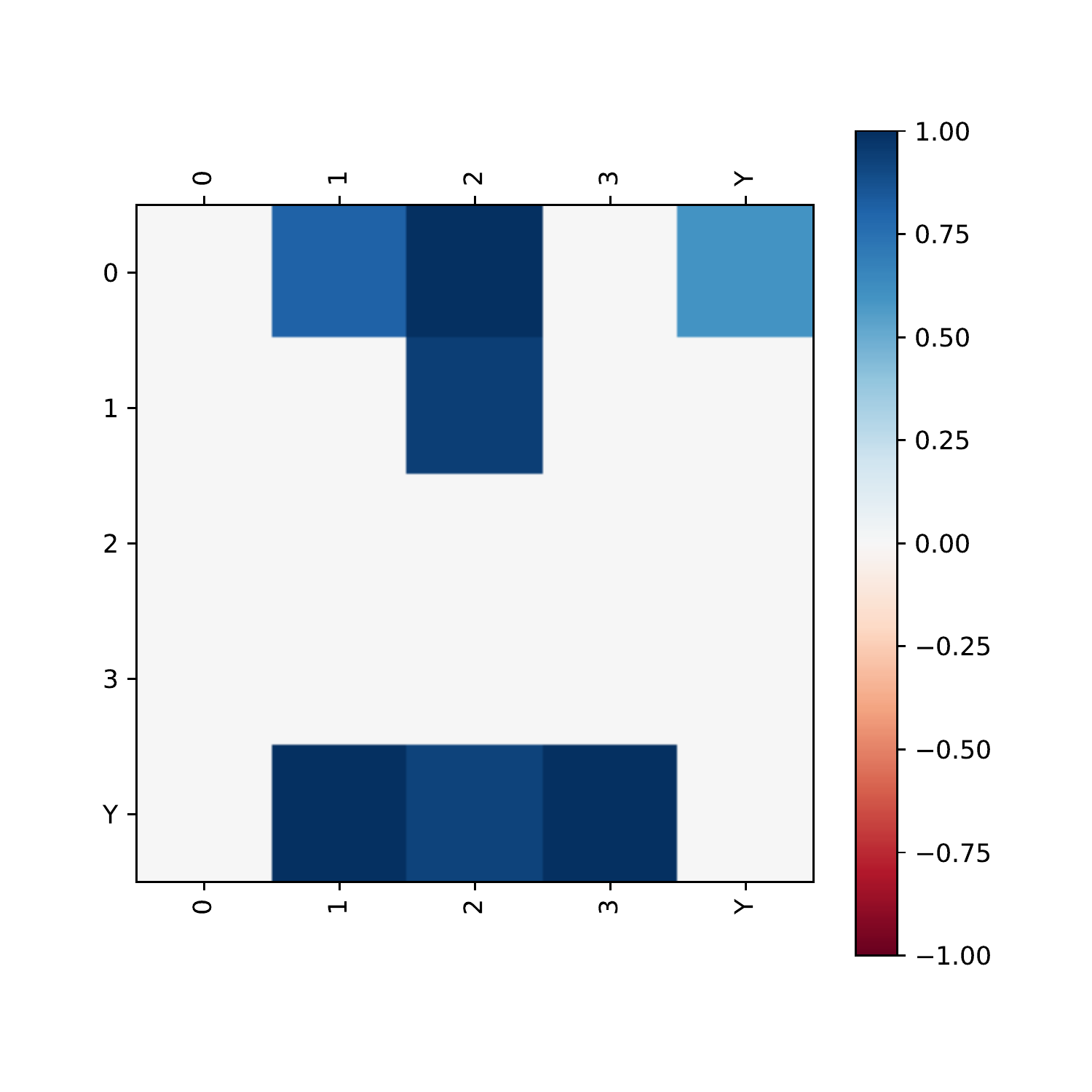}
\end{subfigure}
 \vspace{-0.4cm}
\caption{Estimated matrix under S1 ($n=20$): (a). true whole graph; (b). true NSCG; (c). $\widehat{\mathcal{G}}$ by NSCSL with TE; (d). $\widehat{\mathcal{G}}$ by NSCSL with DE; (e).  $\widehat{\mathcal{G}}$ by NOTEARS; (f). $\widehat{\mathcal{G}}$ by PC; (g). $\widehat{\mathcal{G}}$ by LiNGAM.}
\label{fig_scen_res2}  
 \vspace{-0.25cm}
 \end{figure}

 \begin{figure} 
\centering
\begin{subfigure}[]
  \centering
  \includegraphics[width=0.45\linewidth]{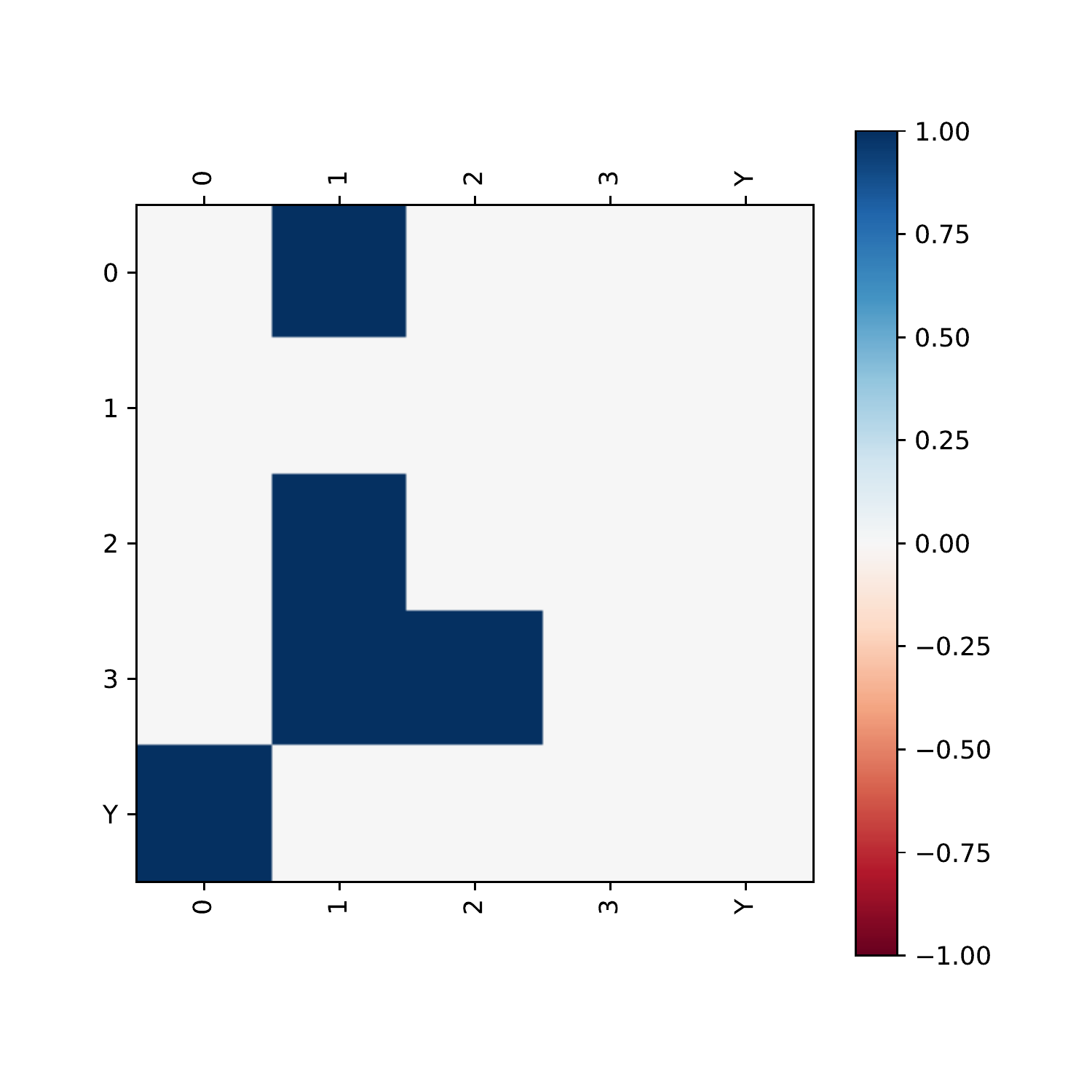} 
\end{subfigure}%
\begin{subfigure}[]
  \centering
  \includegraphics[width=0.45\linewidth]{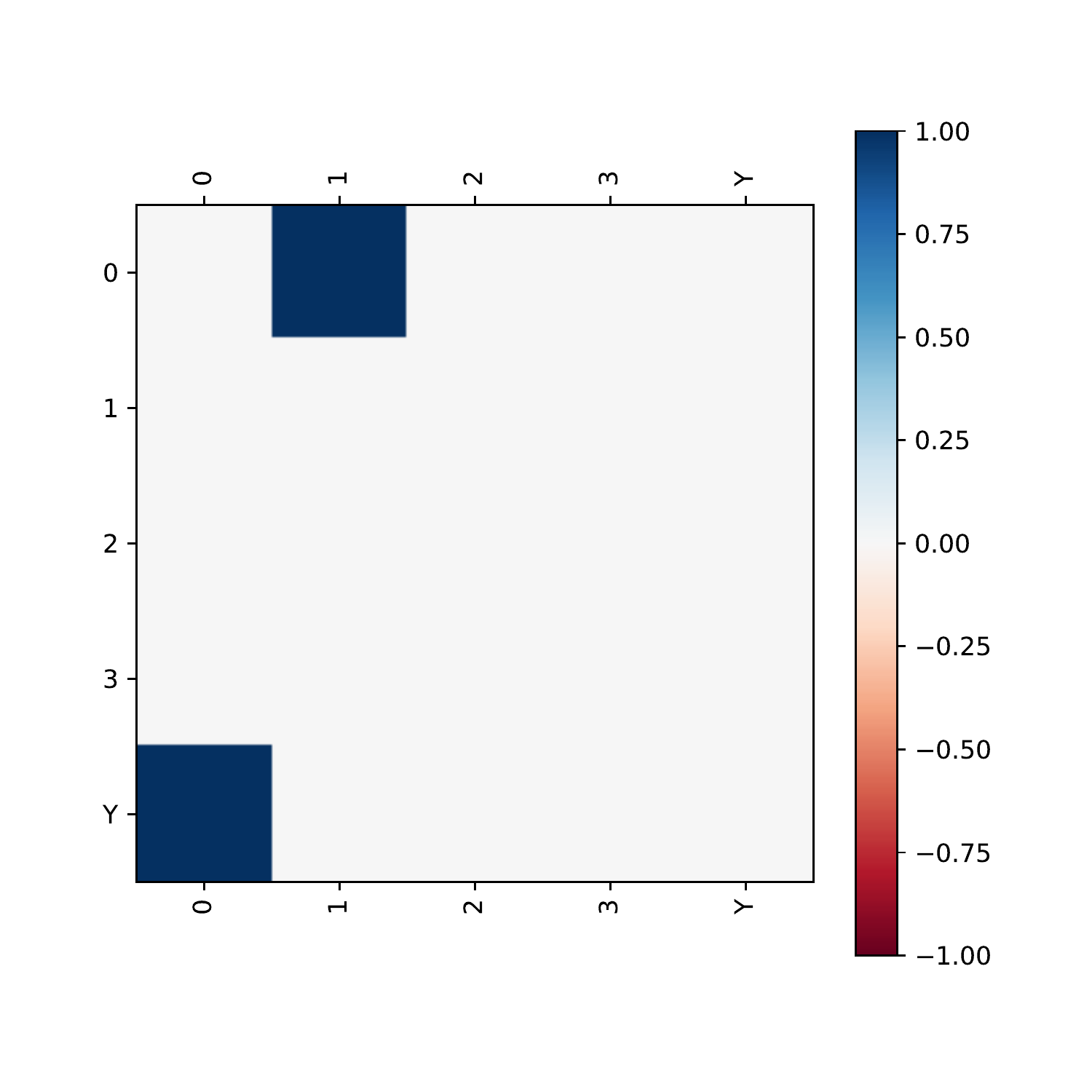} 
\end{subfigure}\\
\begin{subfigure}[] 
  \centering
  \includegraphics[width=0.45\linewidth]{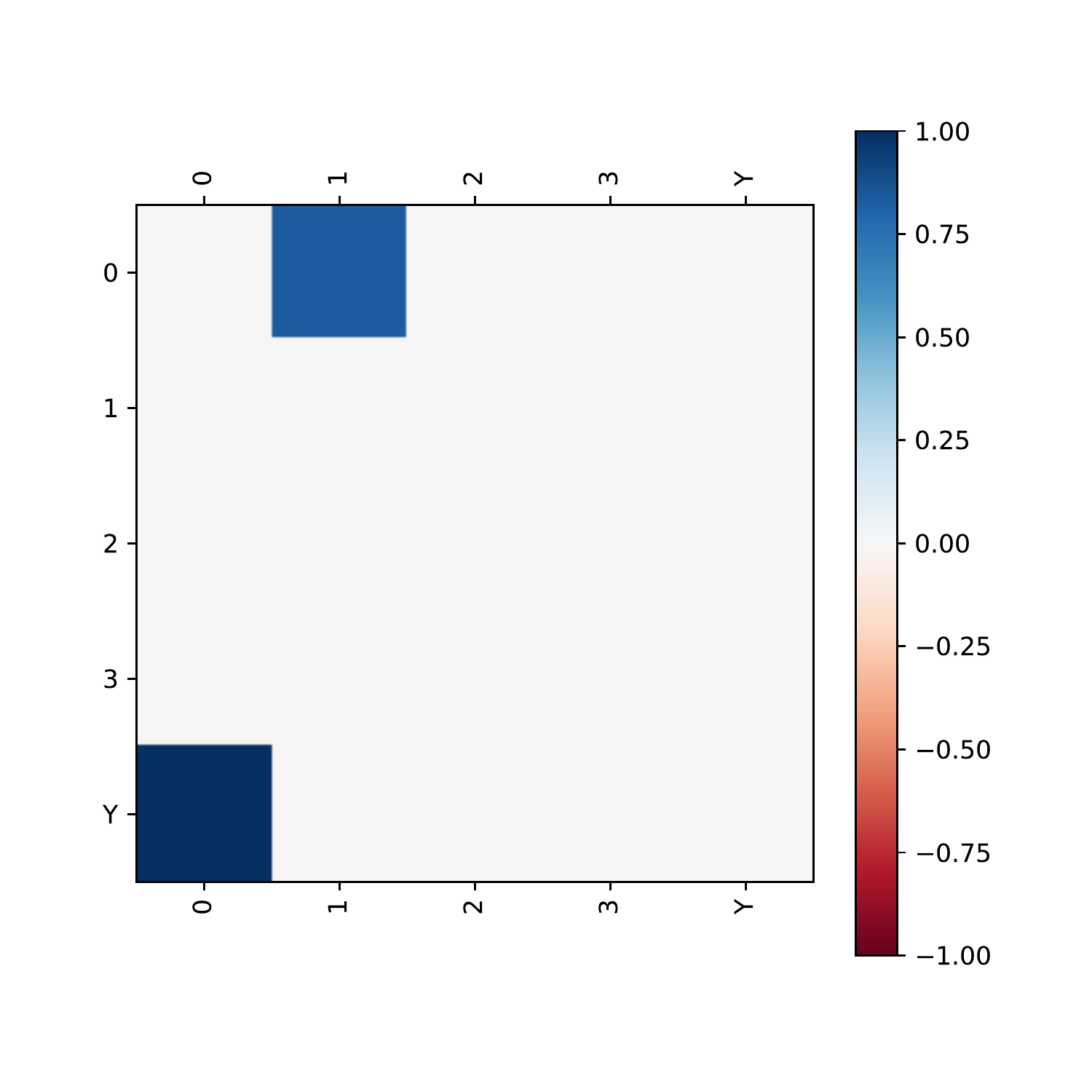} 
\end{subfigure}%
\begin{subfigure}[] 
  \centering
  \includegraphics[width=0.45\linewidth]{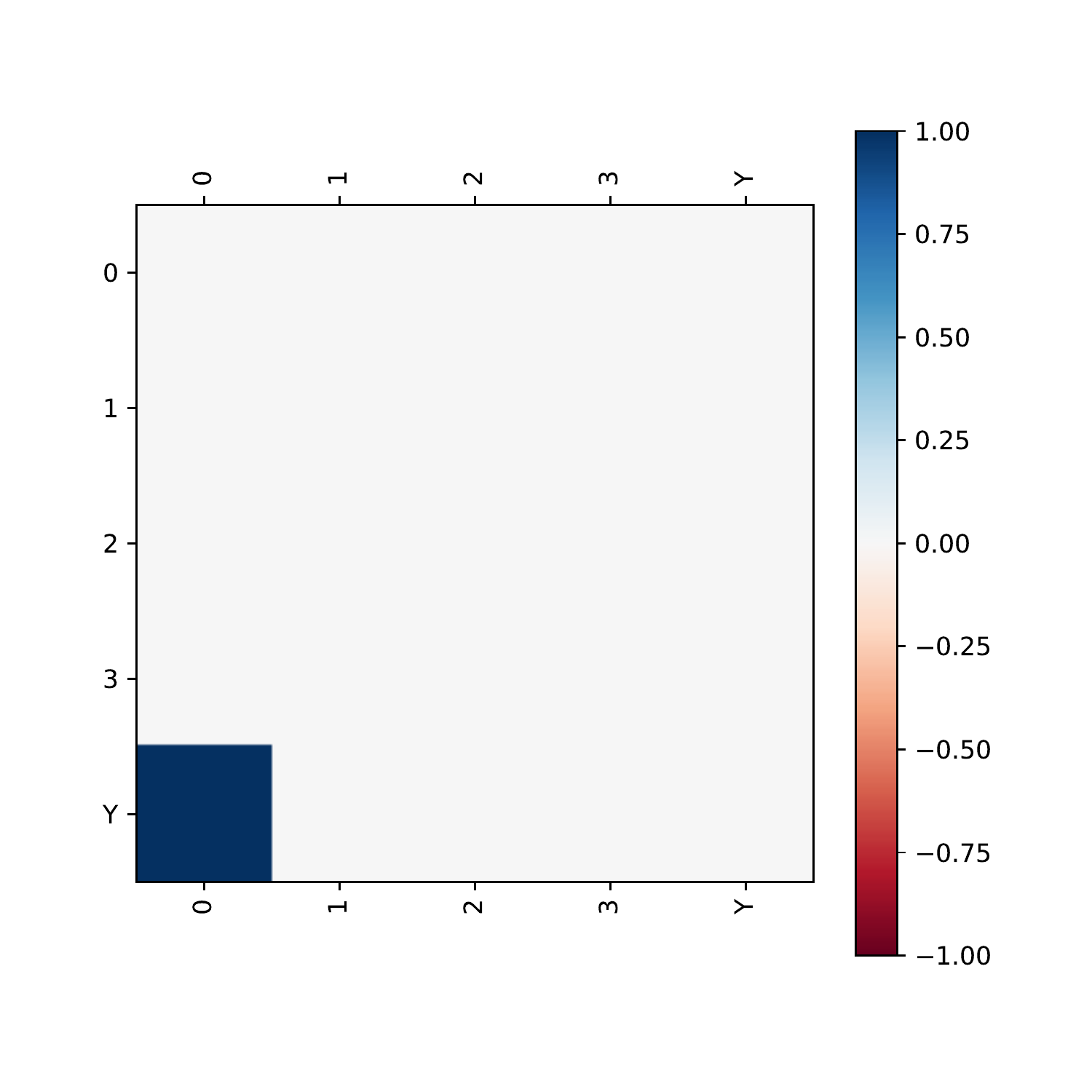} 
\end{subfigure}%
\begin{subfigure}[] 
  \centering
  \includegraphics[width=0.31\linewidth]{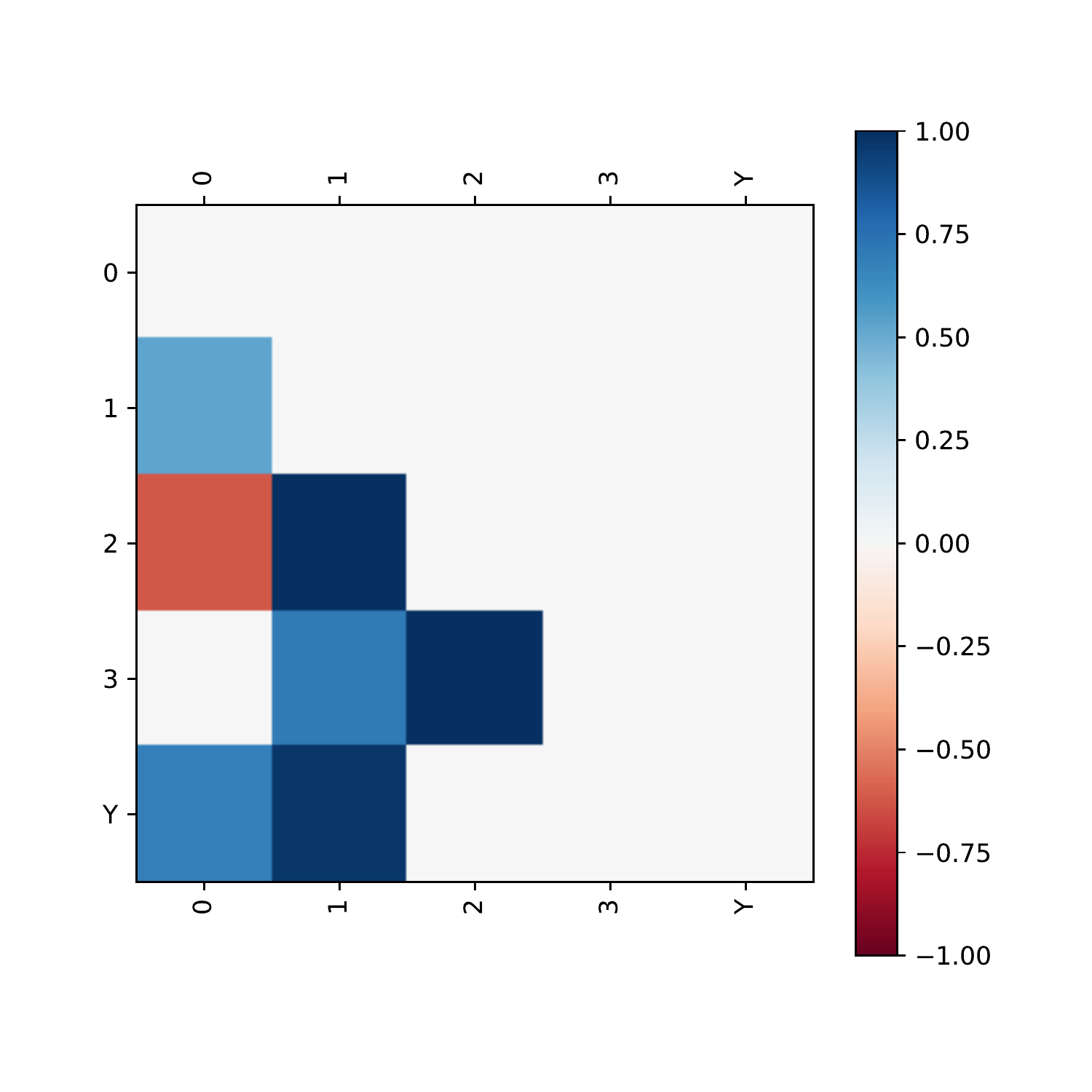} 
\end{subfigure}
\begin{subfigure}[] 
  \centering
  \includegraphics[width=0.31\linewidth]{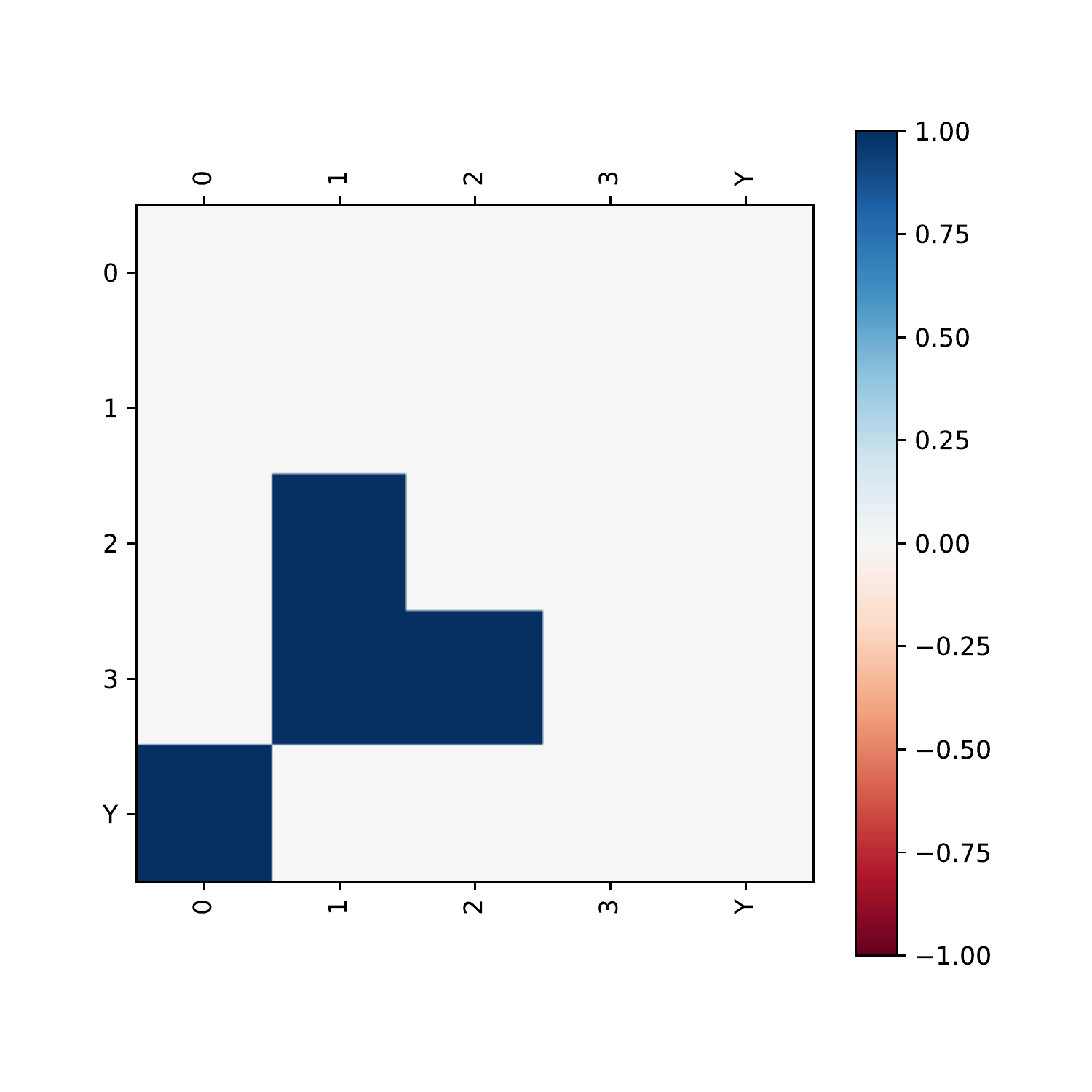} 
\end{subfigure}
\begin{subfigure}[] 
  \centering
  \includegraphics[width=0.31\linewidth]{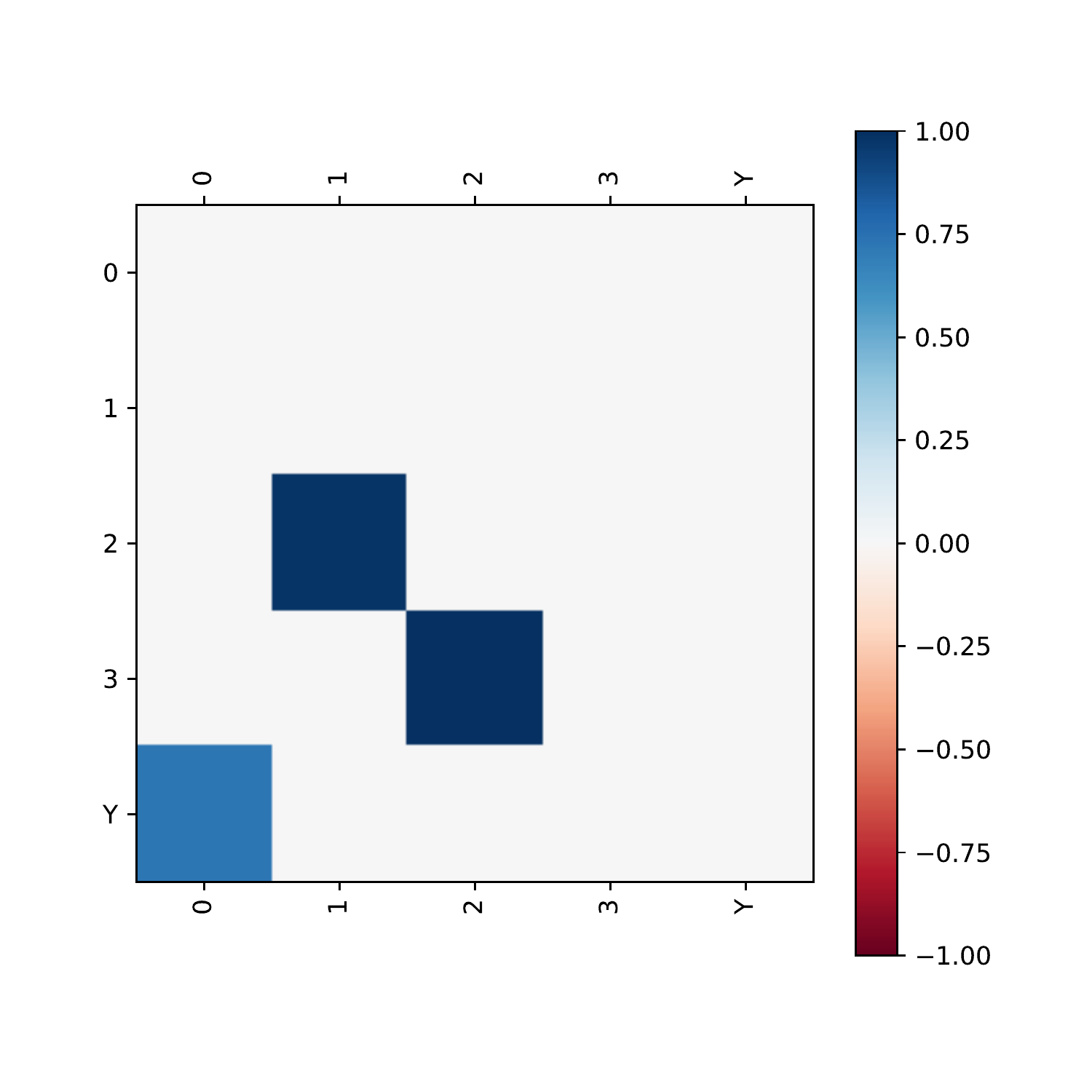}
\end{subfigure}
 \vspace{-0.35cm}
\caption{Estimated matrix under S2 ($n=20$): (a). true whole graph; (b). true NSCG; (c). $\widehat{\mathcal{G}}$ by NSCSL with TE; (d). $\widehat{\mathcal{G}}$ by NSCSL with DE; (e).  $\widehat{\mathcal{G}}$ by NOTEARS; (f). $\widehat{\mathcal{G}}$ by PC; (g). $\widehat{\mathcal{G}}$ by LiNGAM.}
\label{fig_scen_res4}  
 \vspace{-0.4cm}
 \end{figure}

 \begin{figure}[!t]
\centering
\begin{subfigure}[]
  \centering
  \includegraphics[width=0.45\linewidth]{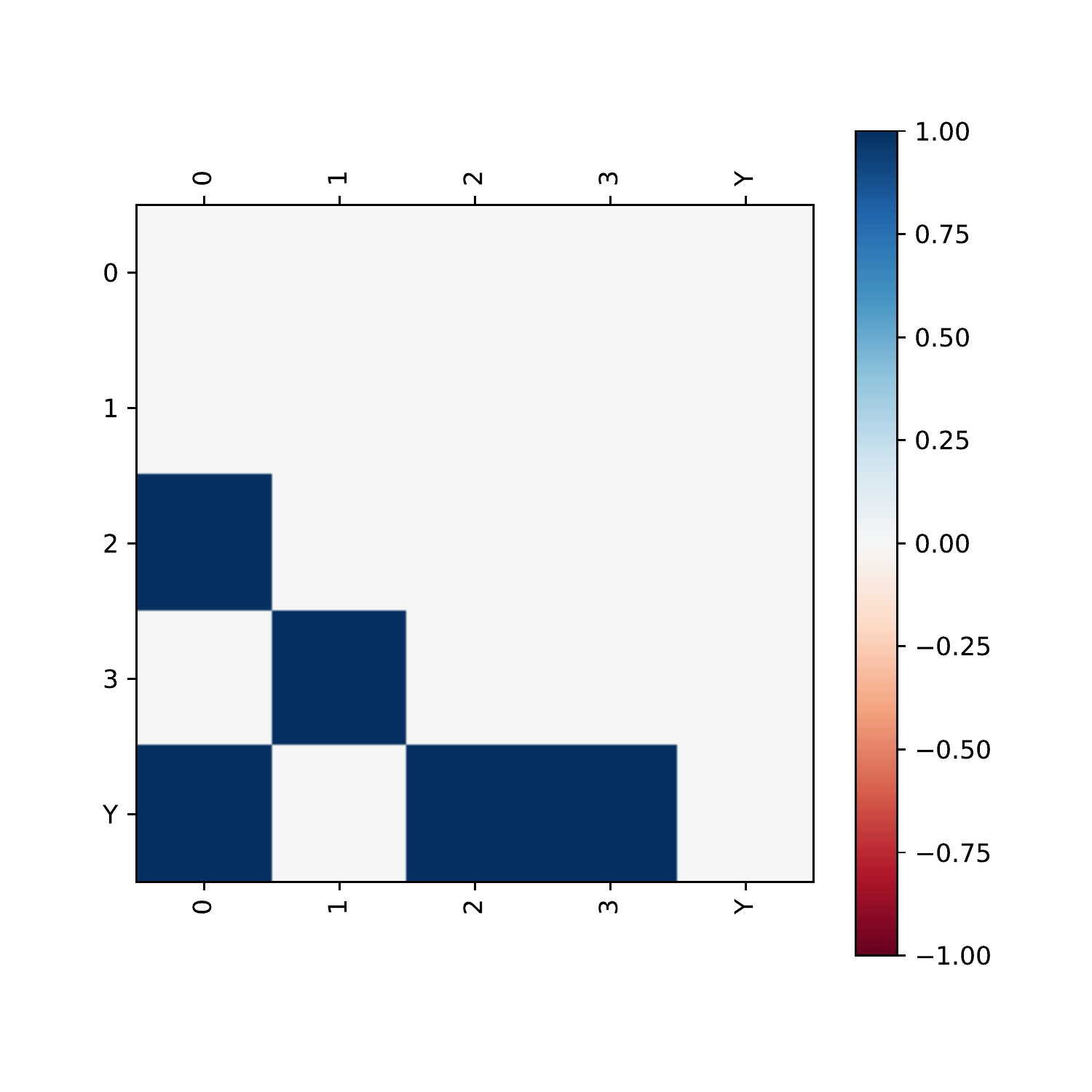} 
\end{subfigure}%
\begin{subfigure}[]
  \centering
  \includegraphics[width=0.45\linewidth]{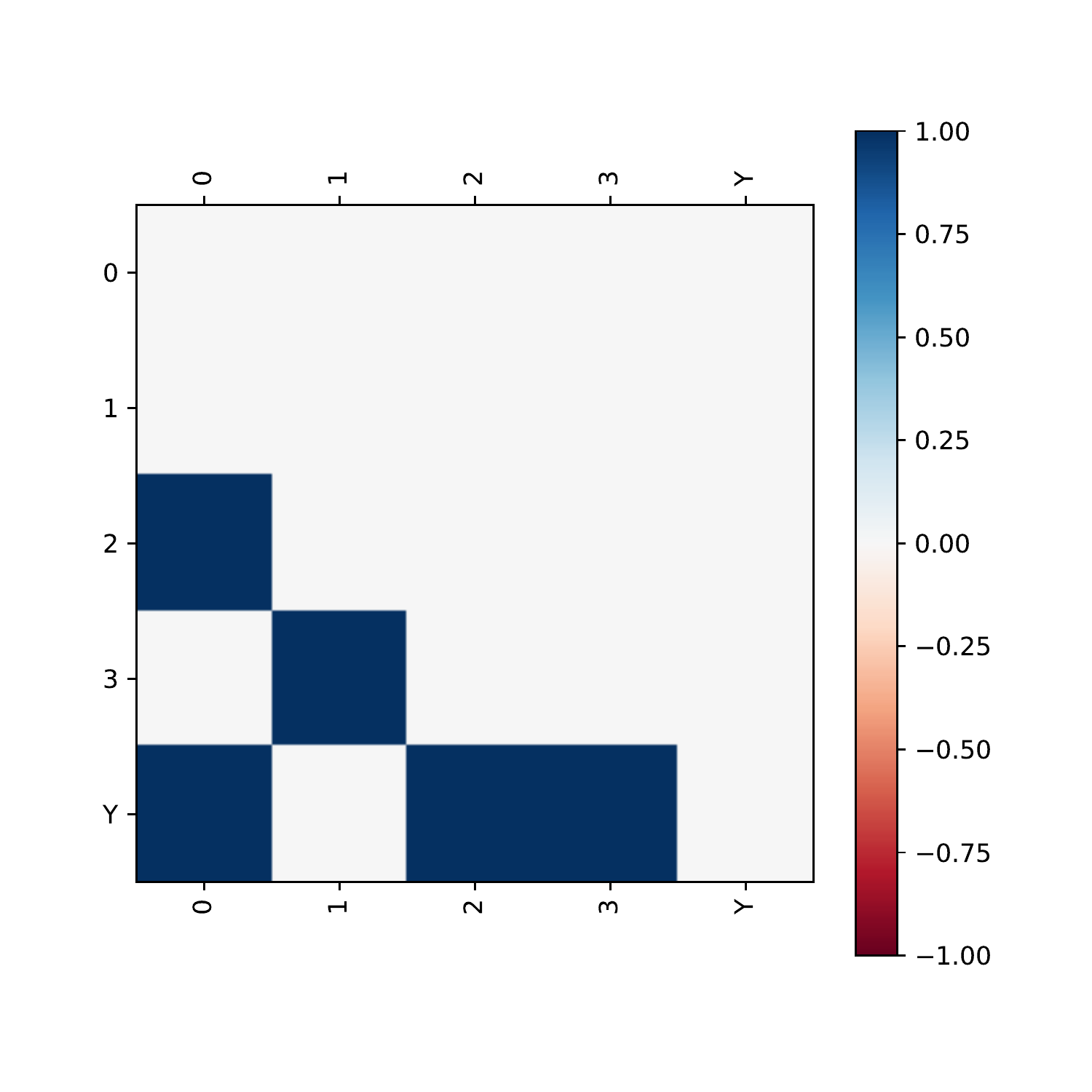} 
\end{subfigure}\\
\begin{subfigure}[] 
  \centering
  \includegraphics[width=0.45\linewidth]{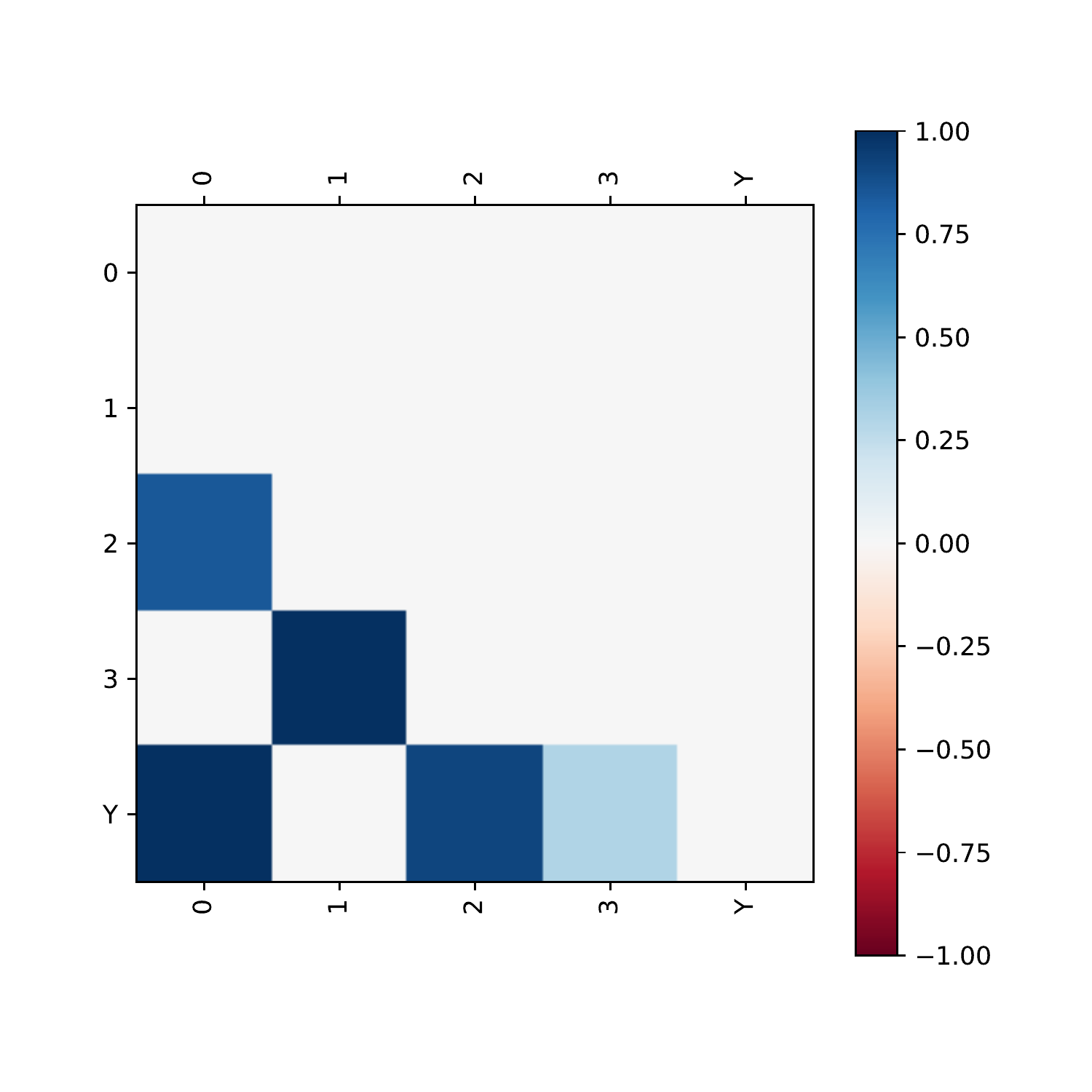} 
\end{subfigure}%
\begin{subfigure}[] 
  \centering
  \includegraphics[width=0.45\linewidth]{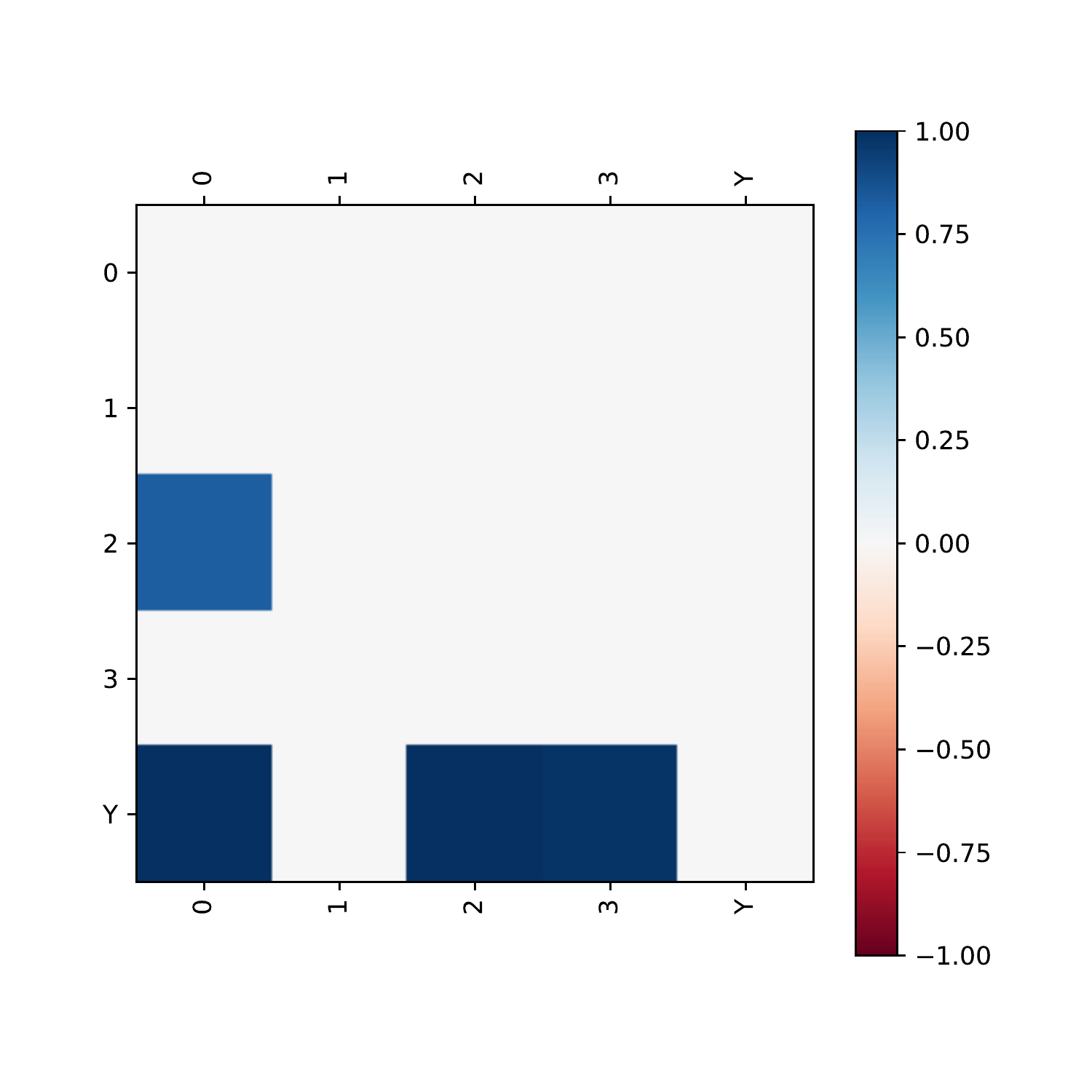} 
\end{subfigure}%
\begin{subfigure}[] 
  \centering
  \includegraphics[width=0.31\linewidth]{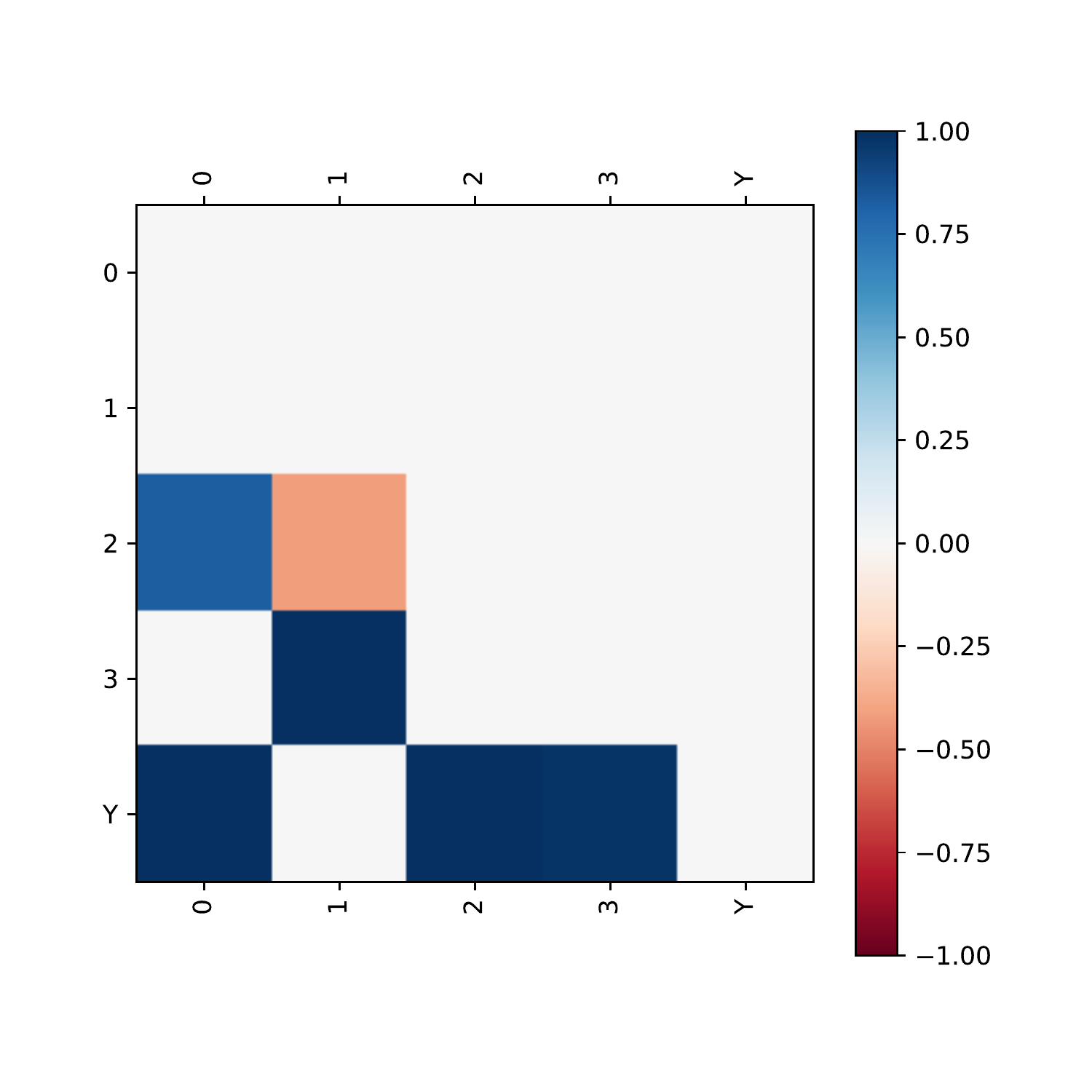} 
\end{subfigure}
\begin{subfigure}[] 
  \centering
  \includegraphics[width=0.31\linewidth]{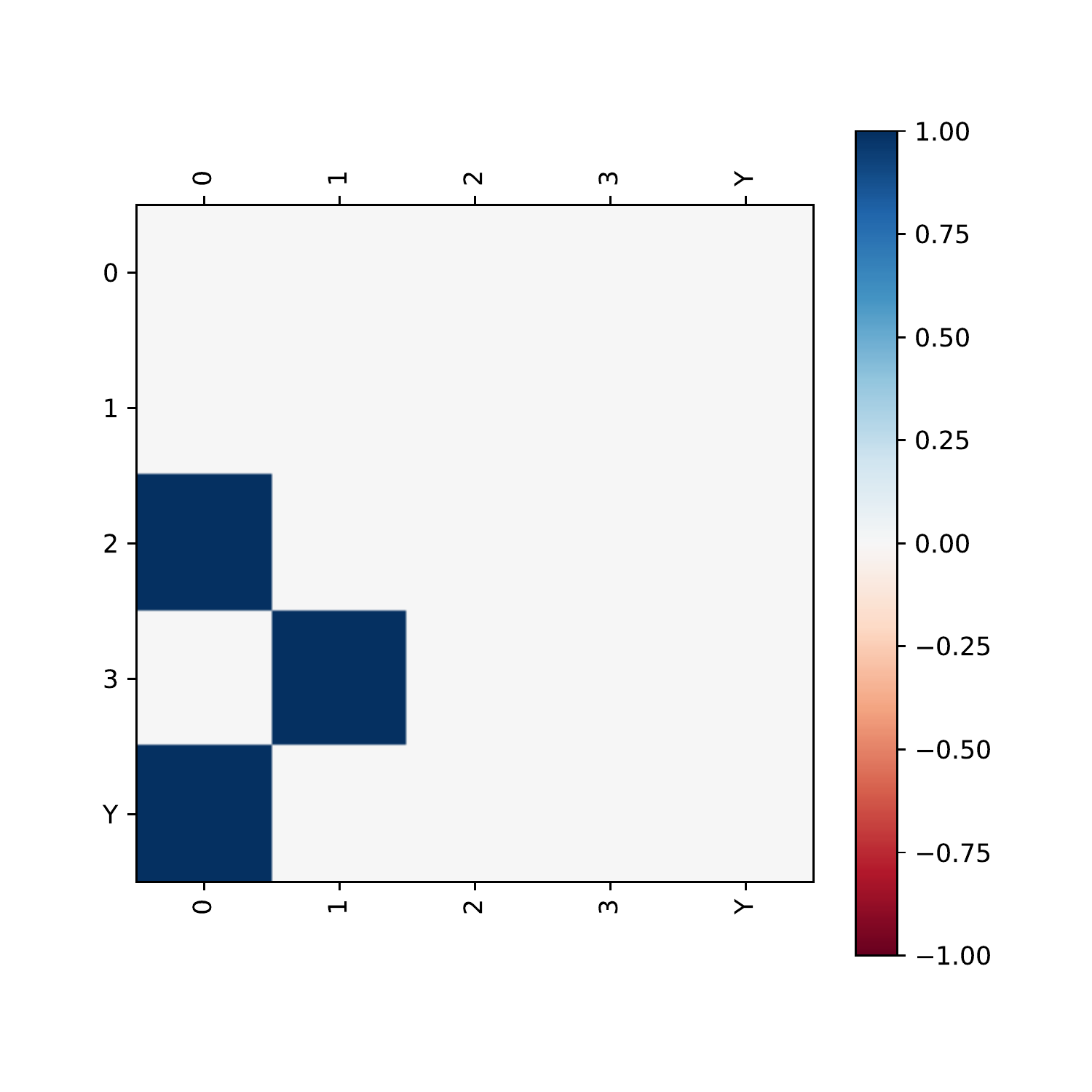} 
\end{subfigure}
\begin{subfigure}[] 
  \centering
  \includegraphics[width=0.31\linewidth]{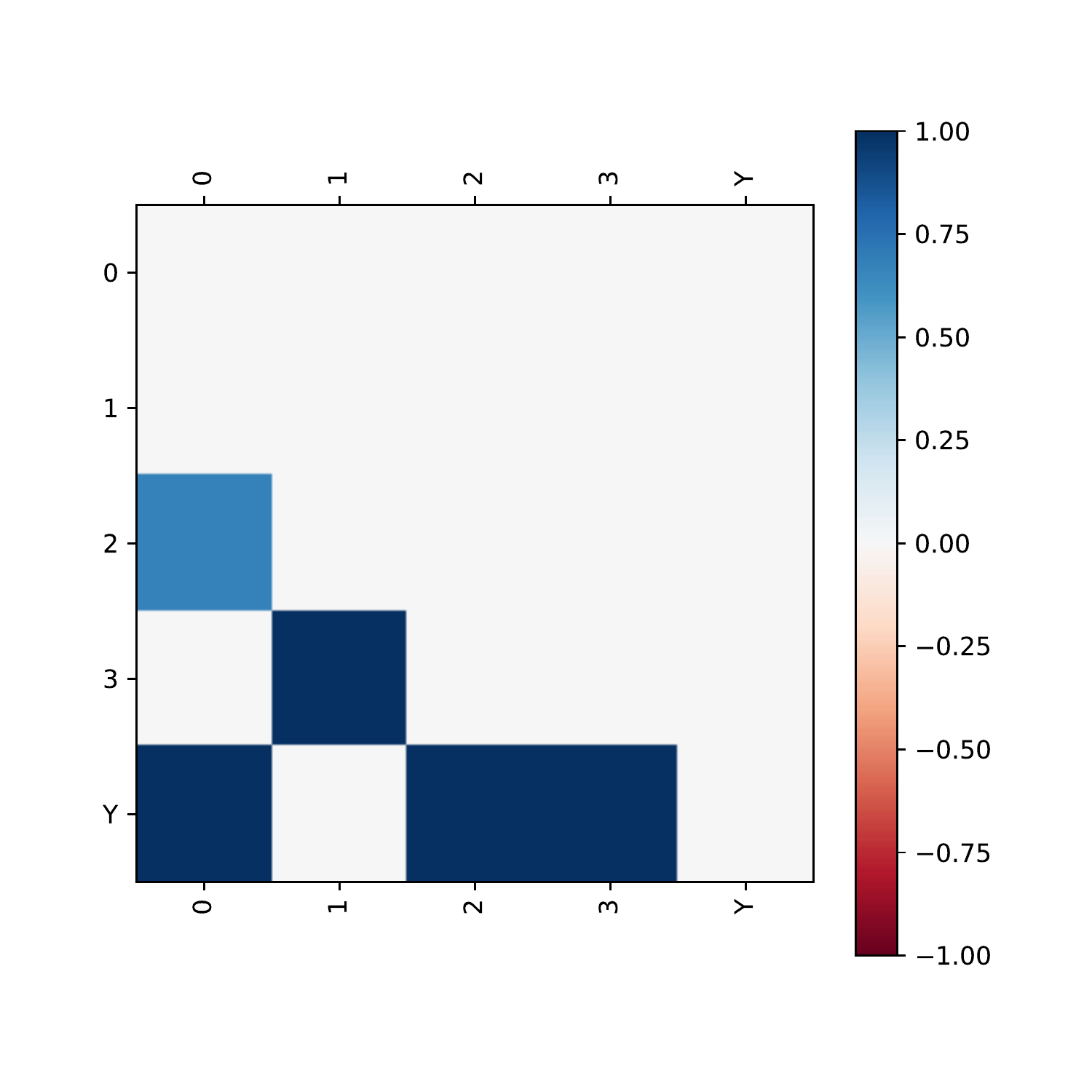}
\end{subfigure}
 \vspace{-0.35cm}
\caption{Estimated matrix under S3 ($n=20$): (a). true whole graph; (b). true NSCG; (c). $\widehat{\mathcal{G}}$ by NSCSL with TE; (d). $\widehat{\mathcal{G}}$ by NSCSL with DE; (e).  $\widehat{\mathcal{G}}$ by NOTEARS; (f). $\widehat{\mathcal{G}}$ by PC; (g). $\widehat{\mathcal{G}}$ by LiNGAM.}
\label{fig_scen_res01}  
 \vspace{-0.25cm}
 \end{figure}

 \begin{figure} 
\centering
\begin{subfigure}[]
  \centering
  \includegraphics[width=0.45\textwidth]{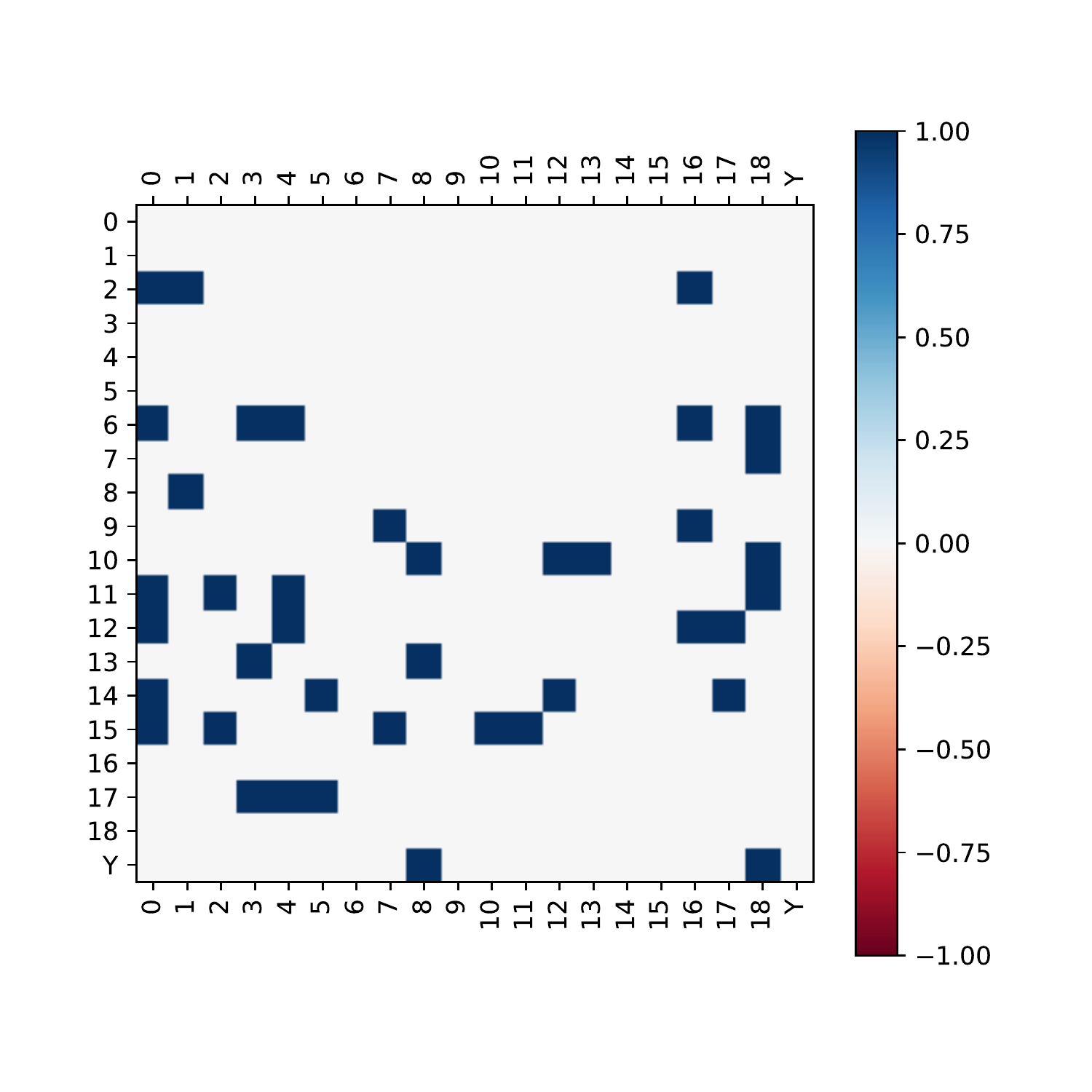} 
\end{subfigure}%
\begin{subfigure}[]
  \centering
  \includegraphics[width=0.45\textwidth]{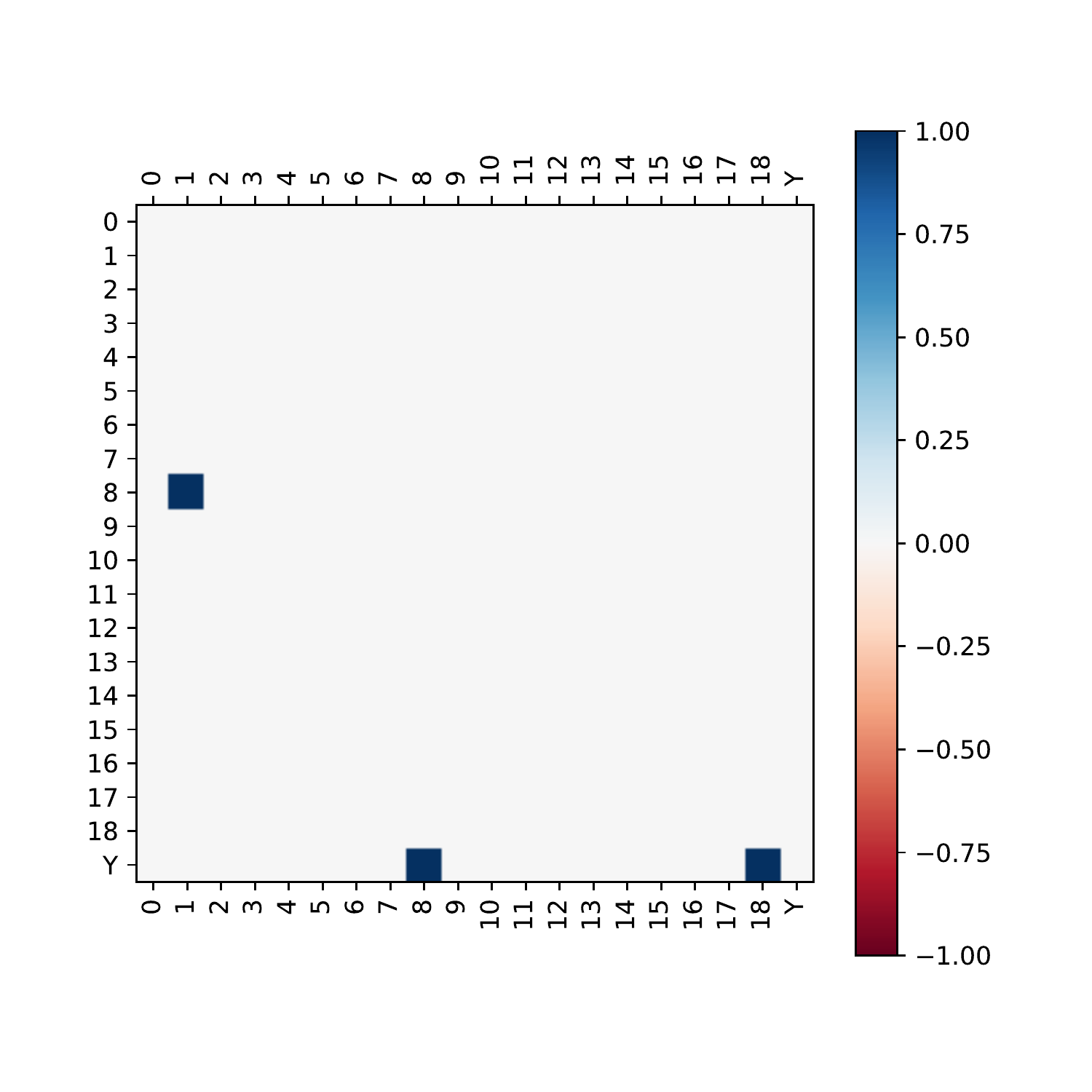} 
\end{subfigure}\\
\begin{subfigure}[] 
  \centering
  \includegraphics[width=0.45\textwidth]{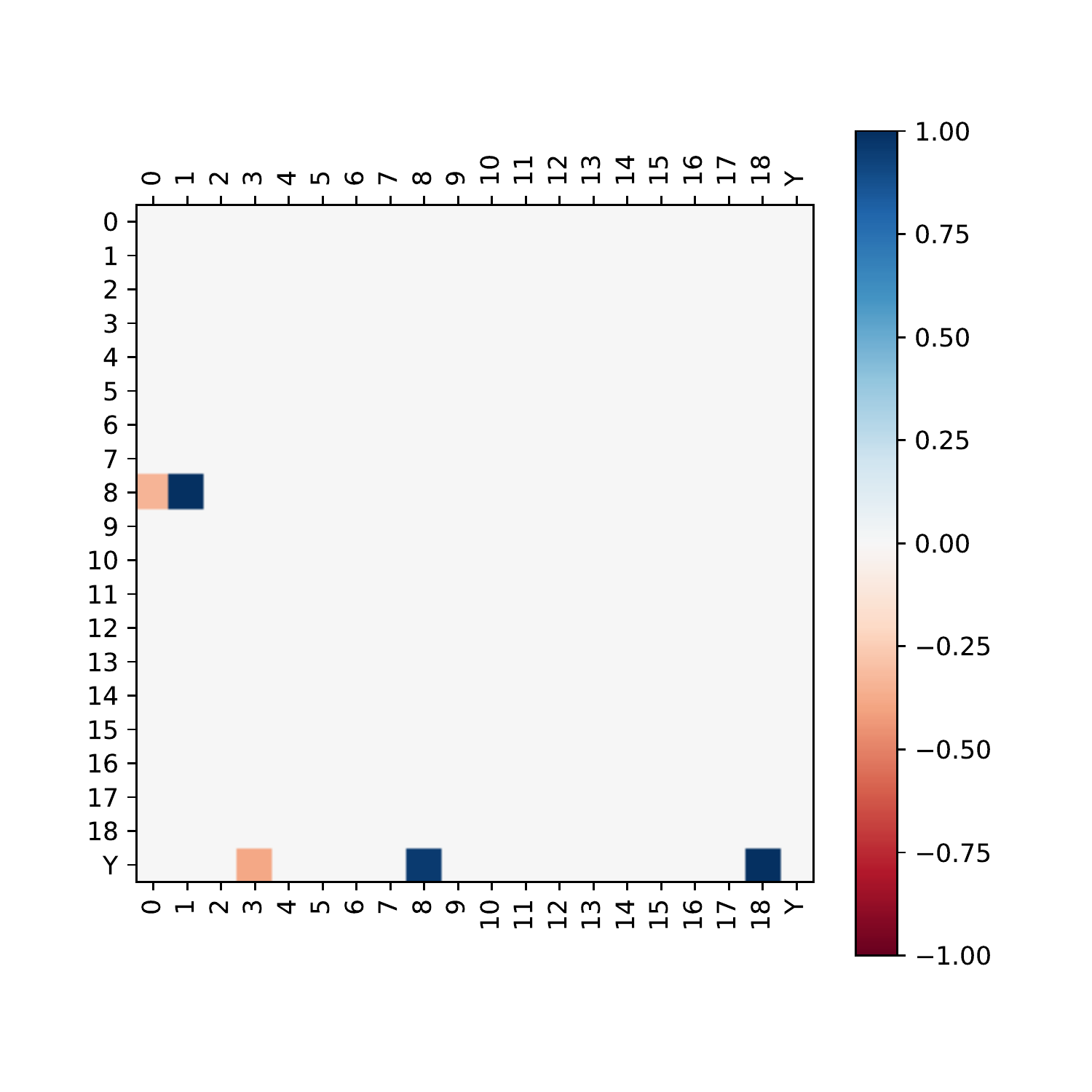} 
\end{subfigure}%
\begin{subfigure}[] 
  \centering
  \includegraphics[width=0.45\textwidth]{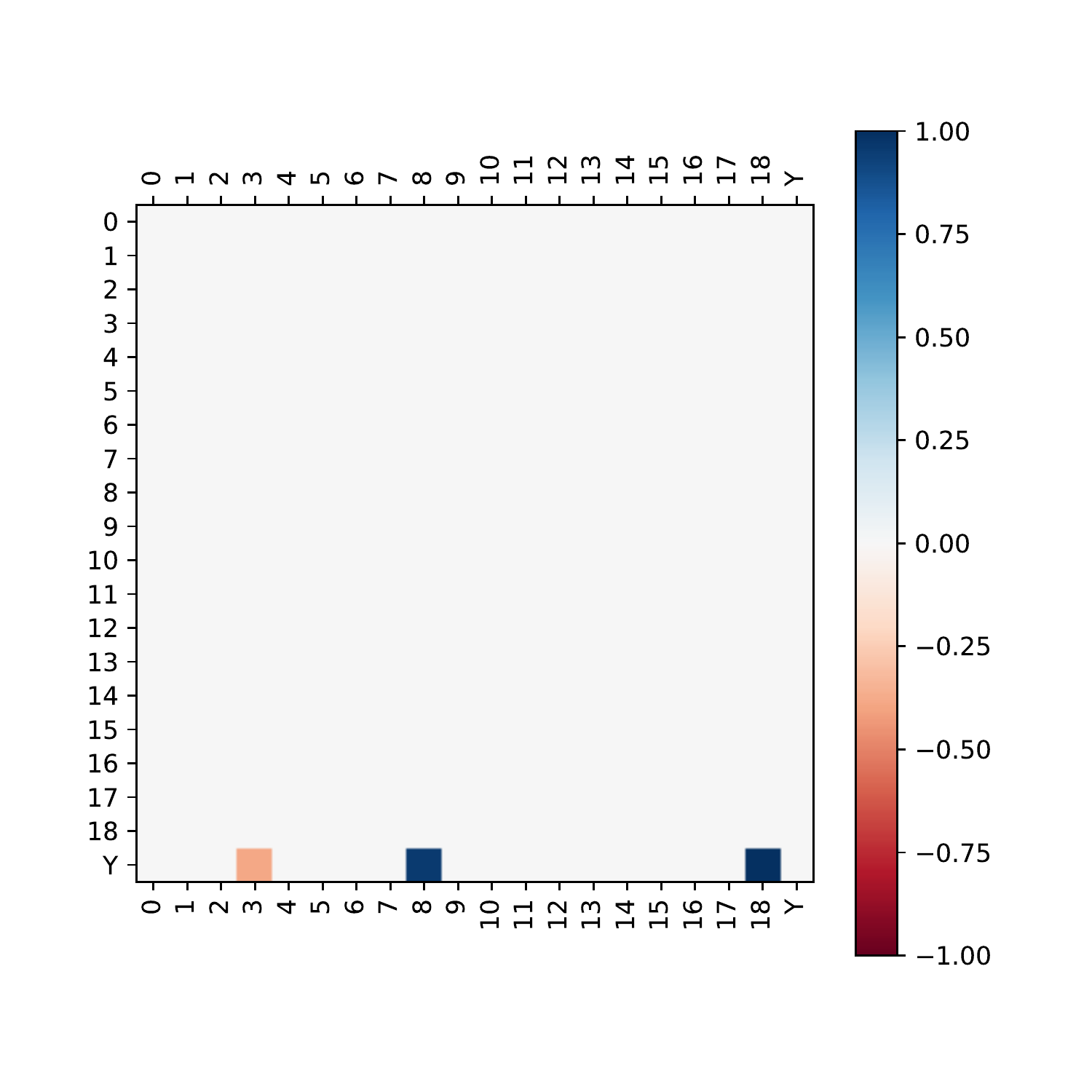} 
\end{subfigure}%
\begin{subfigure}[] 
  \centering
  \includegraphics[width=0.31\linewidth]{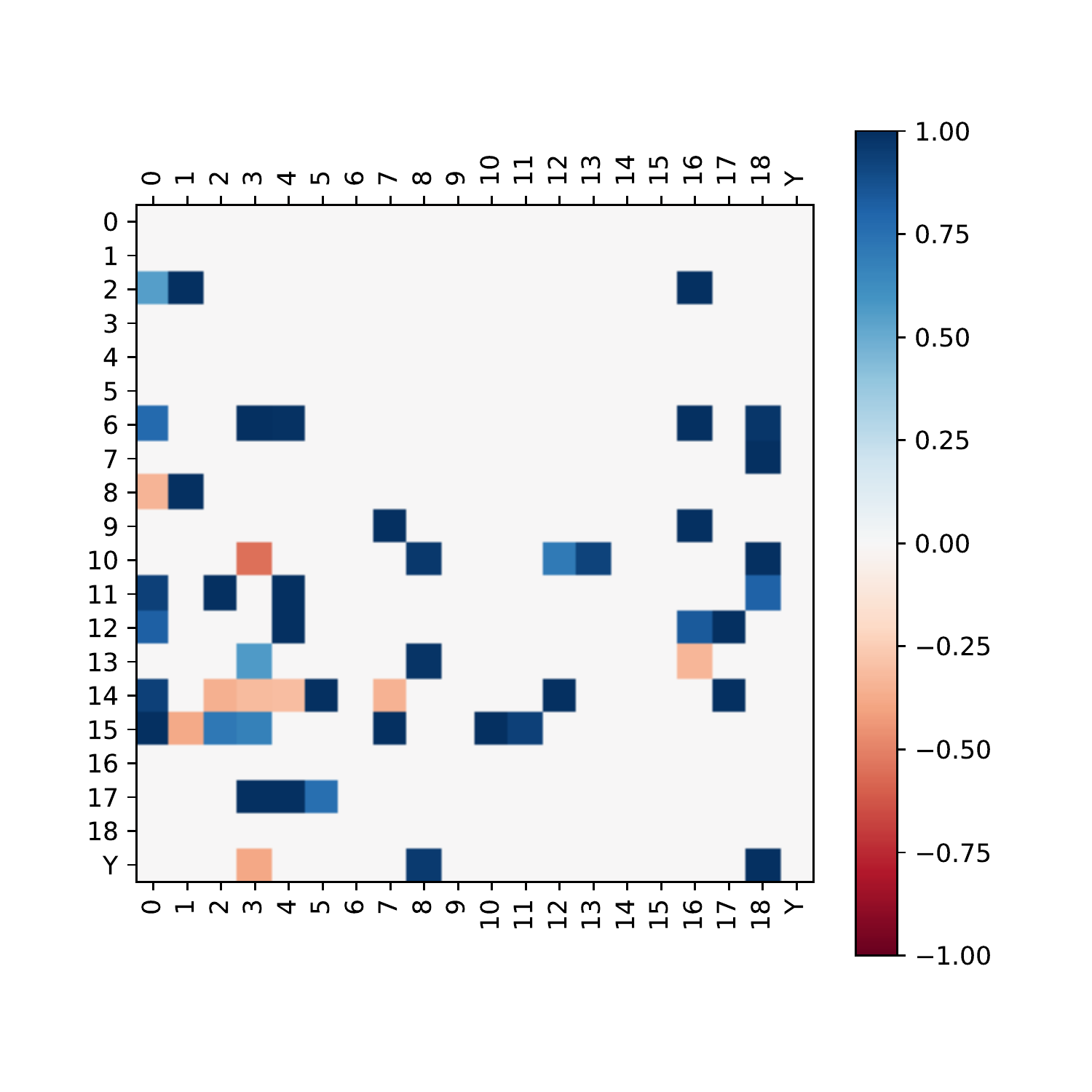} 
\end{subfigure}
\begin{subfigure}[] 
  \centering
  \includegraphics[width=0.31\linewidth]{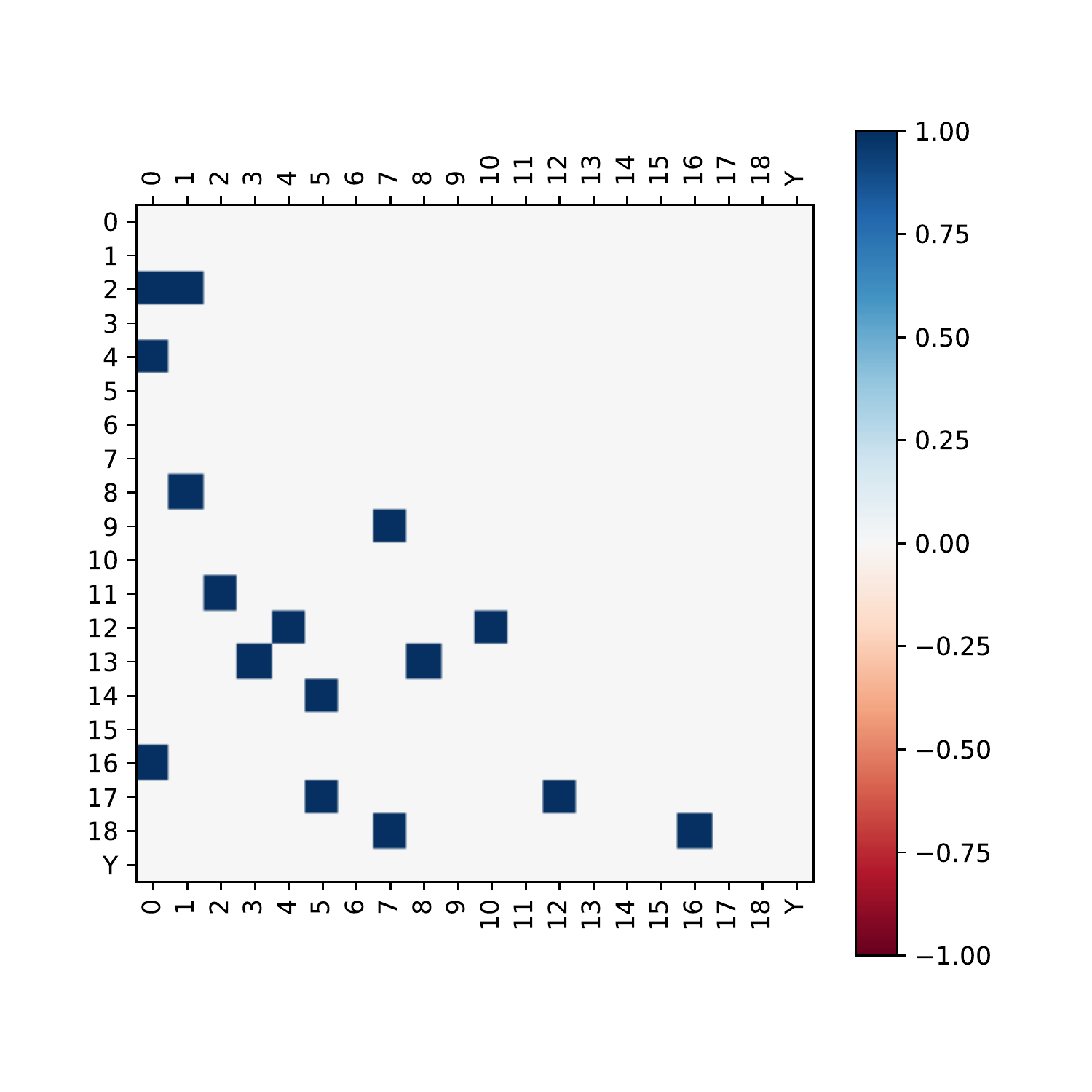} 
\end{subfigure}
\begin{subfigure}[] 
  \centering
  \includegraphics[width=0.31\linewidth]{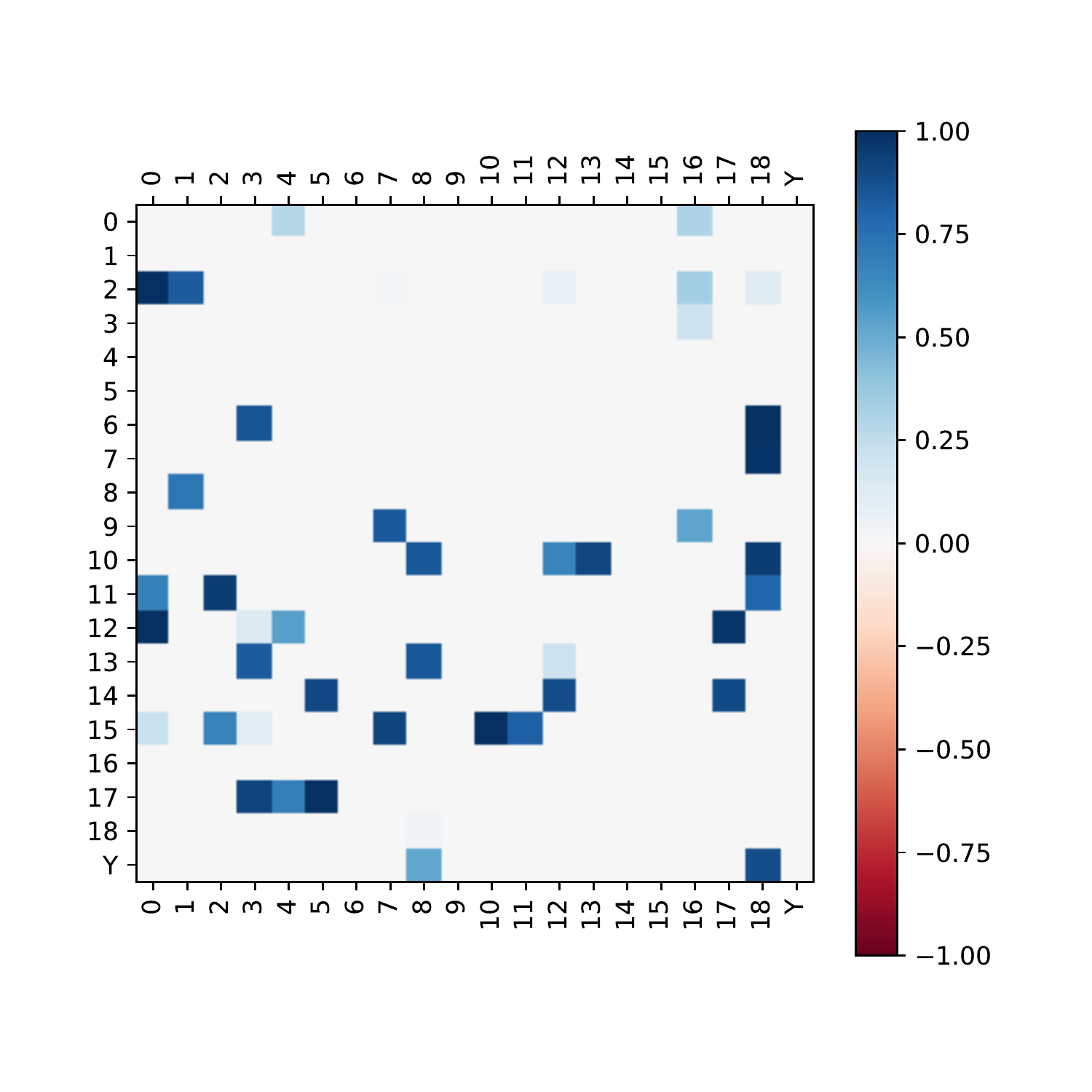}
\end{subfigure}
 \vspace{-0.35cm}
\caption{Estimated matrix  under S4 ($n=100$):  (a). true whole graph; (b). true NSCG; (c). $\widehat{\mathcal{G}}$ by NSCSL with TE; (d). $\widehat{\mathcal{G}}$ by NSCSL with DE; (e).  $\widehat{\mathcal{G}}$ by NOTEARS; (f). $\widehat{\mathcal{G}}$ by PC; (g). $\widehat{\mathcal{G}}$ by LiNGAM.}
\label{fig_scen_res6}  
 \vspace{-0.35cm}
 \end{figure}

 \begin{figure}[!t]
\centering
\begin{subfigure}[]
  \centering
  \includegraphics[width=0.45\linewidth]{figs/S3_True_Whole_MT.pdf} 
\end{subfigure}%
\begin{subfigure}[]
  \centering
  \includegraphics[width=0.45\linewidth]{figs/S3_True_NS_MT.pdf} 
\end{subfigure}\\
\begin{subfigure}[] 
  \centering
  \includegraphics[width=0.45\linewidth]{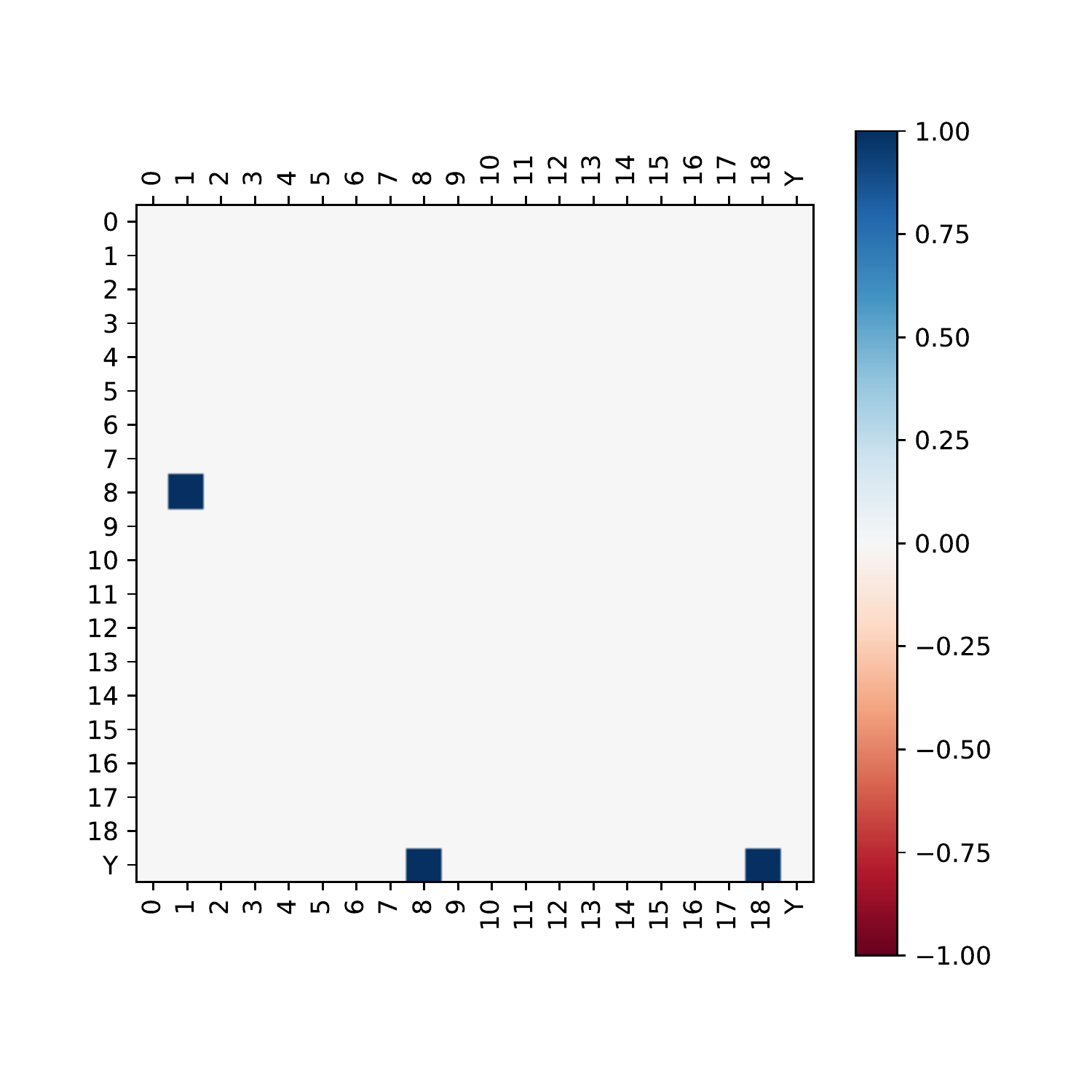} 
\end{subfigure}%
\begin{subfigure}[] 
  \centering
  \includegraphics[width=0.45\linewidth]{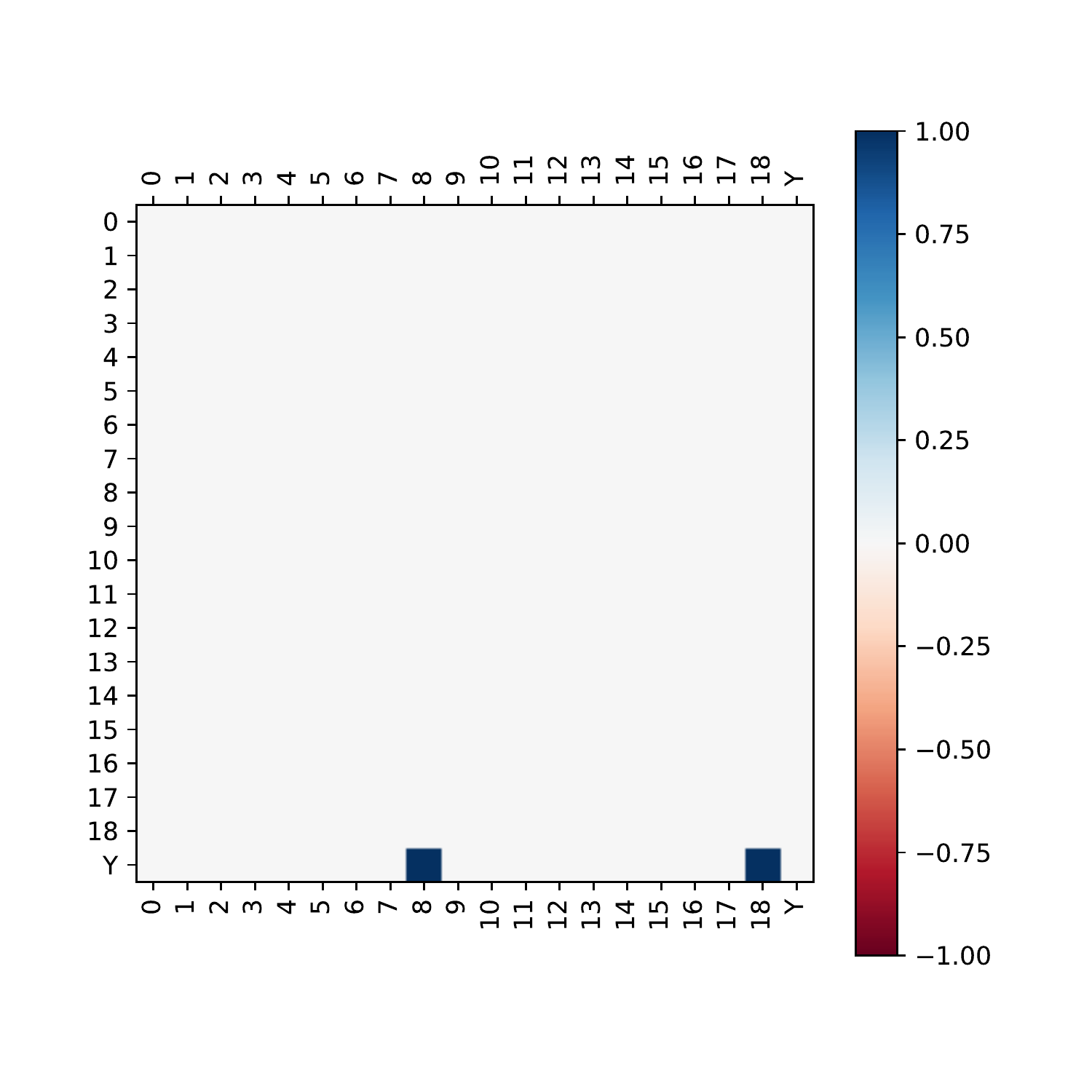} 
\end{subfigure}\\%
\begin{subfigure}[] 
  \centering
  \includegraphics[width=0.31\linewidth]{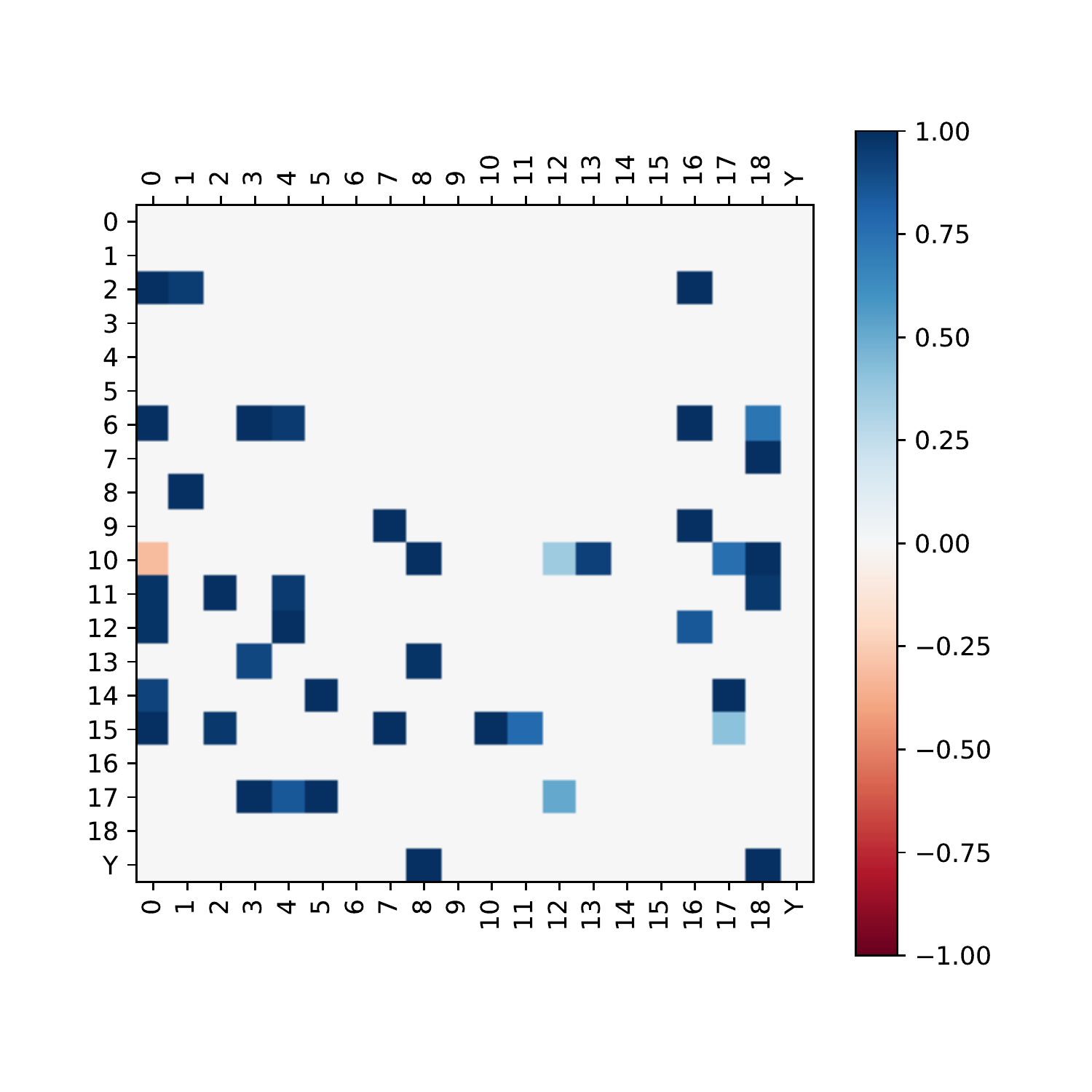} 
\end{subfigure}
\begin{subfigure}[] 
  \centering
  \includegraphics[width=0.31\linewidth]{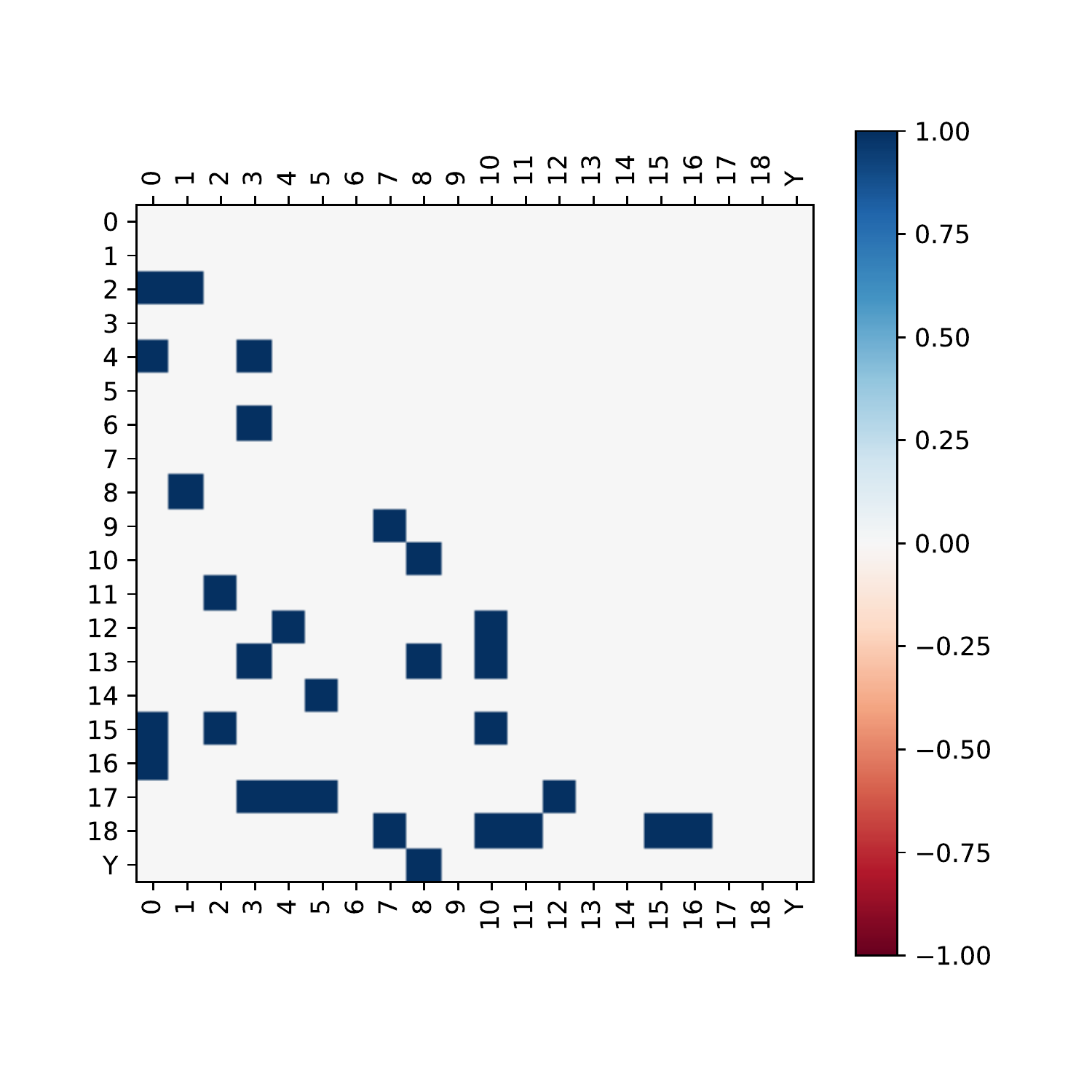} 
\end{subfigure}
\begin{subfigure}[] 
  \centering
  \includegraphics[width=0.31\linewidth]{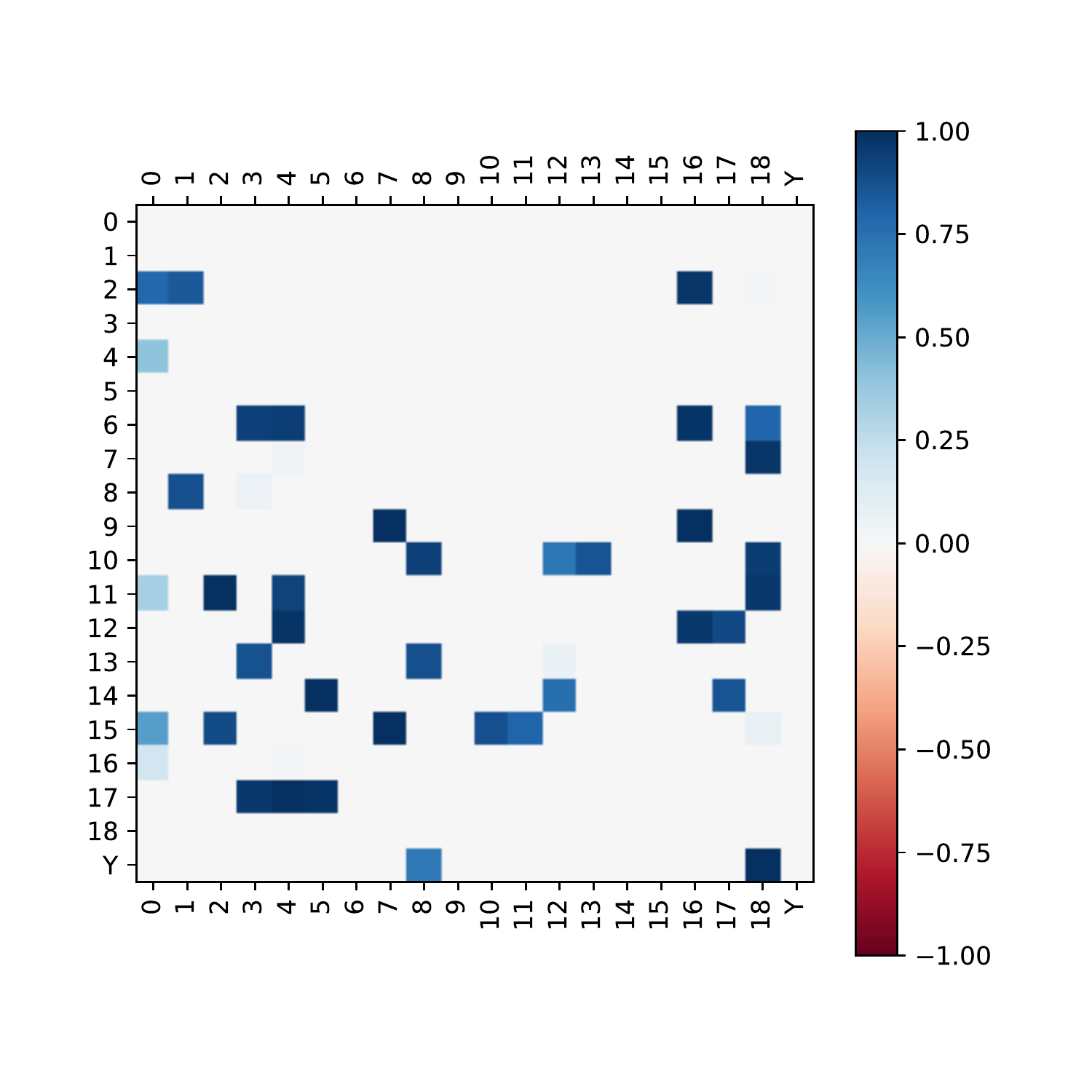}
\end{subfigure}
 \vspace{-0.35cm}
\caption{Estimated matrix under S4 ($n=300$): (a). true whole graph; (b). true NSCG; (c). $\widehat{\mathcal{G}}$ by NSCSL with TE; (d). $\widehat{\mathcal{G}}$ by NSCSL with DE; (e).  $\widehat{\mathcal{G}}$ by NOTEARS; (f). $\widehat{\mathcal{G}}$ by PC; (g). $\widehat{\mathcal{G}}$ by LiNGAM.}
\label{fig_scen_res8}  
 \vspace{-0.4cm}
 \end{figure}
 
 \clearpage
      \subsection{Additional Simulation Results:  True and Estimated Graphs}
 
\begin{figure}[!tph]
\centering
\begin{subfigure}[]
  \centering
  \includegraphics[width=0.45\textwidth]{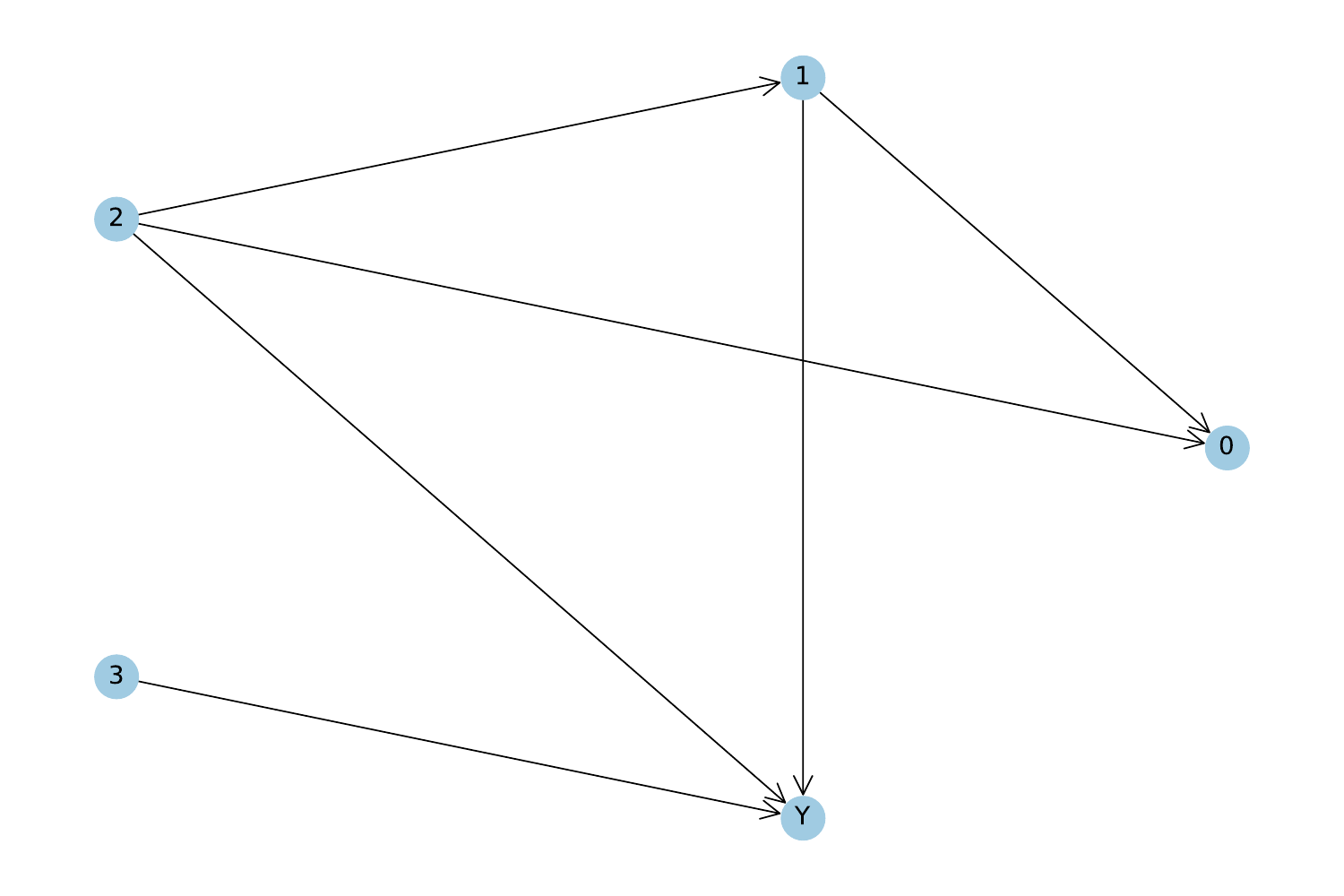} 
\end{subfigure}%
\begin{subfigure}[]
  \centering
  \includegraphics[width=0.45\textwidth]{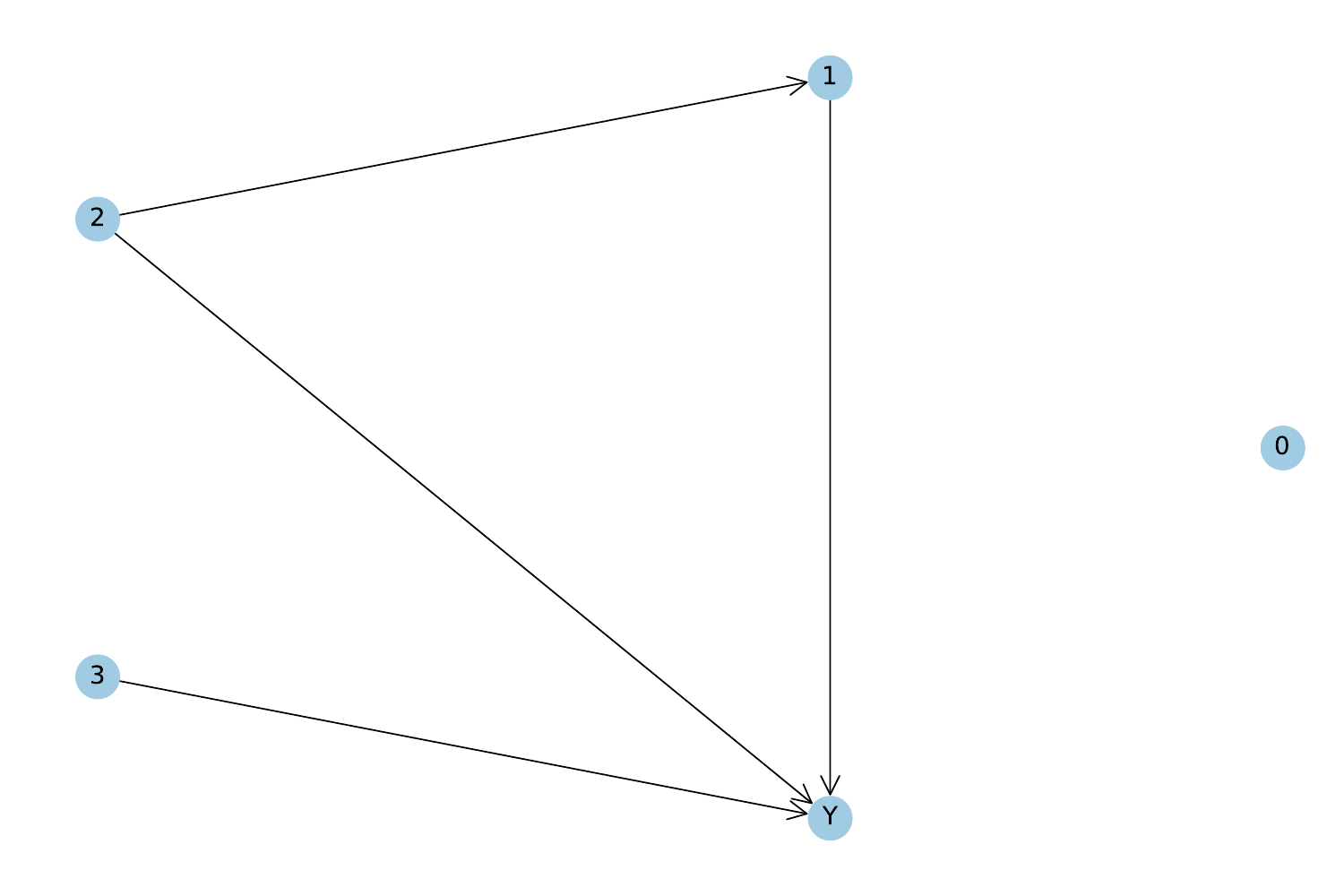} 
\end{subfigure}\\
\begin{subfigure}[] 
  \centering
  \includegraphics[width=0.45\textwidth]{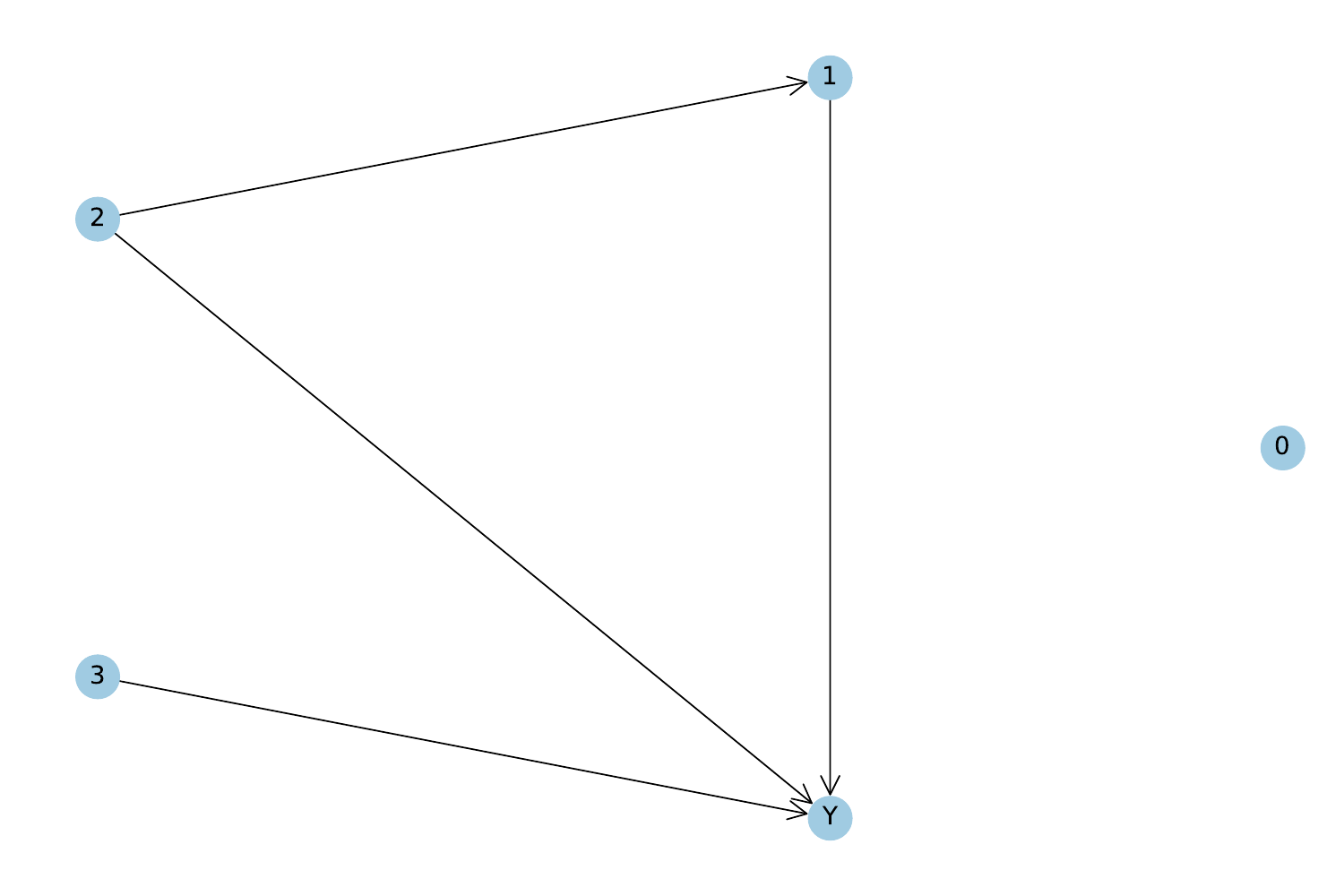} 
\end{subfigure}%
\begin{subfigure}[] 
  \centering
  \includegraphics[width=0.45\textwidth]{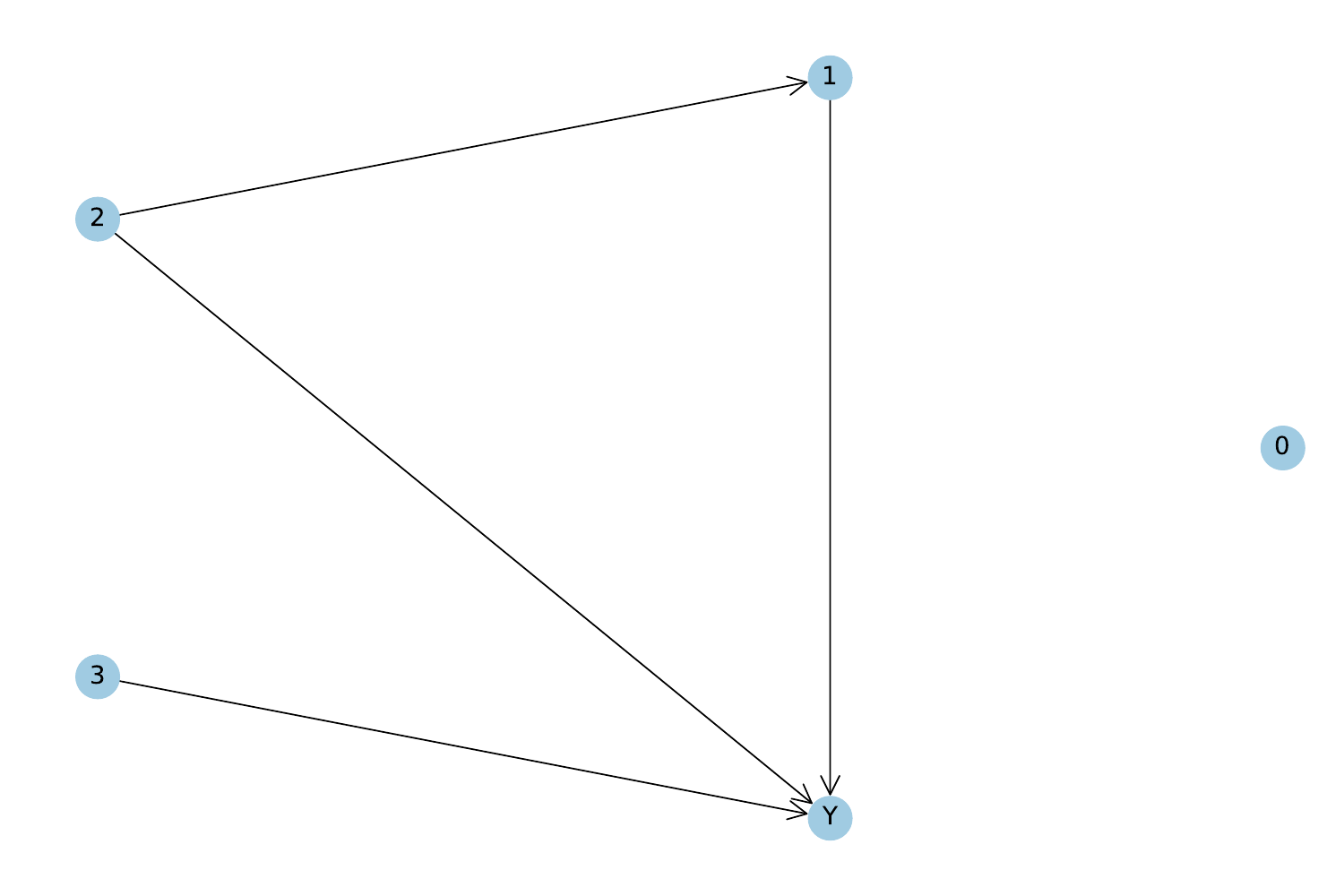} 
\end{subfigure}\\%
\begin{subfigure}[] 
  \centering
  \includegraphics[width=0.31\linewidth]{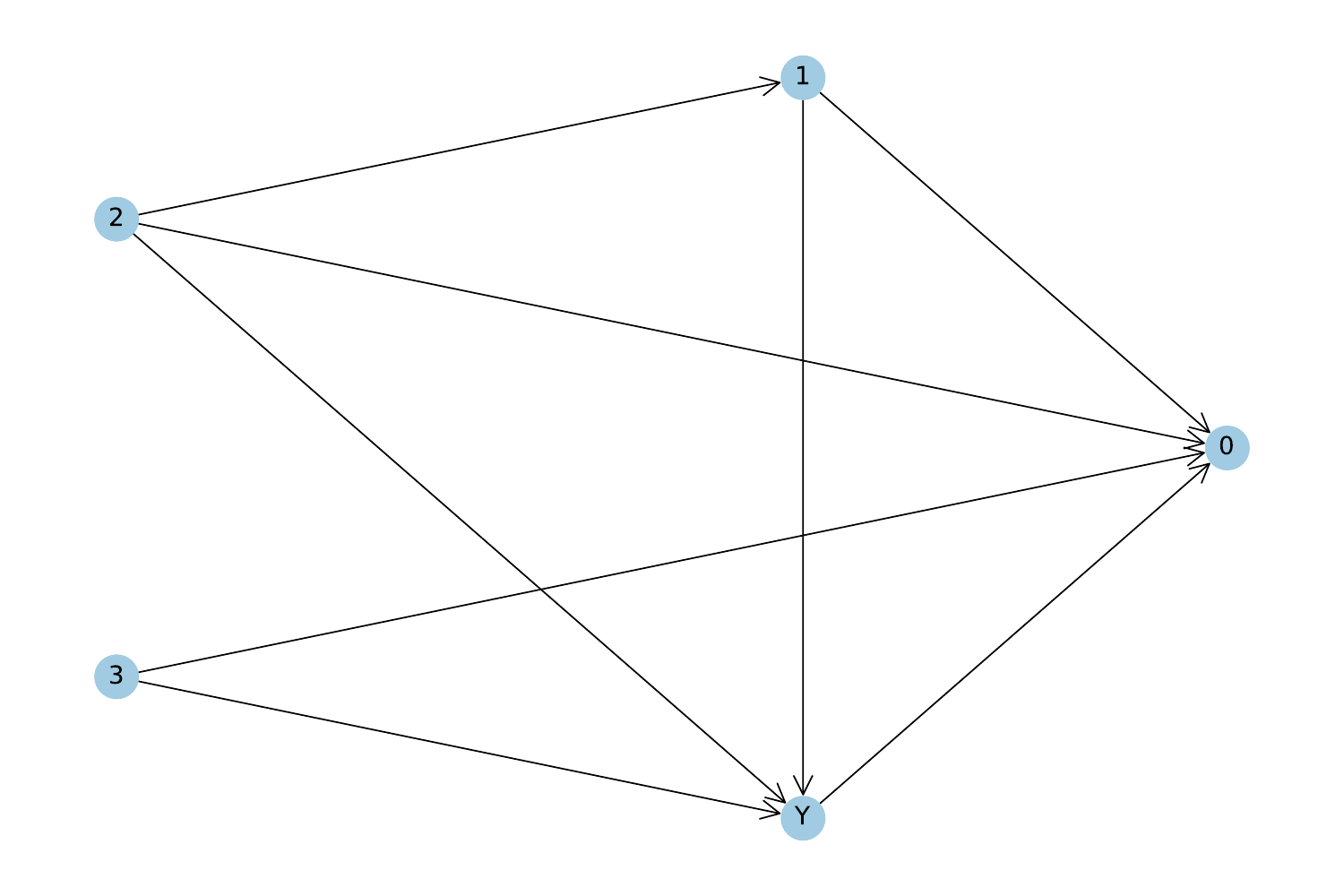} 
\end{subfigure}
\begin{subfigure}[] 
  \centering
  \includegraphics[width=0.31\linewidth]{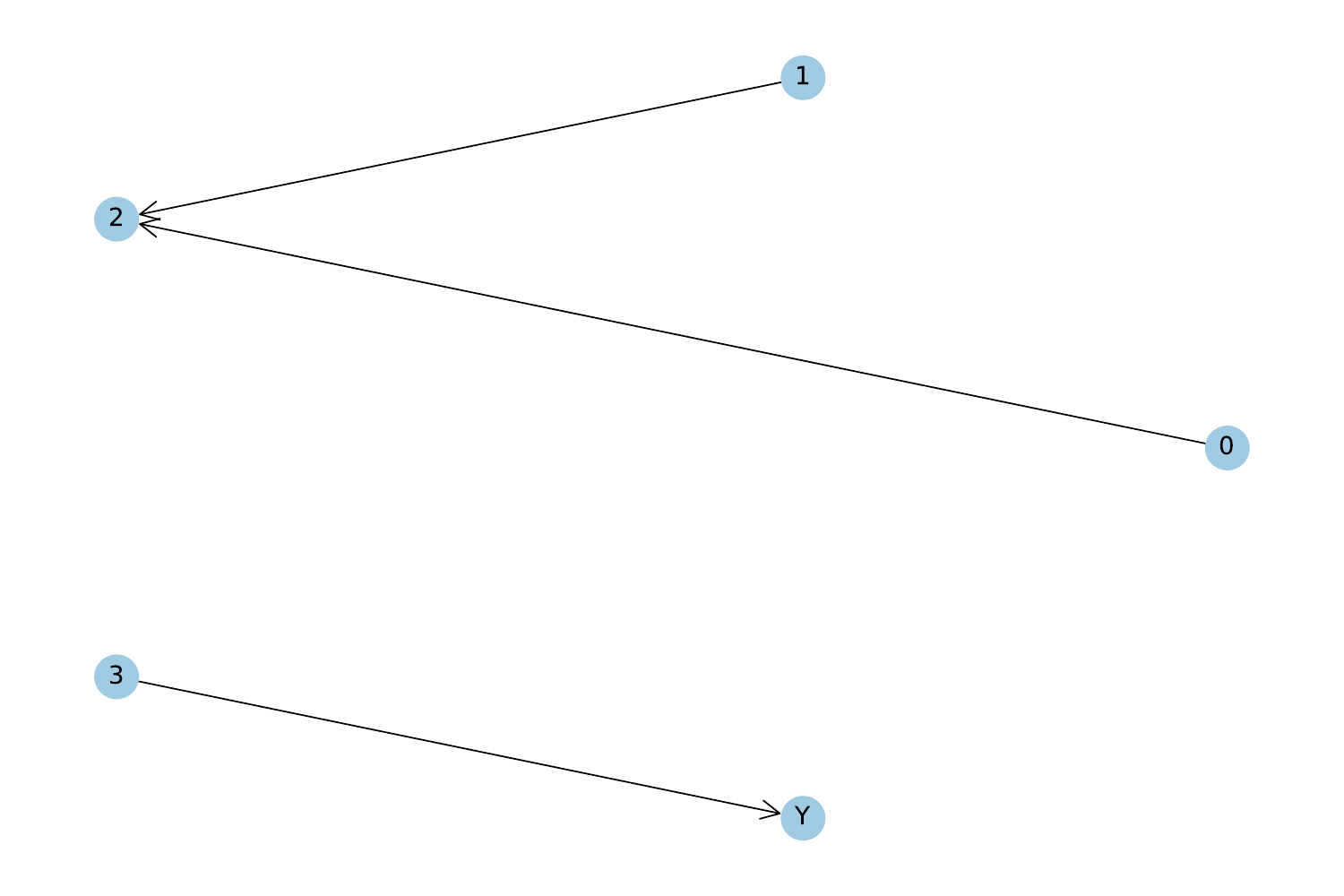} 
\end{subfigure}
\begin{subfigure}[] 
  \centering
  \includegraphics[width=0.31\linewidth]{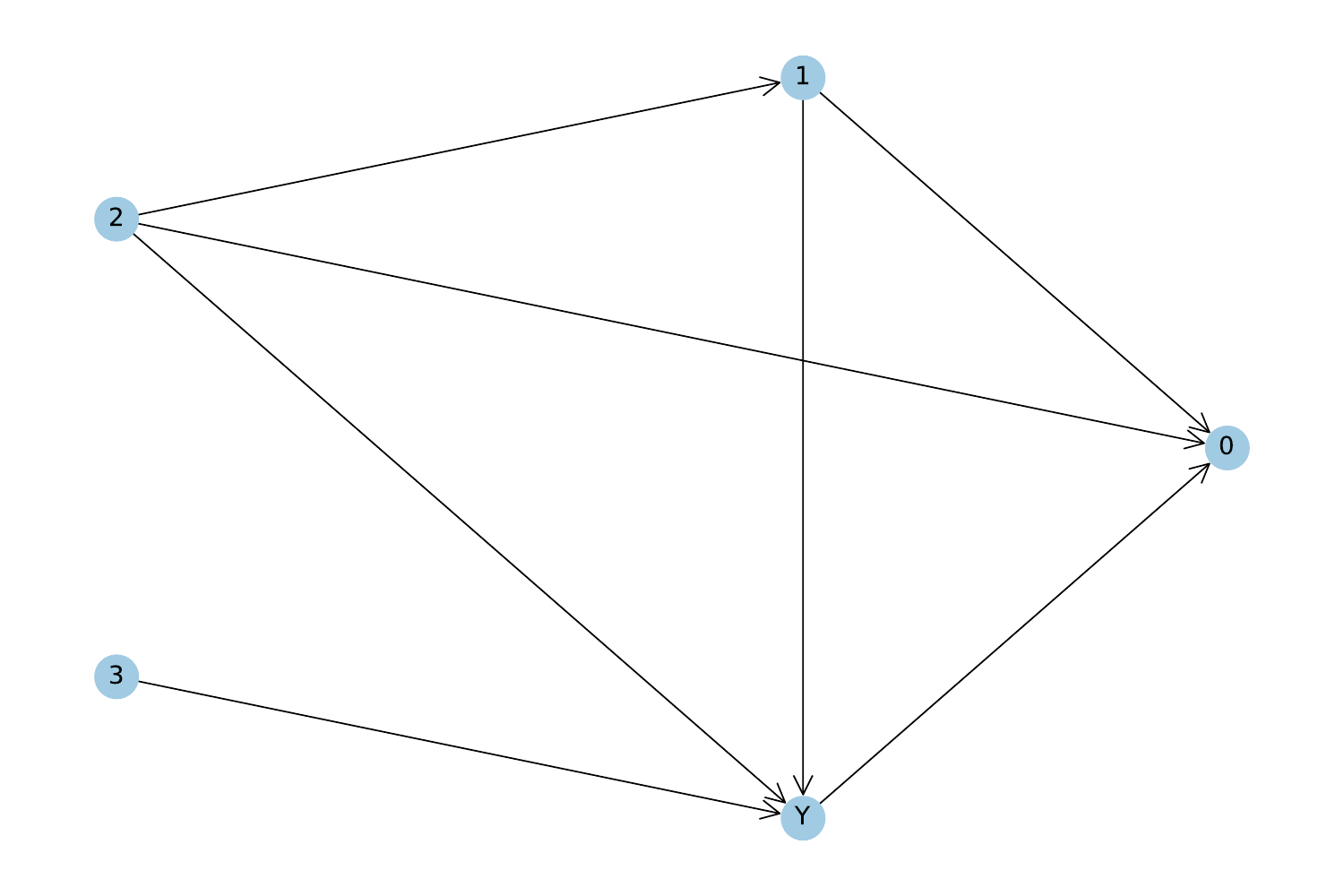}
\end{subfigure}
 \vspace{-0.35cm}
\caption{Graphs under S1 ($n=20$): (a). true whole graph; (b). true NSCG; (c). $\widehat{\mathcal{G}}$ by NSCSL with TE; (d). $\widehat{\mathcal{G}}$ by NSCSL with DE; (e).  $\widehat{\mathcal{G}}$ by NOTEARS; (f). $\widehat{\mathcal{G}}$ by PC; (g). $\widehat{\mathcal{G}}$ by LiNGAM.}
\label{fig_scen_res}  
 \vspace{-0.5cm}
 \end{figure}

\begin{figure}[!htp]
\centering
\begin{subfigure}[]
  \centering
  \includegraphics[width=0.45\textwidth]{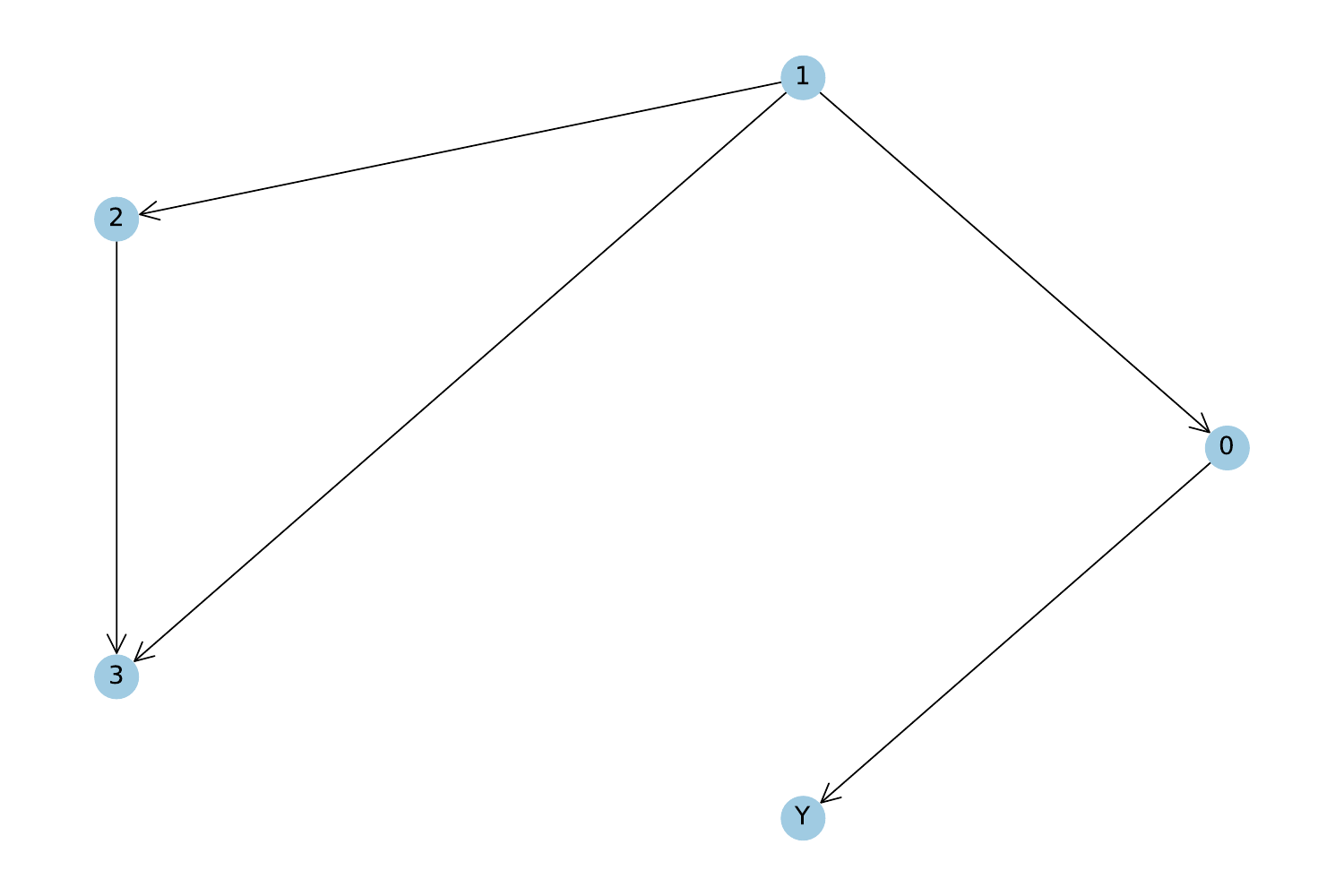} 
\end{subfigure}%
\begin{subfigure}[]
  \centering
  \includegraphics[width=0.45\textwidth]{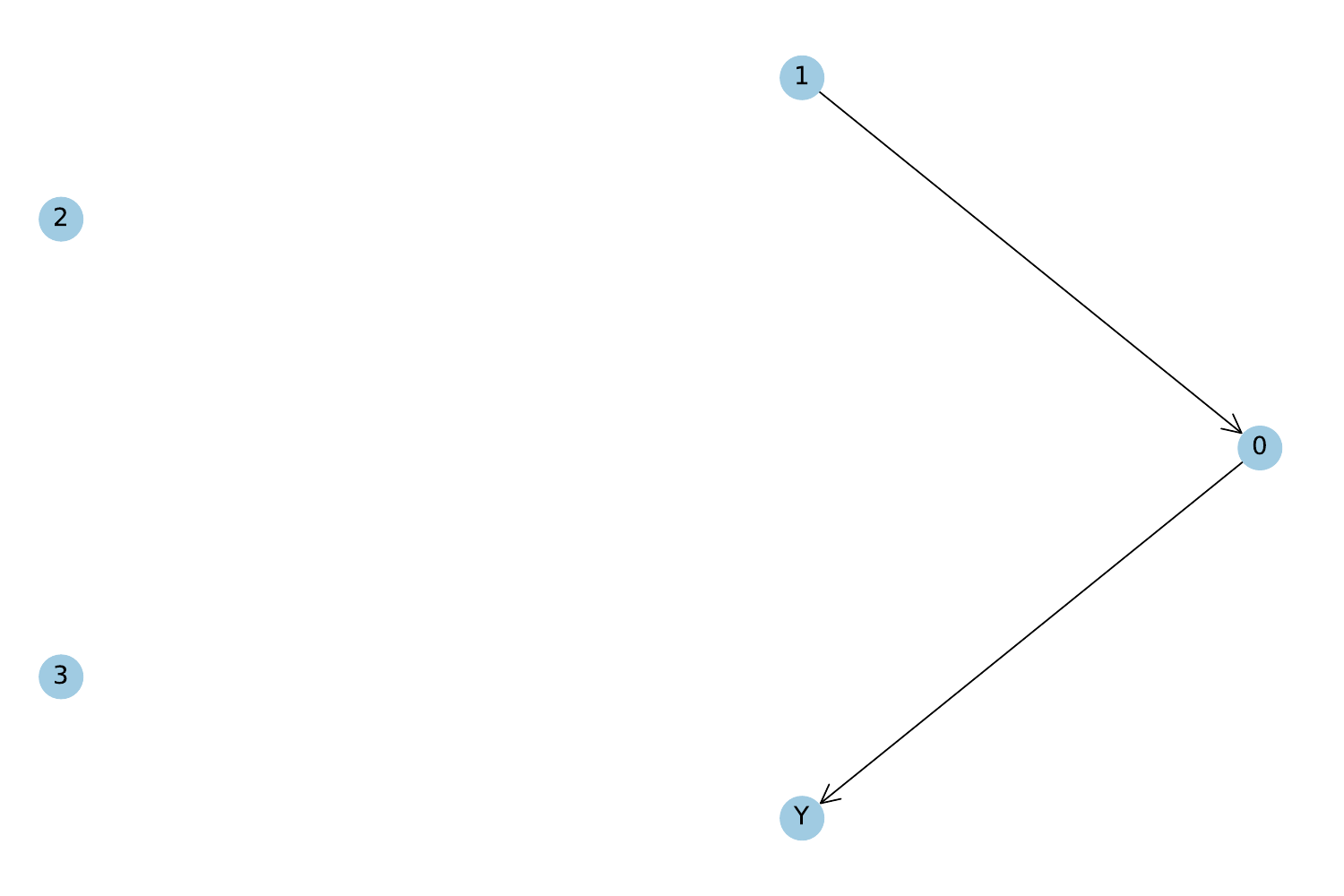} 
\end{subfigure}\\
\begin{subfigure}[] 
  \centering
  \includegraphics[width=0.45\textwidth]{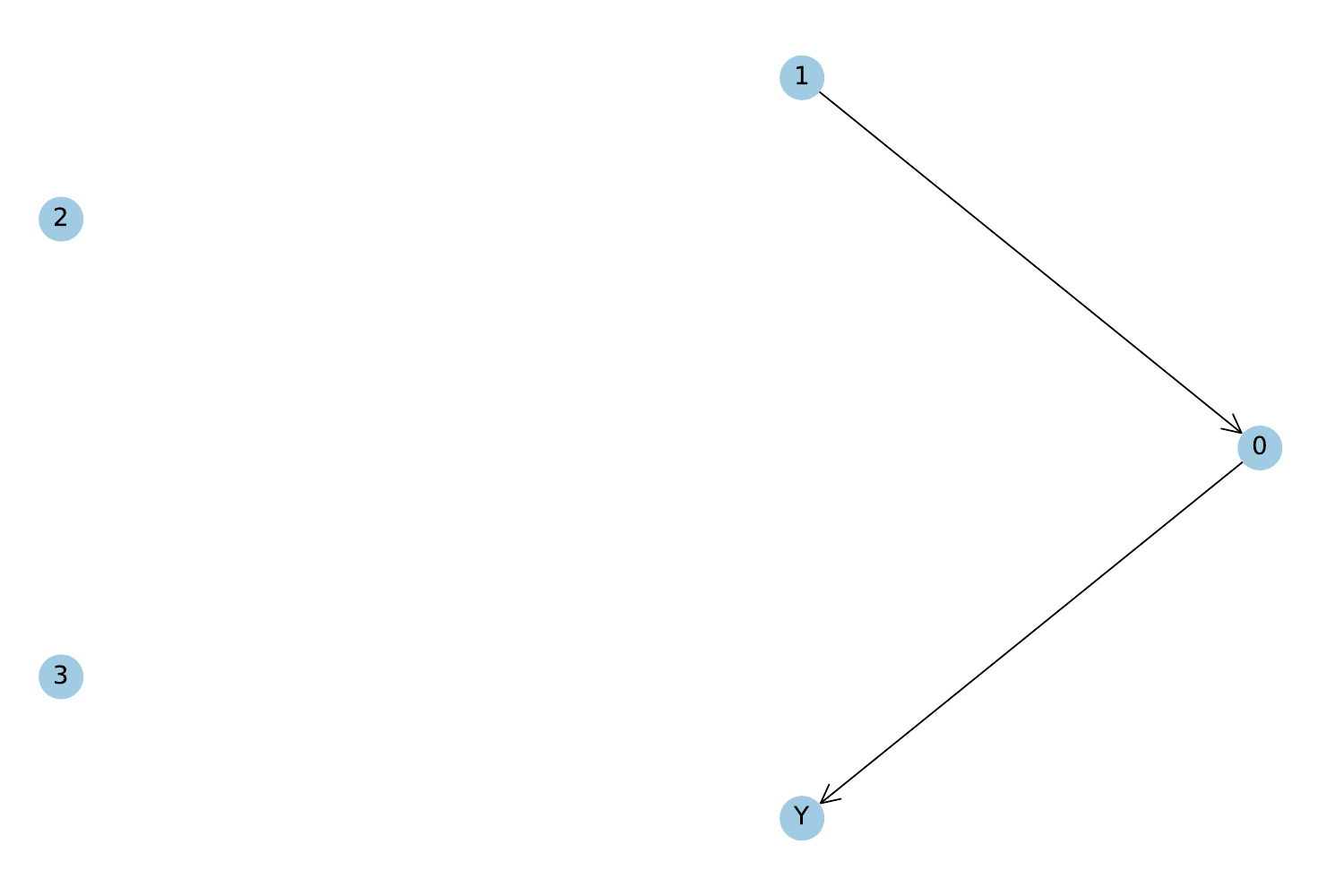} 
\end{subfigure}%
\begin{subfigure}[] 
  \centering
  \includegraphics[width=0.45\textwidth]{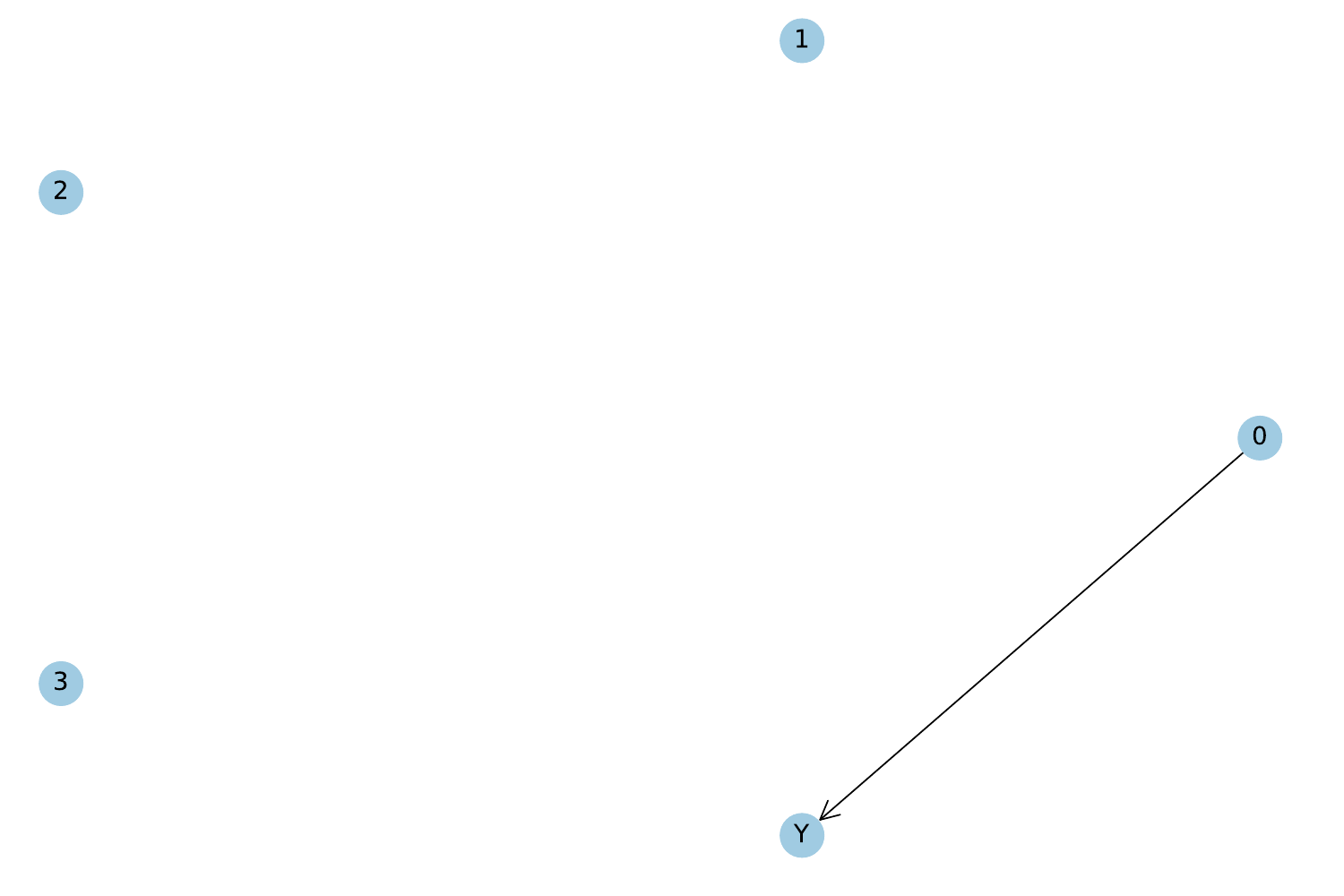} 
\end{subfigure}\\%
\begin{subfigure}[] 
  \centering
  \includegraphics[width=0.31\linewidth]{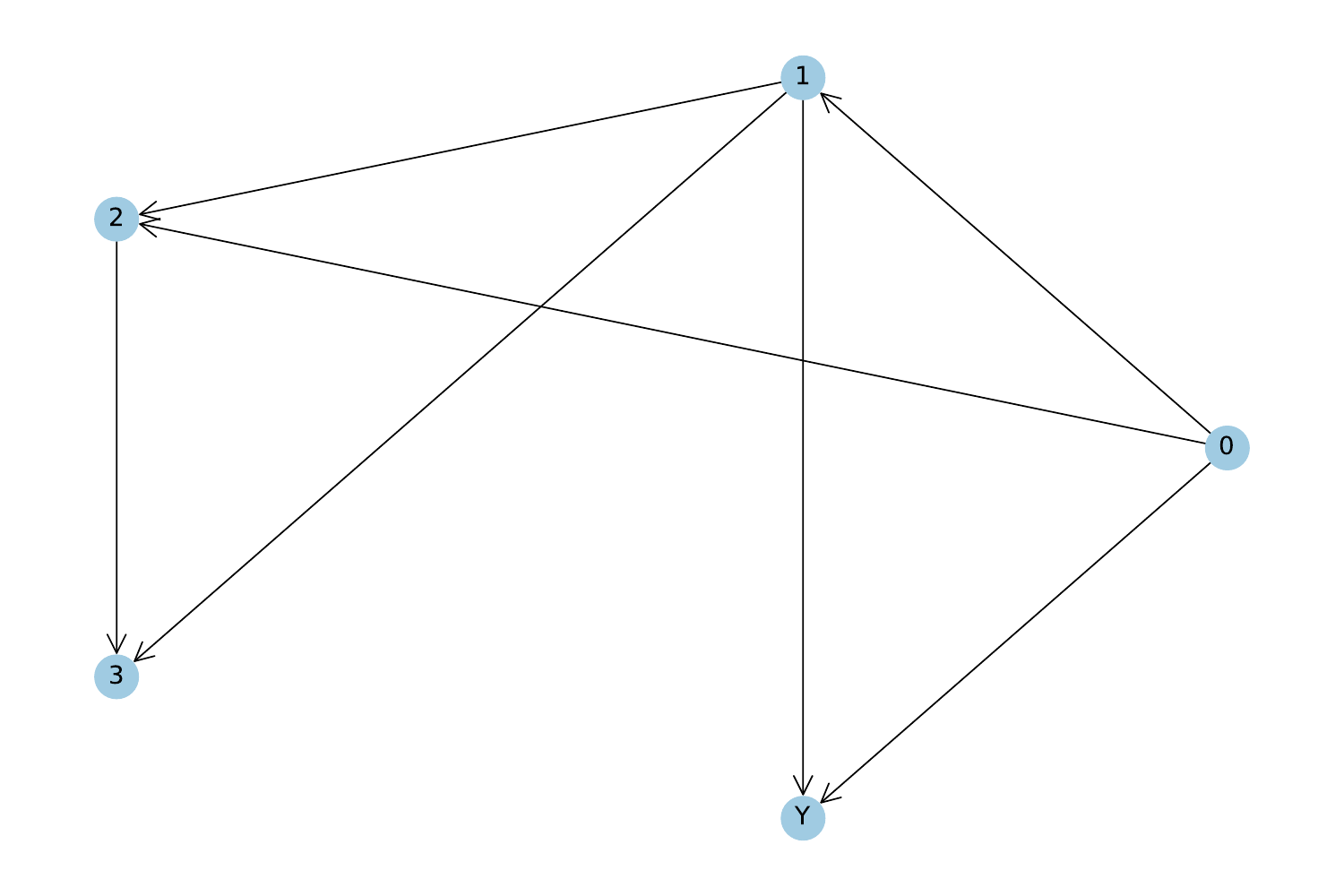} 
\end{subfigure}
\begin{subfigure}[] 
  \centering
  \includegraphics[width=0.31\linewidth]{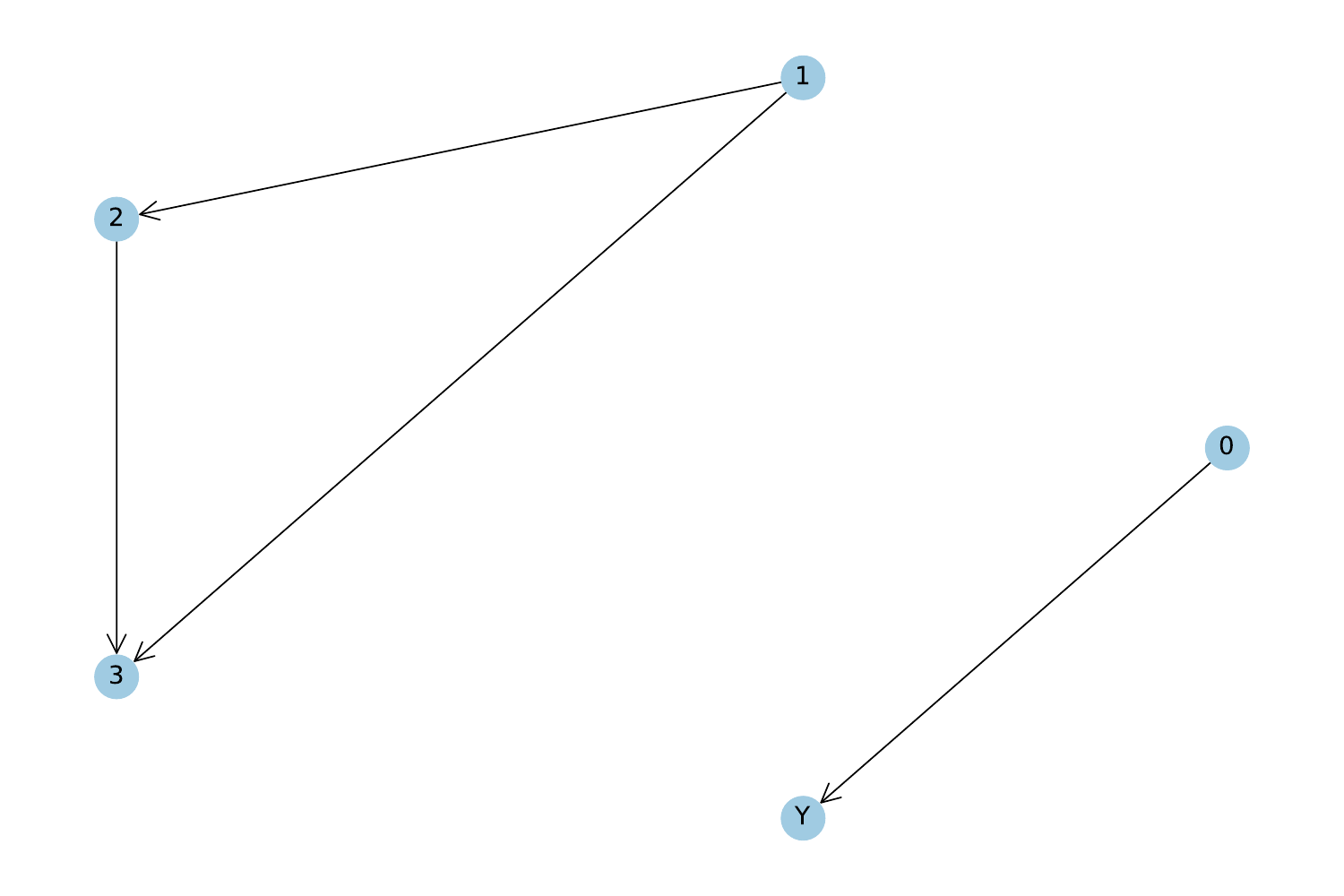} 
\end{subfigure}
\begin{subfigure}[] 
  \centering
  \includegraphics[width=0.31\linewidth]{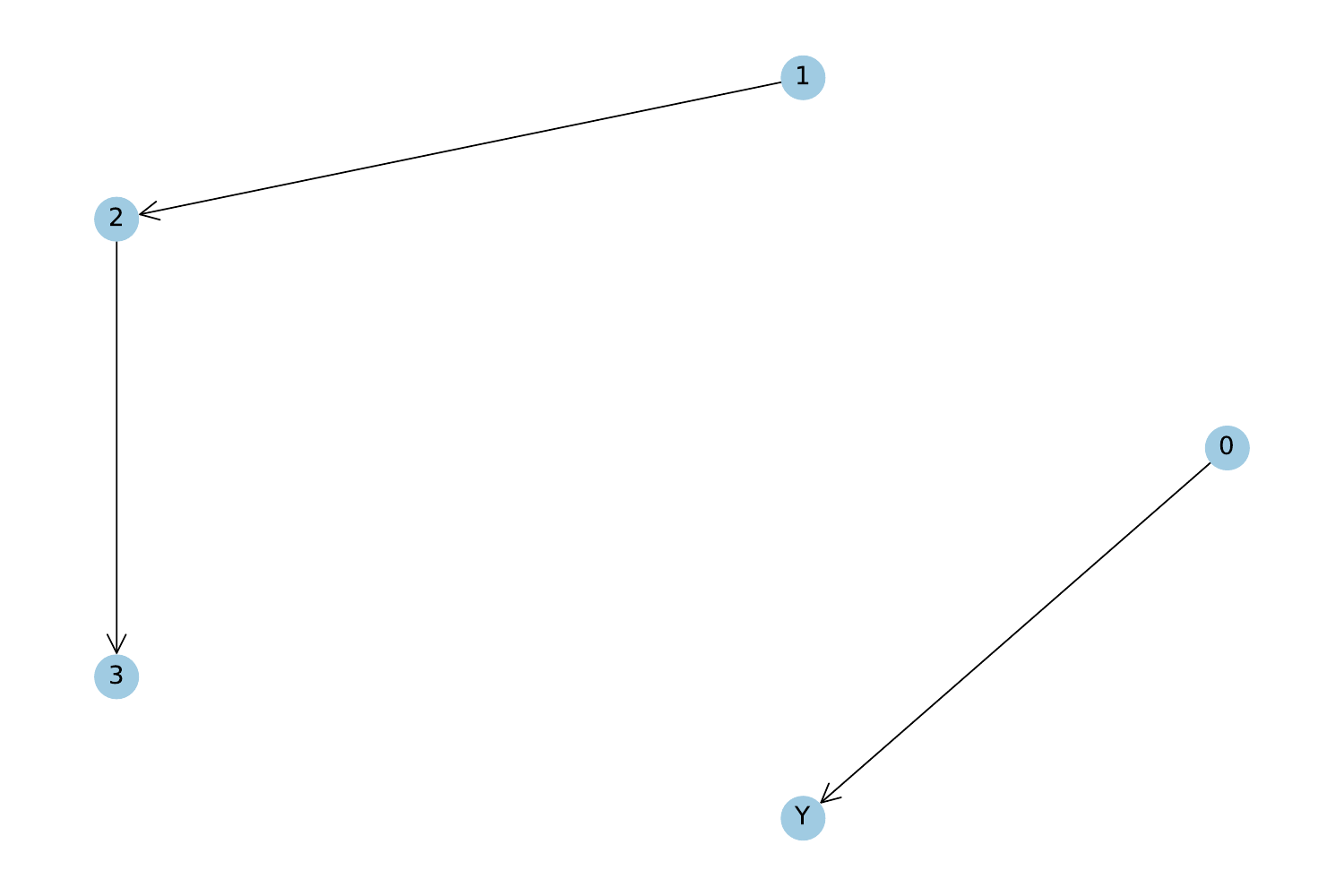}
\end{subfigure}
 \vspace{-0.35cm}
\caption{Graphs under S2 ($n=20$): (a). true whole graph; (b). true NSCG; (c). $\widehat{\mathcal{G}}$ by NSCSL with TE; (d). $\widehat{\mathcal{G}}$ by NSCSL with DE; (e).  $\widehat{\mathcal{G}}$ by NOTEARS; (f). $\widehat{\mathcal{G}}$ by PC; (g). $\widehat{\mathcal{G}}$ by LiNGAM.}
\label{fig_scen_res3}  
 \vspace{-0.35cm}
 \end{figure}

 \begin{figure}[!hpt] 
\centering
\begin{subfigure}[]
  \centering
  \includegraphics[width=0.45\textwidth]{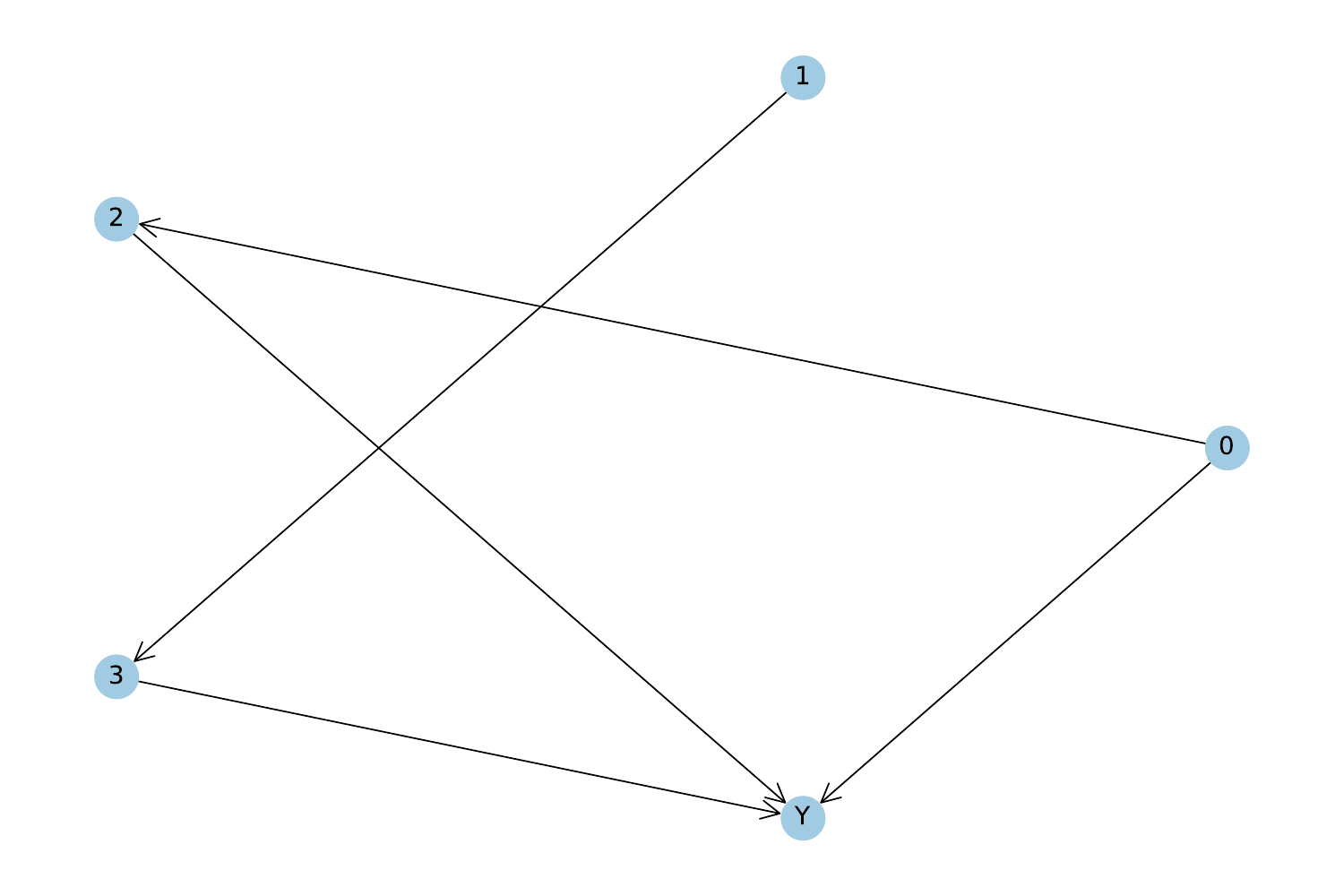} 
\end{subfigure}%
\begin{subfigure}[]
  \centering
  \includegraphics[width=0.45\textwidth]{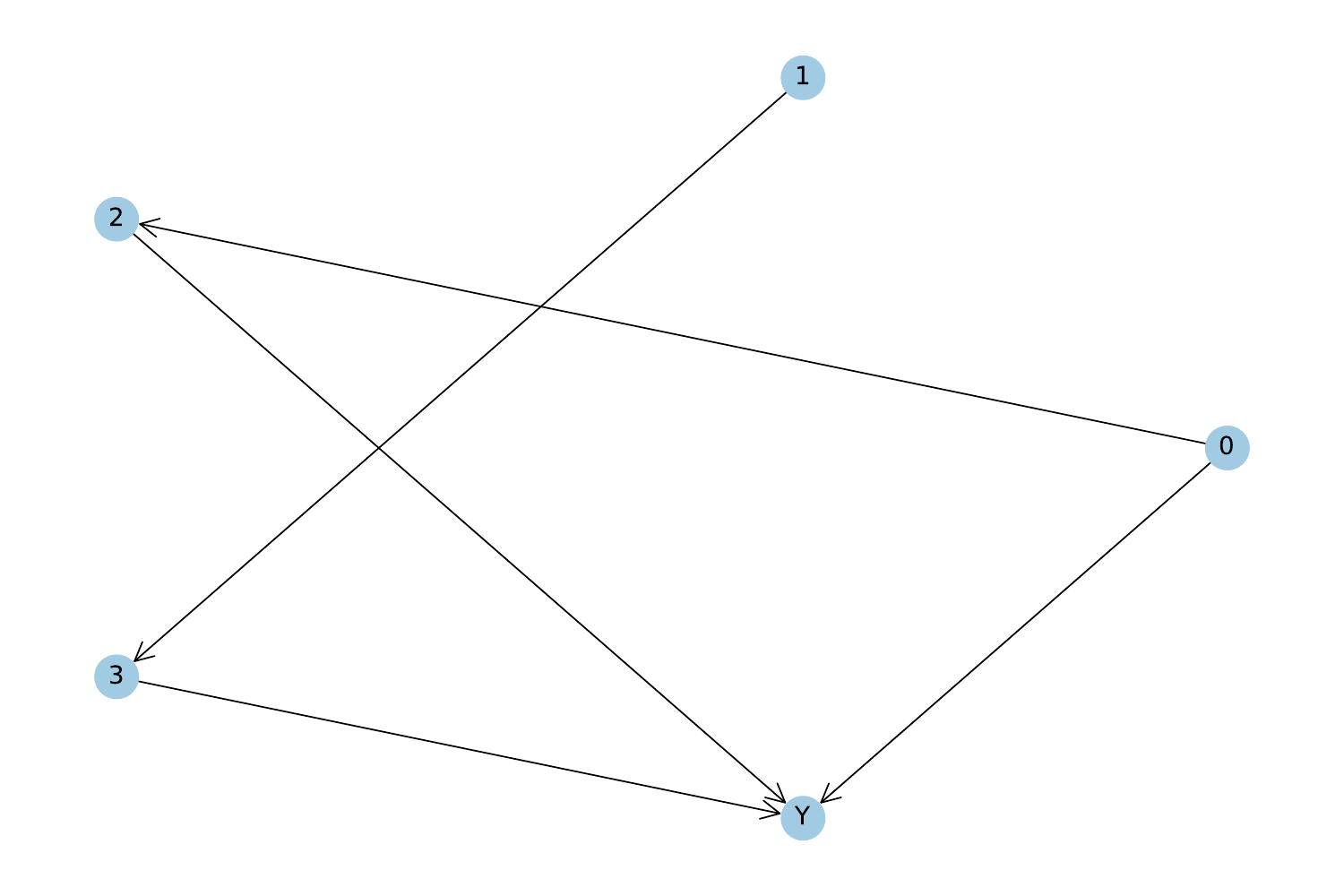} 
\end{subfigure}\\
\begin{subfigure}[] 
  \centering
  \includegraphics[width=0.45\textwidth]{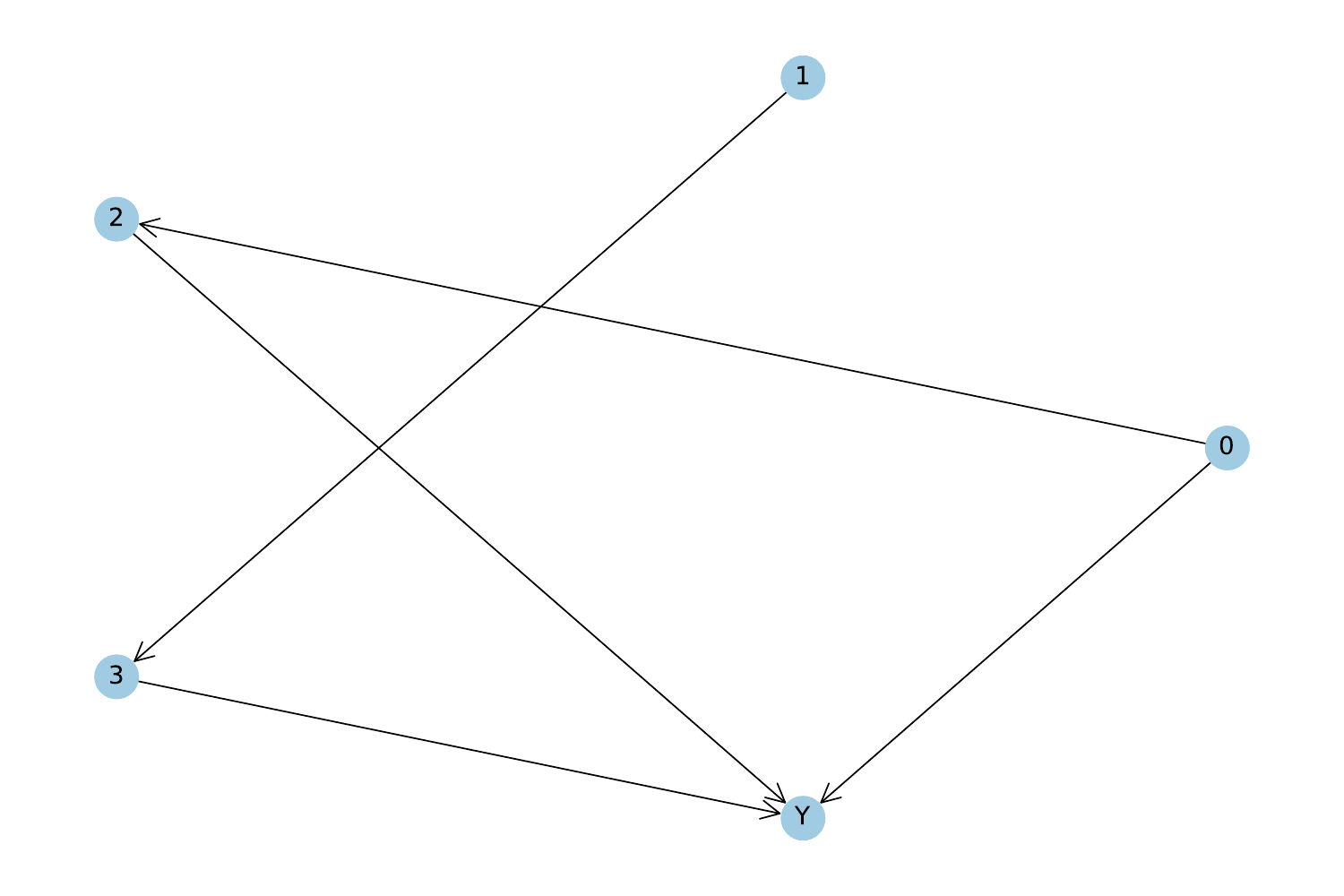} 
\end{subfigure}%
\begin{subfigure}[] 
  \centering
  \includegraphics[width=0.45\textwidth]{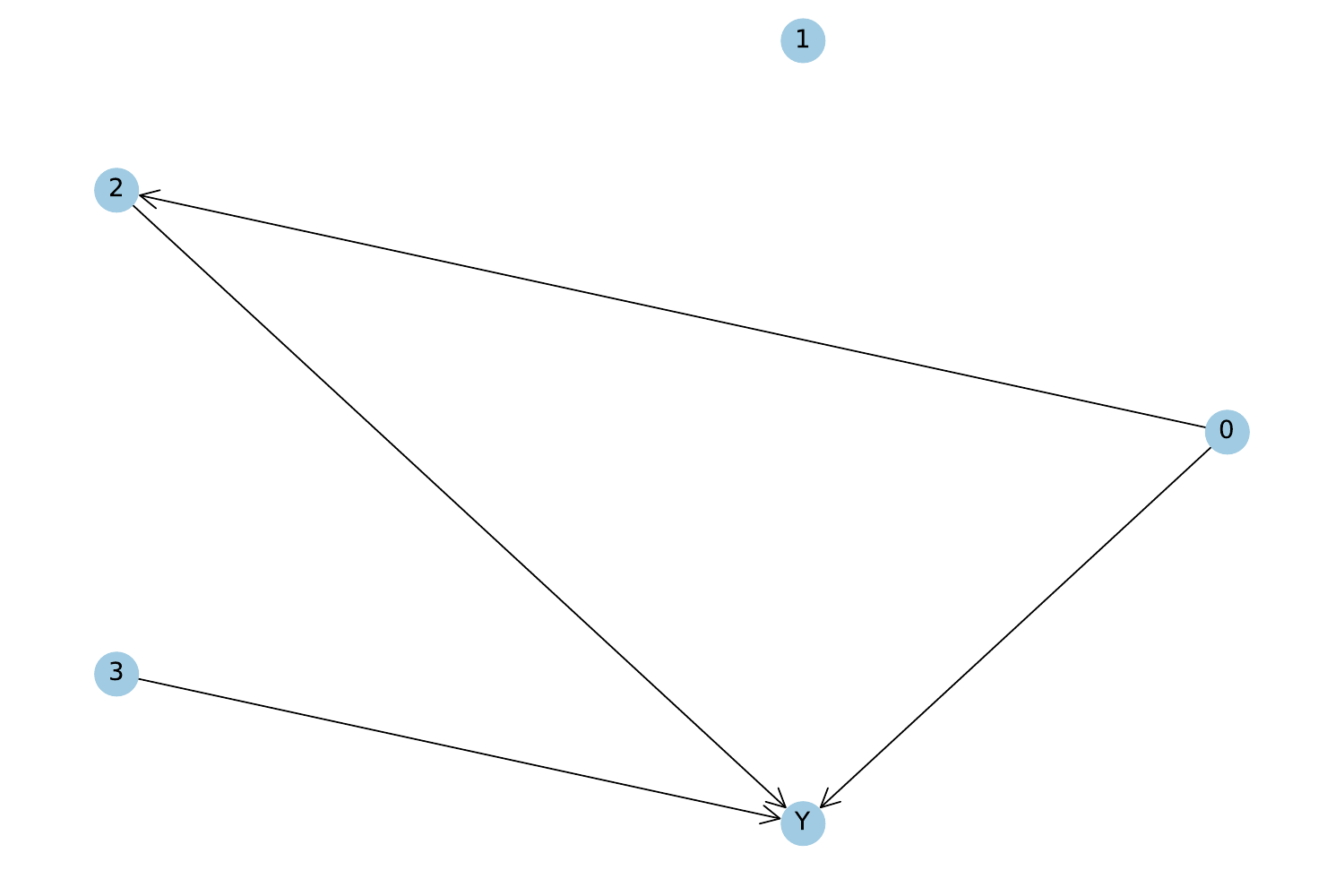} 
\end{subfigure}\\%
\begin{subfigure}[] 
  \centering
  \includegraphics[width=0.31\linewidth]{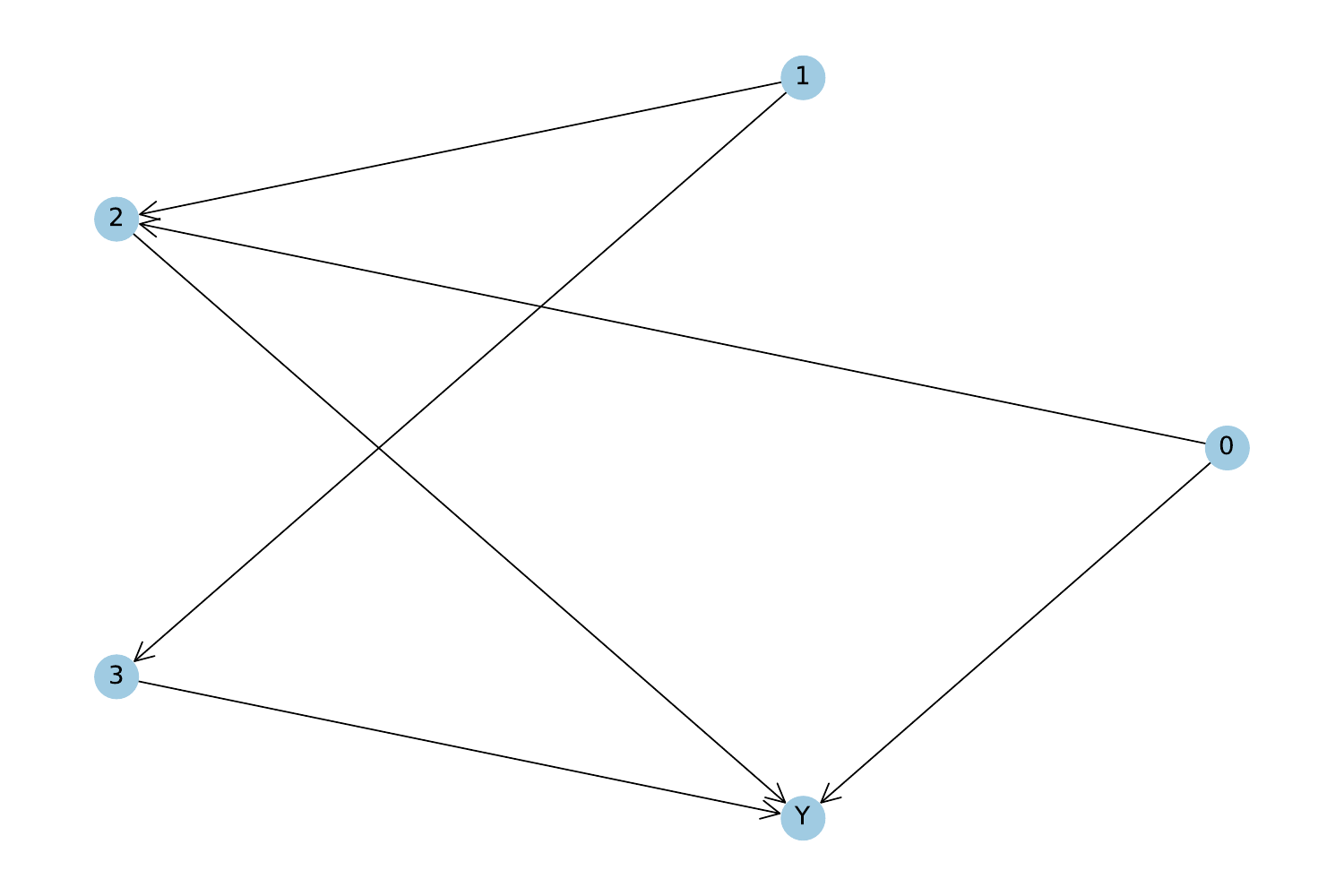} 
\end{subfigure}
\begin{subfigure}[] 
  \centering
  \includegraphics[width=0.31\linewidth]{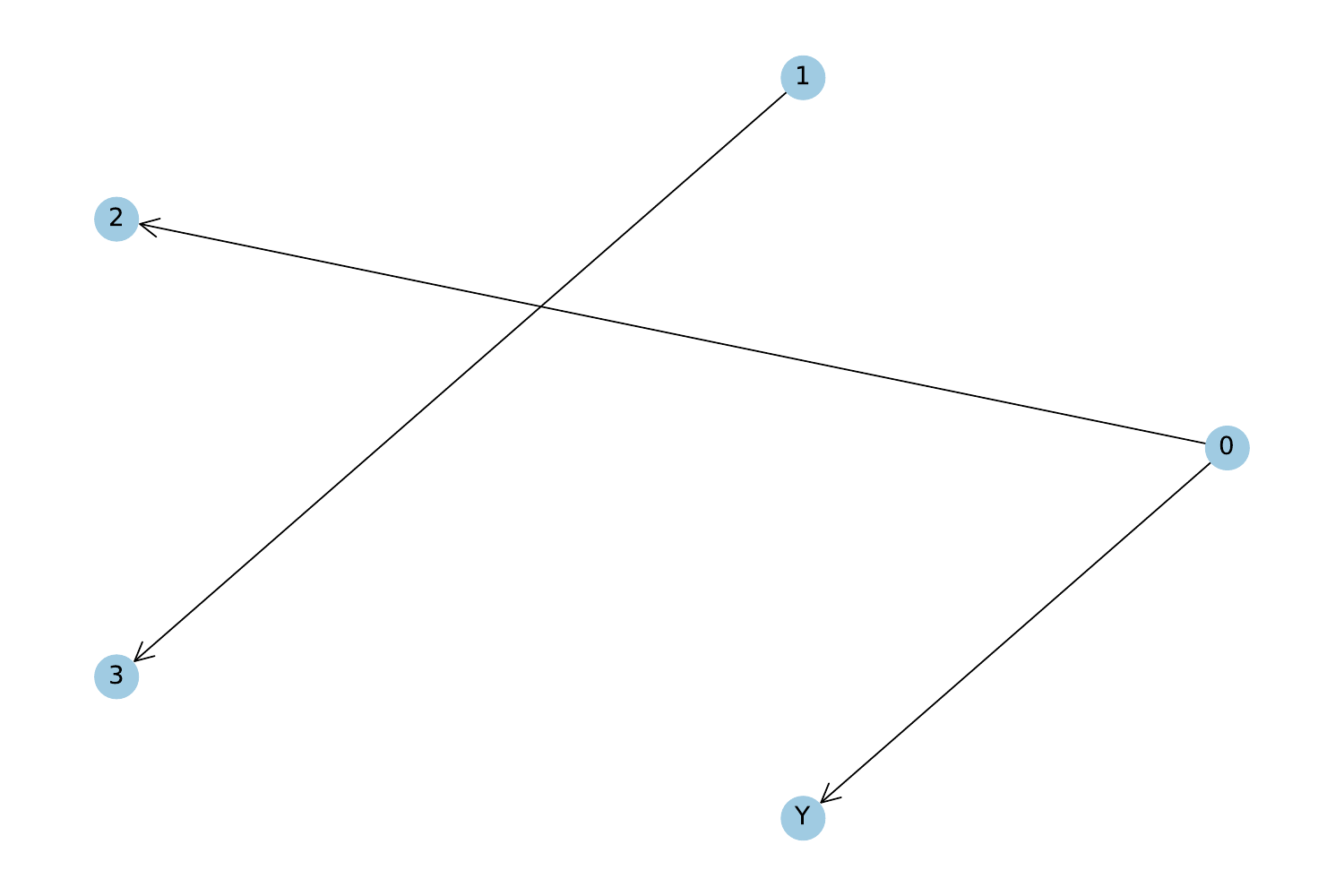} 
\end{subfigure}
\begin{subfigure}[] 
  \centering
  \includegraphics[width=0.31\linewidth]{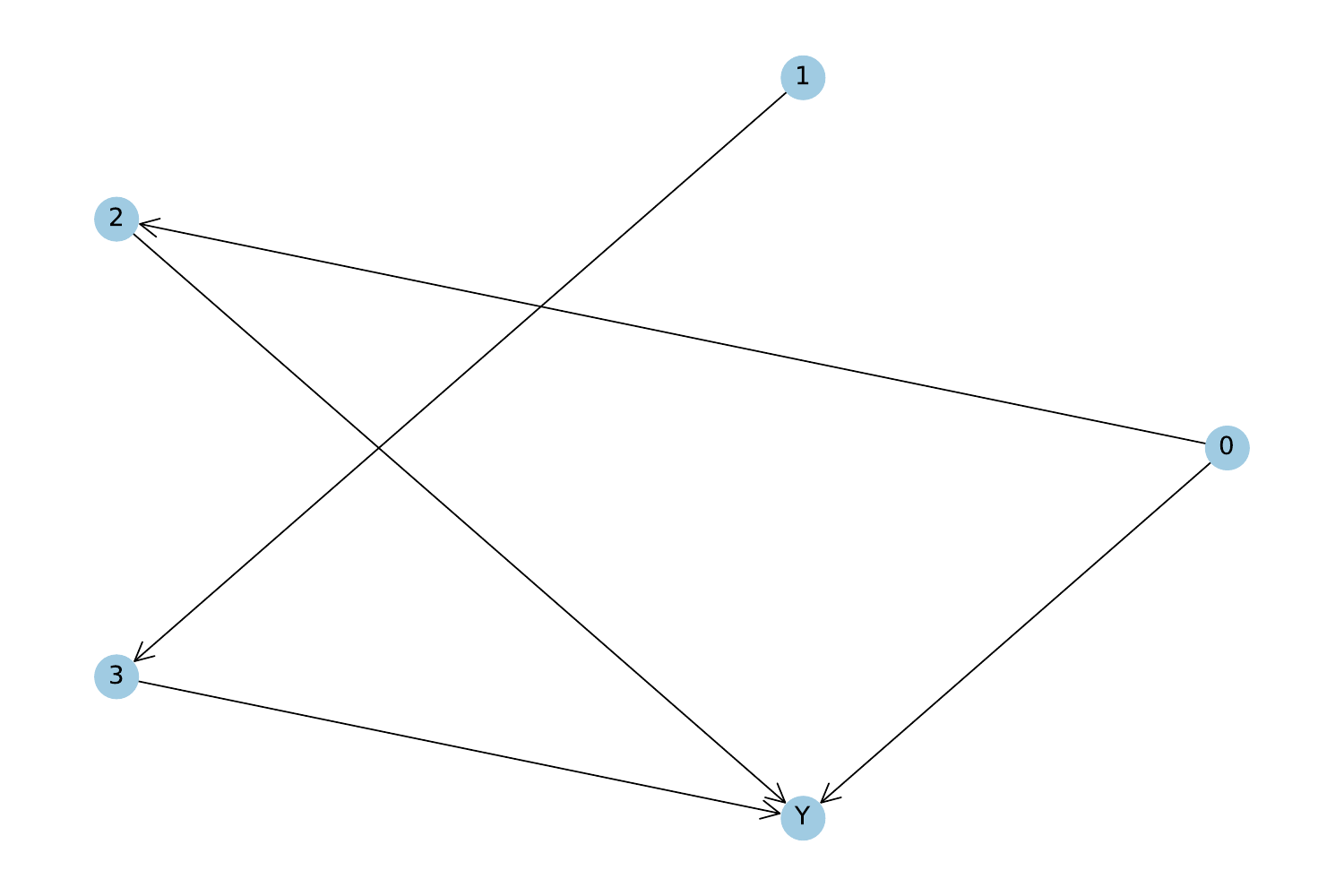}
\end{subfigure}
 \vspace{-0.35cm}
\caption{Graphs under S3 ($n=20$): (a). true whole graph; (b). true NSCG; (c). $\widehat{\mathcal{G}}$ by NSCSL with TE; (d). $\widehat{\mathcal{G}}$ by NSCSL with DE; (e).  $\widehat{\mathcal{G}}$ by NOTEARS; (f). $\widehat{\mathcal{G}}$ by PC; (g). $\widehat{\mathcal{G}}$ by LiNGAM.}
\label{fig_scen_res00}  
 \vspace{-0.35cm}
 \end{figure}

\begin{figure} 
\centering
\begin{subfigure}[]
  \centering
  \includegraphics[width=0.50\textwidth]{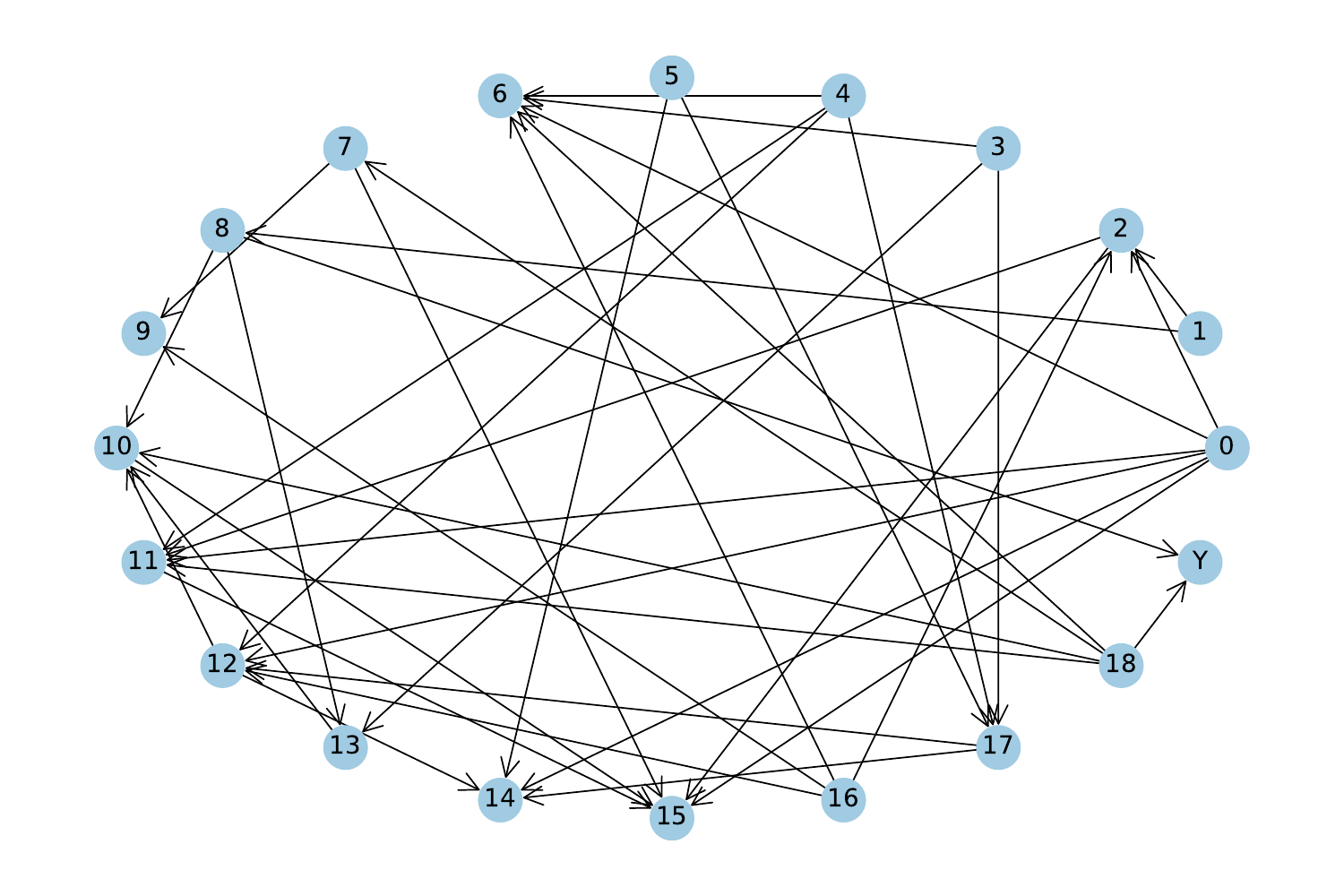} 
\end{subfigure}%
\begin{subfigure}[]
  \centering
  \includegraphics[width=0.45\textwidth]{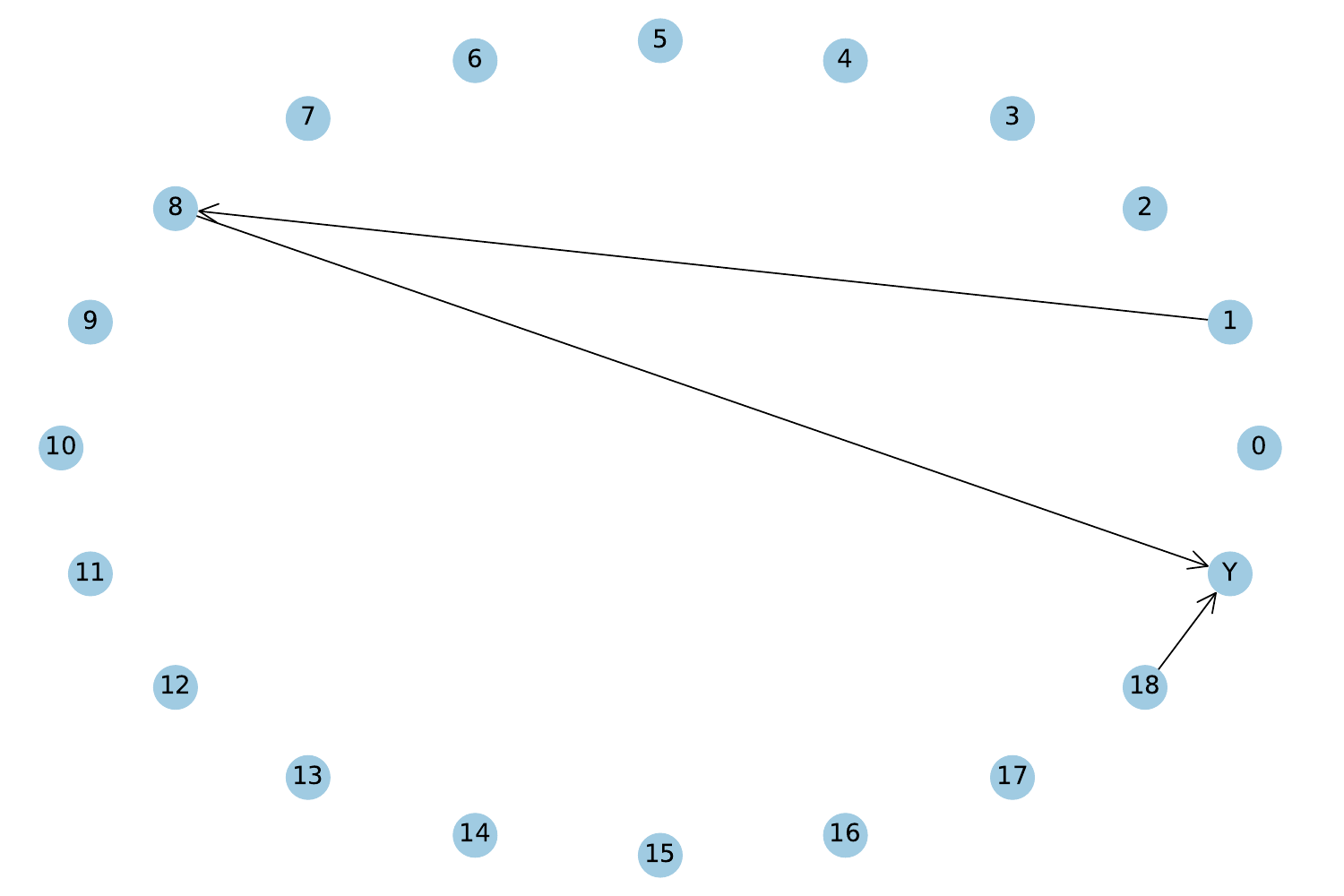} 
\end{subfigure}\\
\begin{subfigure}[] 
  \centering
  \includegraphics[width=0.45\textwidth]{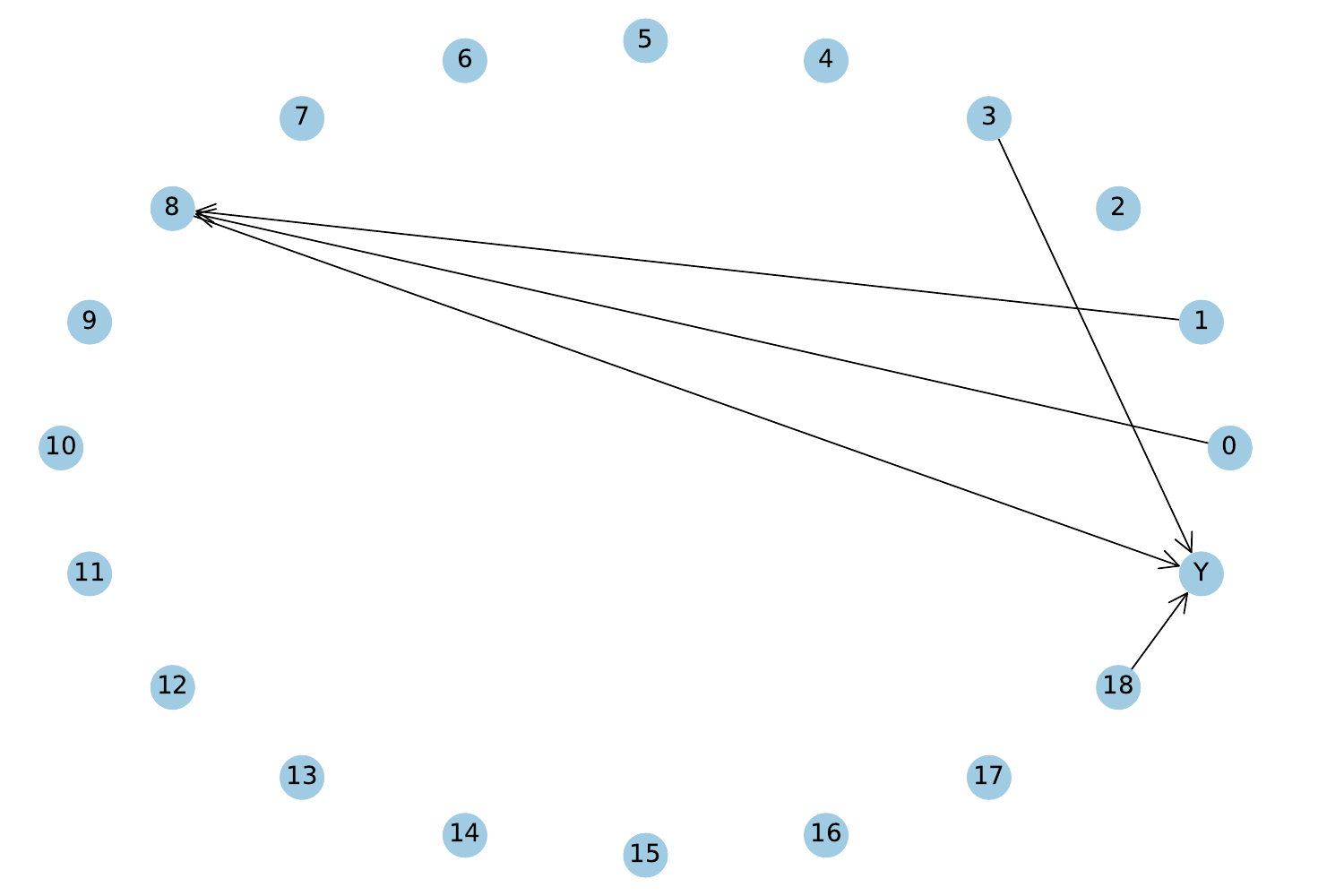} 
\end{subfigure}%
\begin{subfigure}[] 
  \centering
  \includegraphics[width=0.45\textwidth]{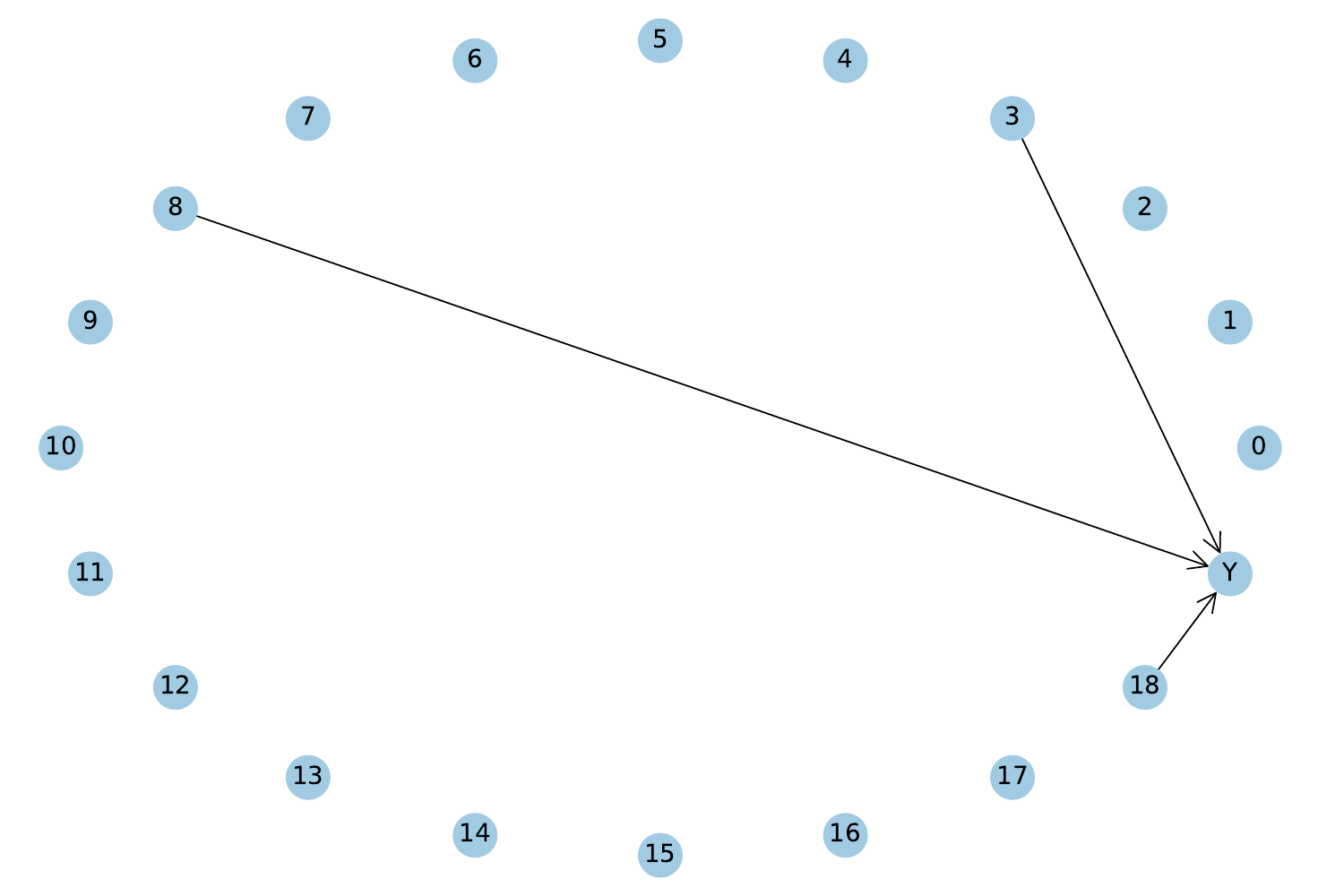} 
\end{subfigure}\\%
\begin{subfigure}[] 
  \centering
  \includegraphics[width=0.31\linewidth]{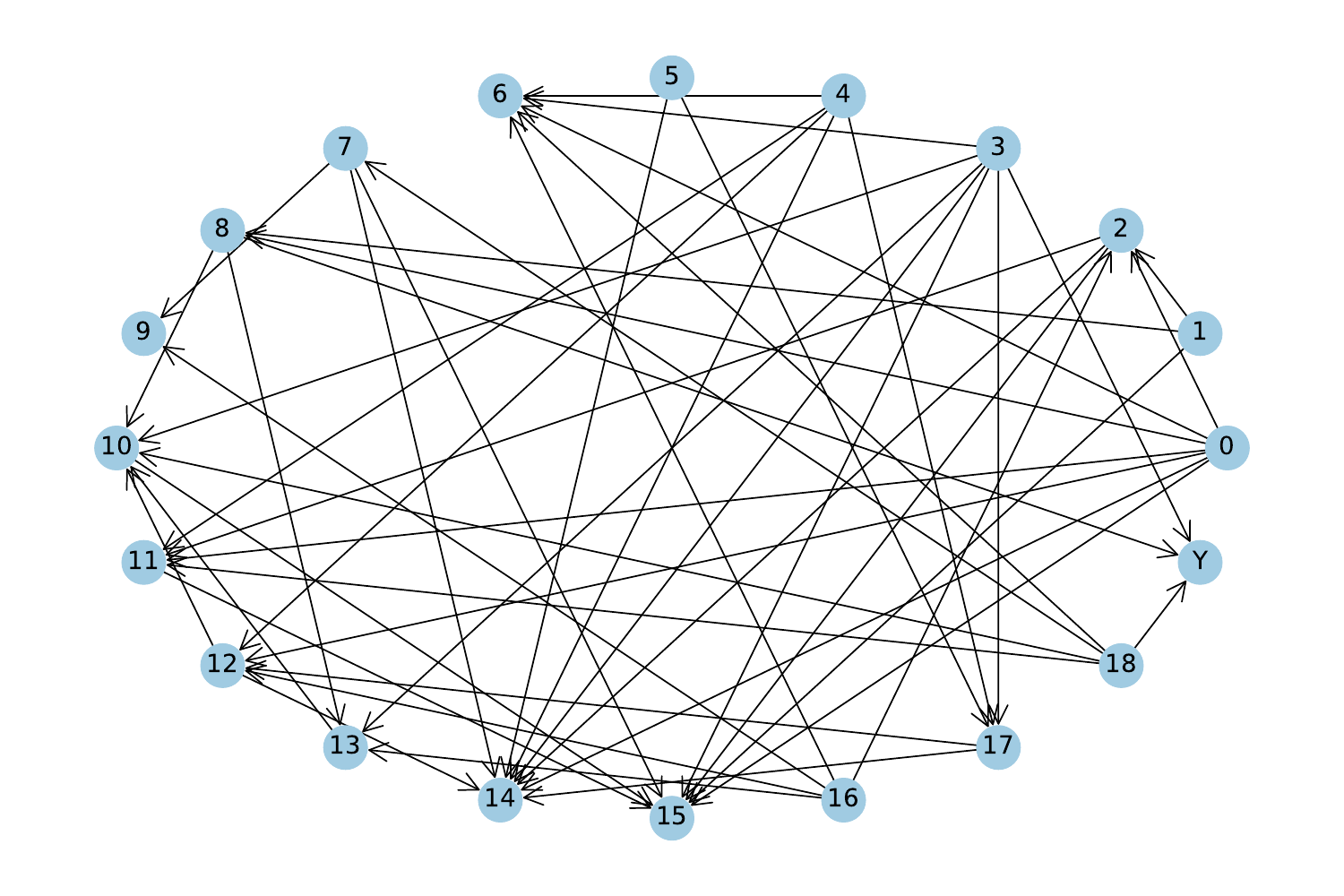} 
\end{subfigure}
\begin{subfigure}[] 
  \centering
  \includegraphics[width=0.31\linewidth]{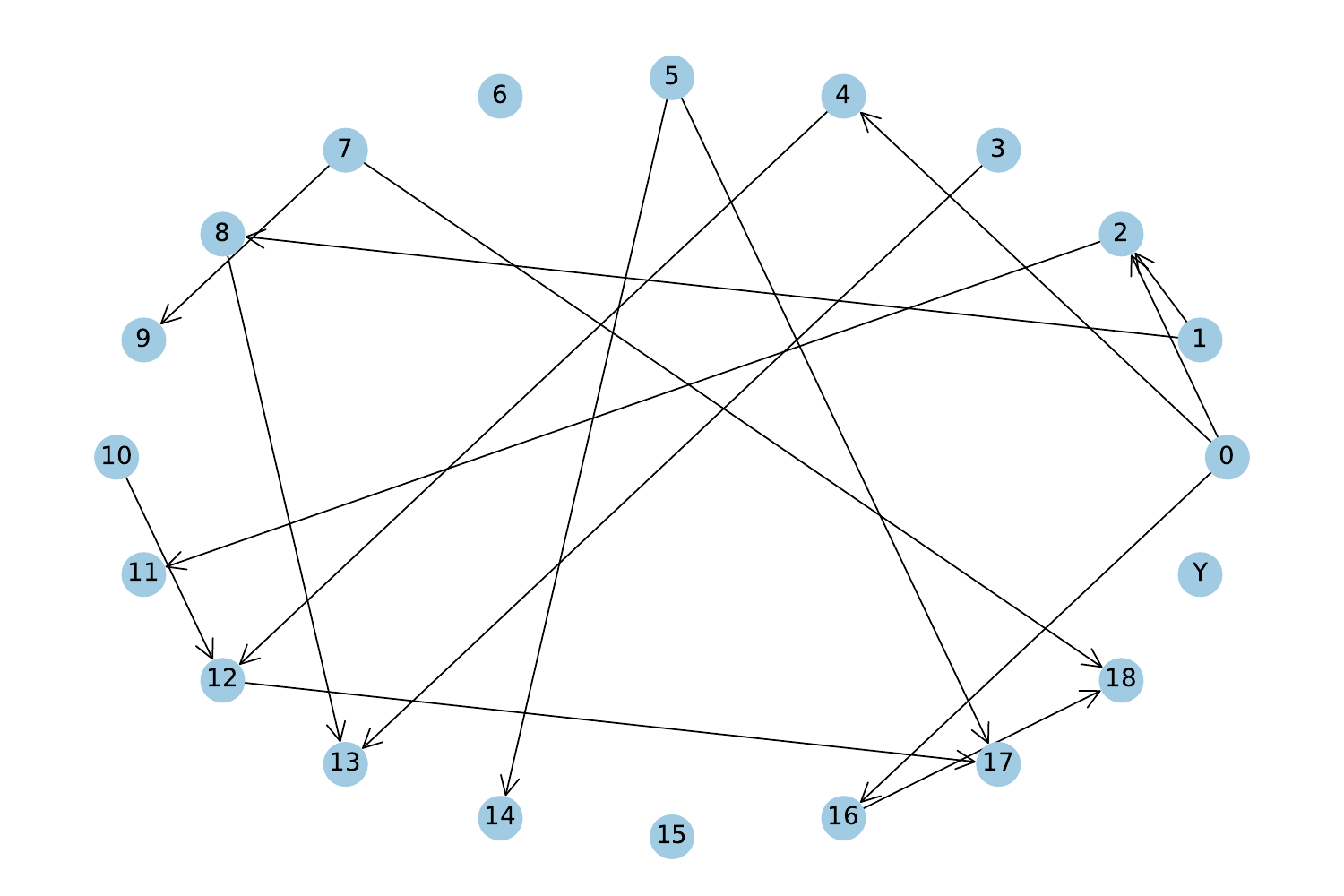} 
\end{subfigure}
\begin{subfigure}[] 
  \centering
  \includegraphics[width=0.31\linewidth]{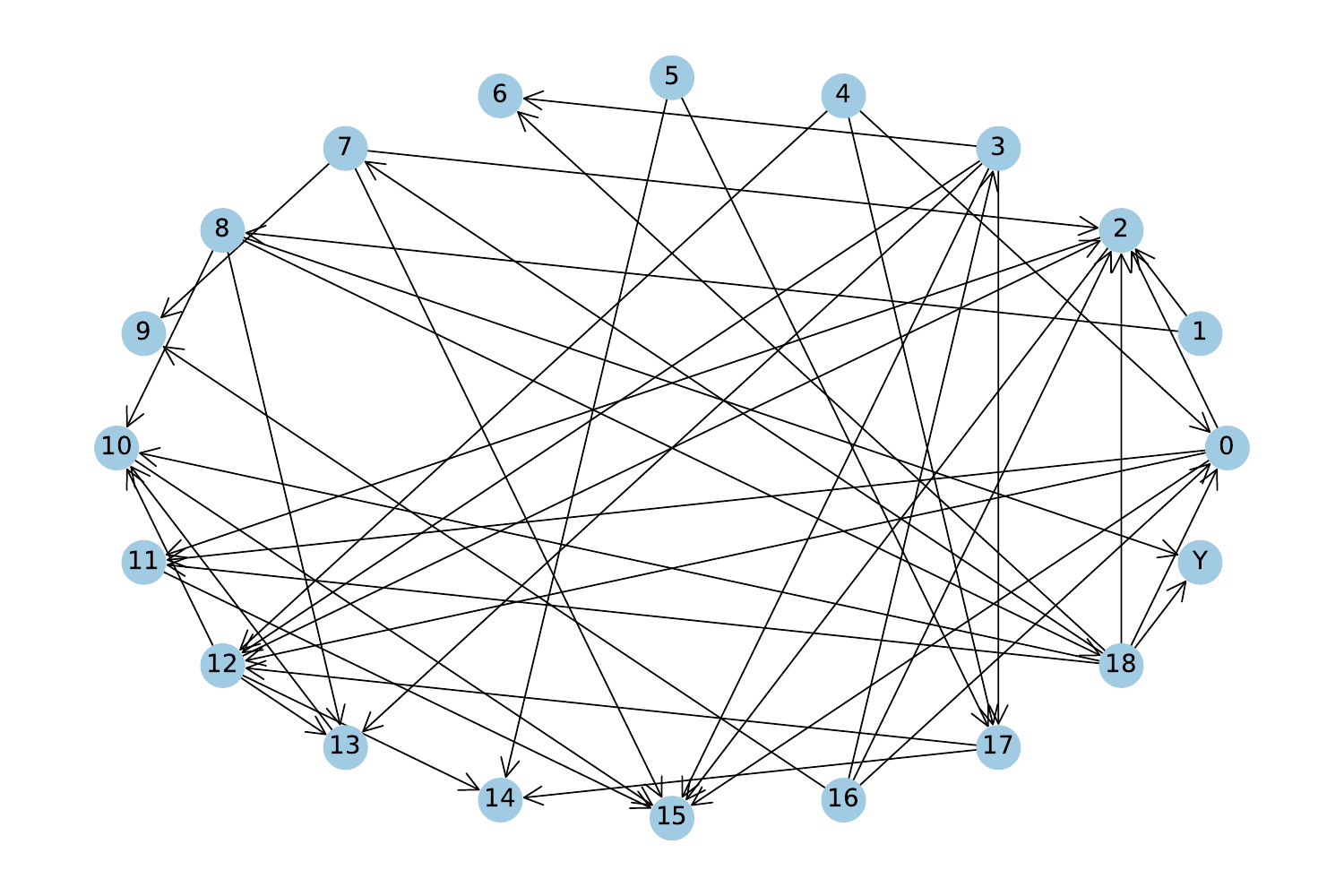}
\end{subfigure}
 \vspace{-0.35cm}
\caption{Graphs under S4 ($n=100$): (a). true whole graph; (b). true NSCG; (c). $\widehat{\mathcal{G}}$ by NSCSL with TE; (d). $\widehat{\mathcal{G}}$ by NSCSL with DE; (e).  $\widehat{\mathcal{G}}$ by NOTEARS; (f). $\widehat{\mathcal{G}}$ by PC; (g). $\widehat{\mathcal{G}}$ by LiNGAM.}
\label{fig_scen_res5}  
 \vspace{-0.35cm}
 \end{figure}

   \begin{figure} 
\centering
\begin{subfigure}[]
  \centering
  \includegraphics[width=0.50\textwidth]{figs/S3_True_Whole_NET.pdf} 
\end{subfigure}%
\begin{subfigure}[]
  \centering
  \includegraphics[width=0.45\textwidth]{figs/S3_True_NS_NET.pdf} 
\end{subfigure}
\begin{subfigure}[] 
  \centering
  \includegraphics[width=0.45\textwidth]{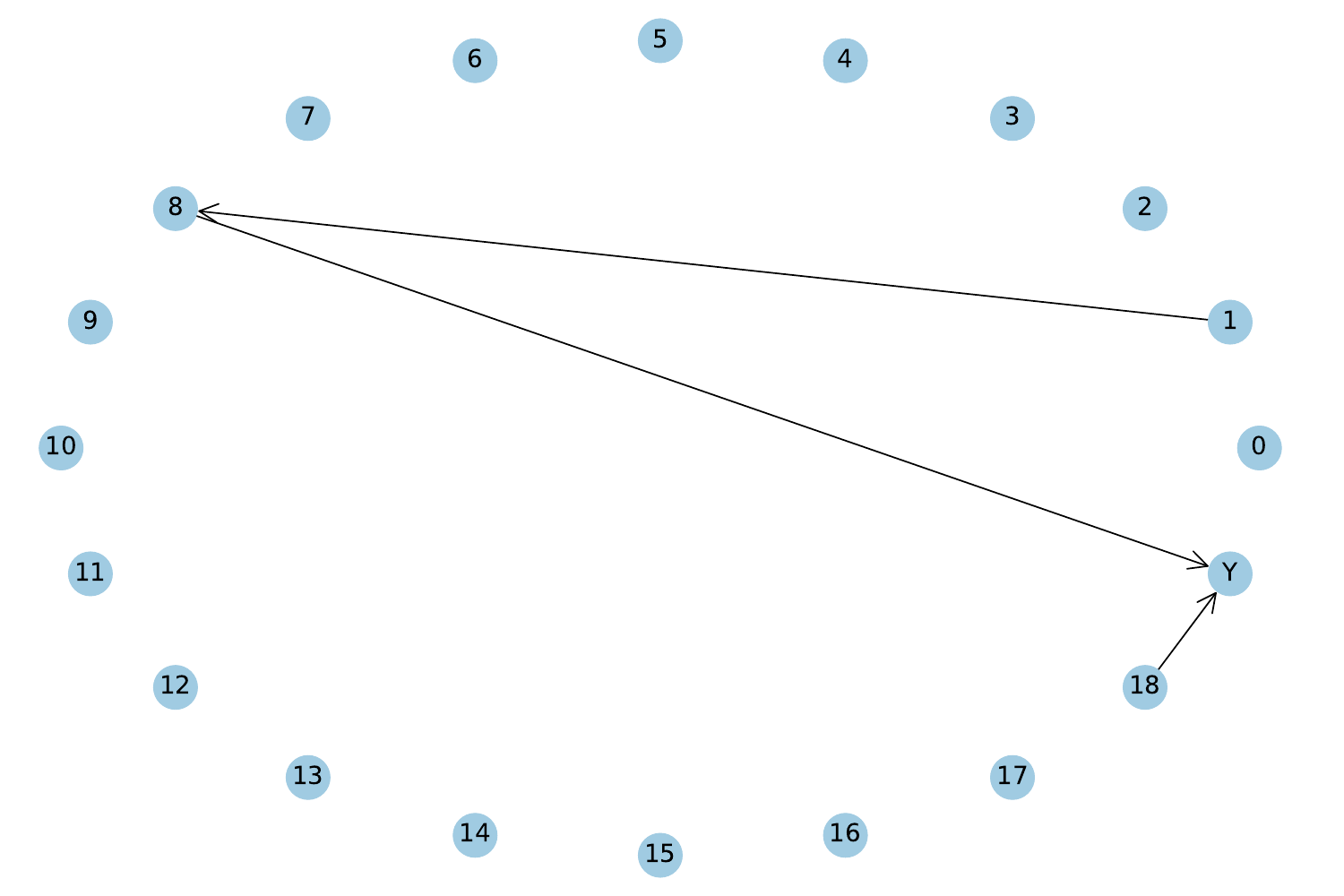} 
\end{subfigure}%
\begin{subfigure}[] 
  \centering
  \includegraphics[width=0.45\textwidth]{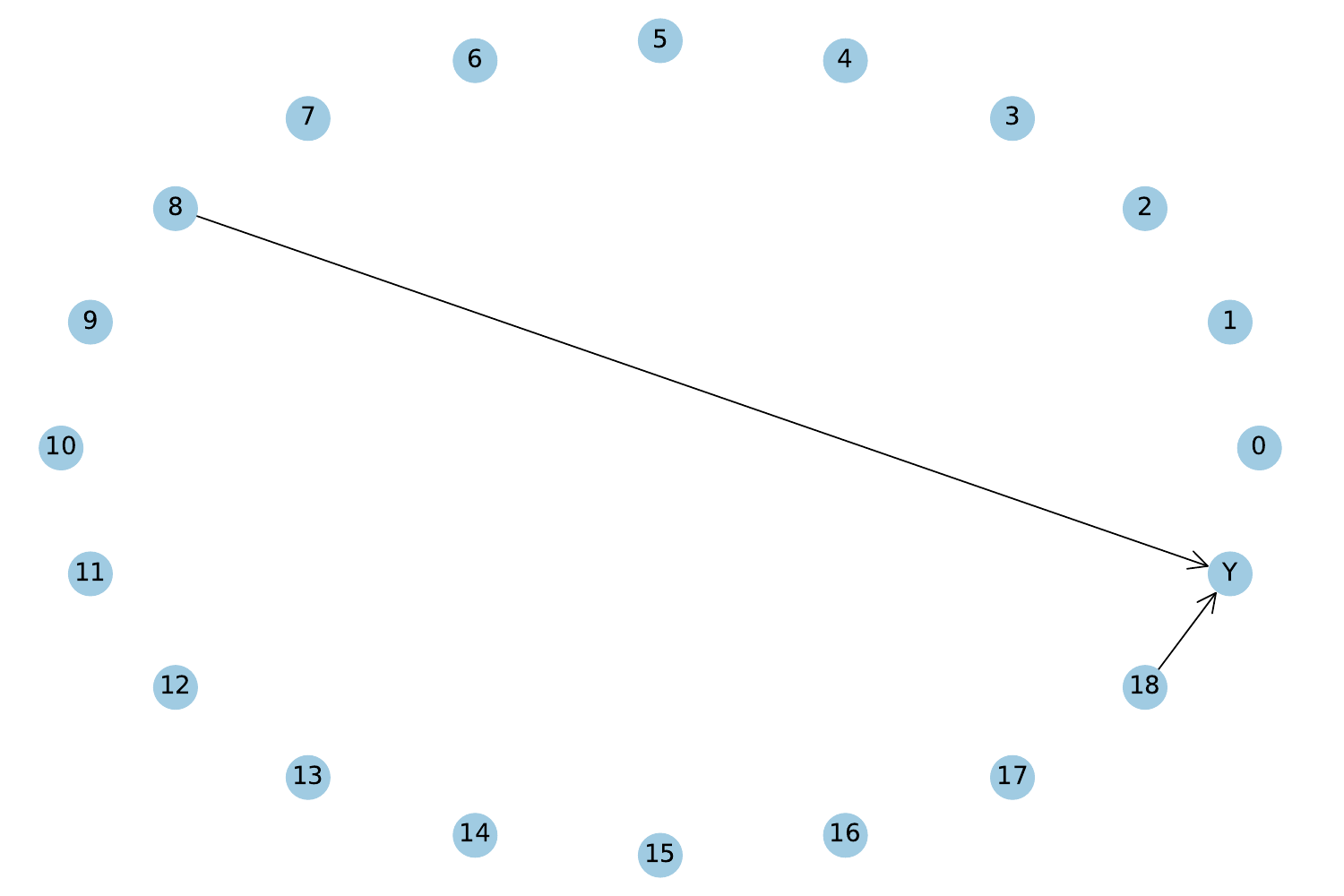} 
\end{subfigure}\\%
\begin{subfigure}[] 
  \centering
  \includegraphics[width=0.31\linewidth]{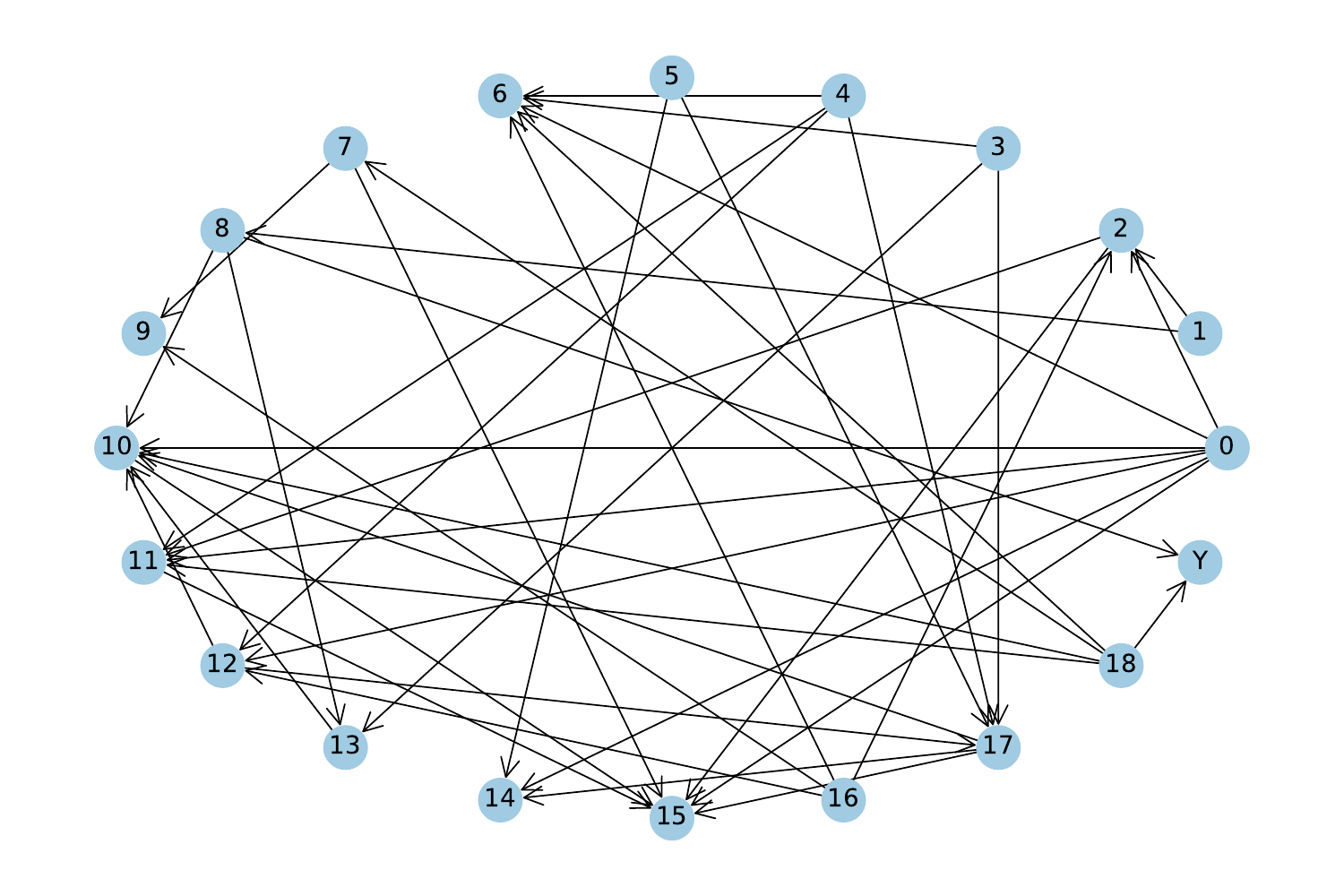} 
\end{subfigure}
\begin{subfigure}[] 
  \centering
  \includegraphics[width=0.31\linewidth]{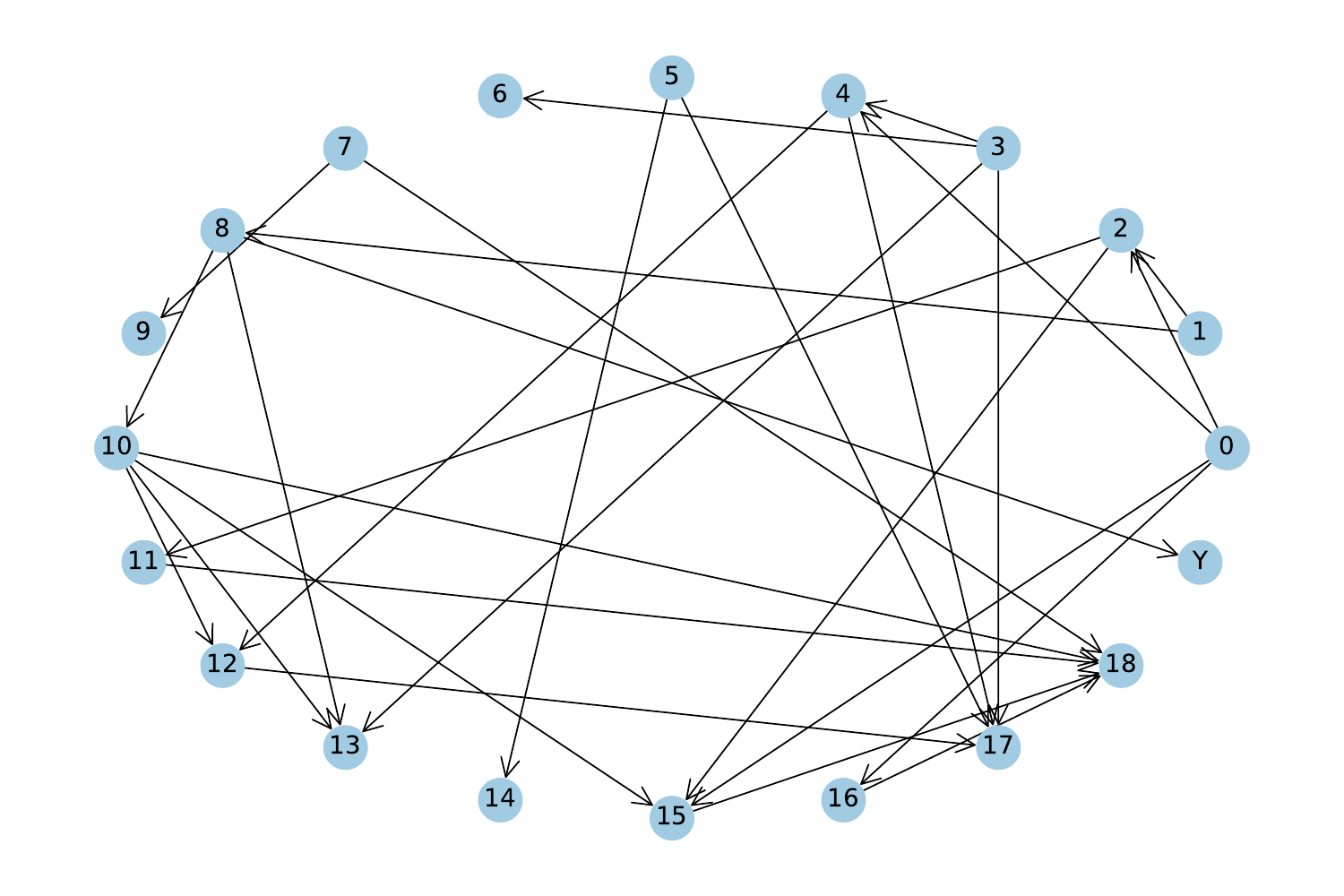} 
\end{subfigure}
\begin{subfigure}[] 
  \centering
  \includegraphics[width=0.31\linewidth]{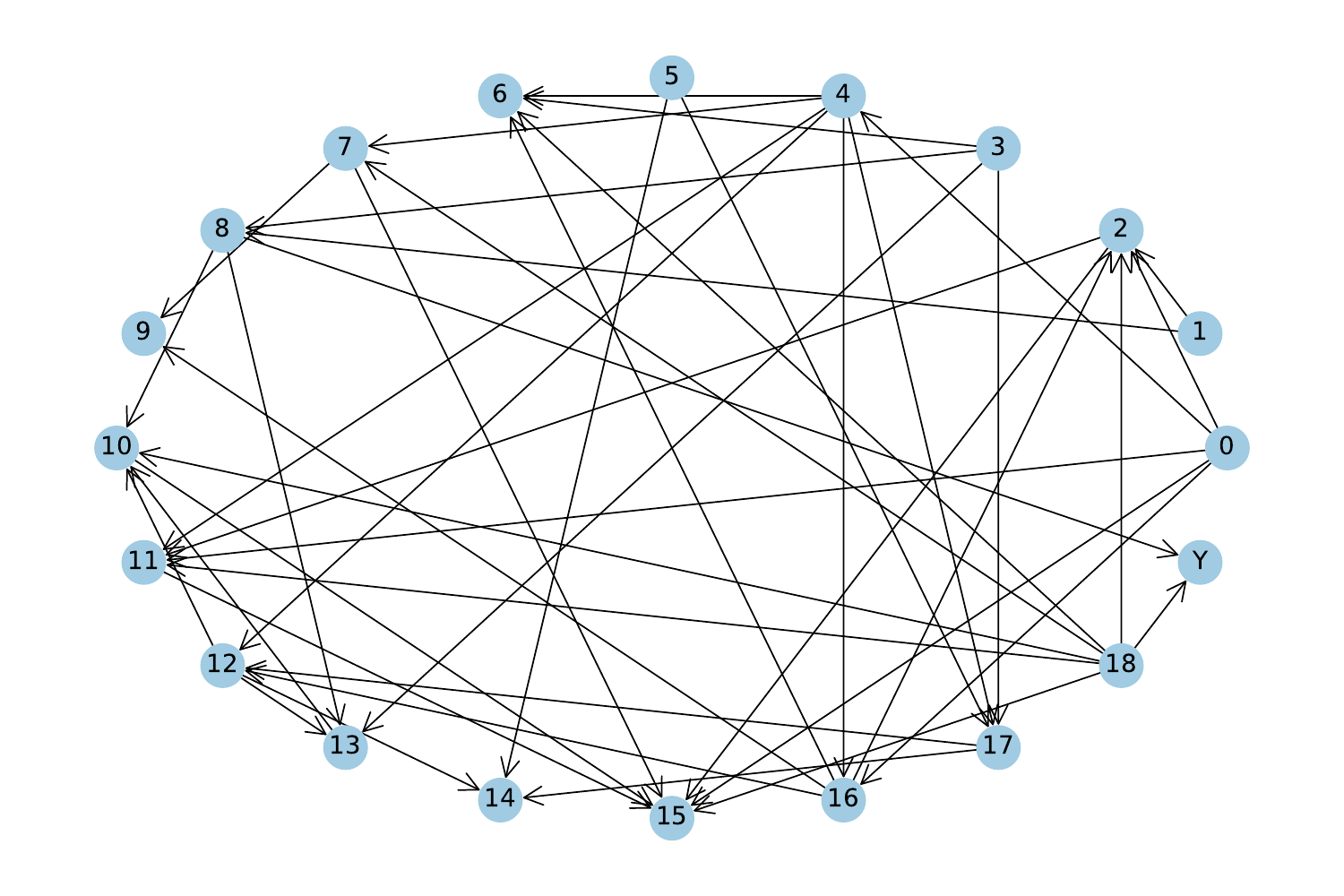}
\end{subfigure}
 \vspace{-0.35cm}
\caption{Graphs  under S4 ($n=300$): (a). true whole graph; (b). true NSCG; (c). $\widehat{\mathcal{G}}$ by NSCSL with TE; (d). $\widehat{\mathcal{G}}$ by NSCSL with DE; (e).  $\widehat{\mathcal{G}}$ by NOTEARS; (f). $\widehat{\mathcal{G}}$ by PC; (g). $\widehat{\mathcal{G}}$ by LiNGAM.}
\label{fig_scen_res7}  
 \vspace{-0.35cm}
 \end{figure}

 \end{appendices} 

 \clearpage
\bibliography{mycite}
\bibliographystyle{nips2023}

\end{document}